%% file: main.tex
\documentclass[10pt,twocolumn,letterpaper]{article}

\usepackage[pagenumbers]{iccv} %

\input{preamble}

\usepackage{colortbl}
\usepackage[dvipsnames]{xcolor}
\definecolor{myGray}{gray}{0.9}

\newcommand{\mc}{\color{red}}

\usepackage[accsupp]{axessibility}
\definecolor{iccvblue}{rgb}{0.21,0.49,0.74}
\usepackage[pagebackref,breaklinks,colorlinks,allcolors=iccvblue]{hyperref}

\def\paperID{11894} %
\def\confName{ICCV}
\def\confYear{2025}

\title{Temporal Rate Reduction Clustering for Human Motion Segmentation}

\author{Xianghan Meng\textsuperscript{\textdagger},~Zhengyu Tong\textsuperscript{\textdagger},~Zhiyuan Huang,~and~Chun-Guang Li\thanks{Corresponding author. \textsuperscript{\textdagger}These two authors are equally contributed.}\\
Beijing University of Posts and Telecommunications, Beijing 100876, P.R. China\\
\texttt{\small \{mengxianghan,tongzhengyu,huangzhiyuan,lichunguang\}@bupt.edu.cn}
}

\begin{document}
\maketitle

\begin{abstract}
Human Motion Segmentation (HMS), which aims to partition videos into non-overlapping human motions, has attracted increasing research attention recently. 
Existing approaches for HMS are mainly dominated by subspace clustering methods, which are grounded on the assumption that high-dimensional temporal data align with a Union-of-Subspaces (UoS) distribution.
However, the frames in video capturing complex human motions with cluttered backgrounds may not align well with the UoS distribution. 
In this paper, we propose a novel approach for HMS, named Temporal Rate Reduction Clustering (\trc), which jointly learns structured  representations and affinity to segment the sequences of frames in video.
Specifically, the structured representations learned by \trc{} enjoy %
temporally consistency and are aligned well with a UoS structure, which is favorable for addressing the HMS task.
We conduct extensive experiments on five benchmark HMS datasets and achieve state-of-the-art performances with different feature extractors. 
The code is available at: \url{https://github.com/mengxianghan123/TR2C}.
\end{abstract} 

\section{Introduction}
\label{Sec:Introduction}
Human motion recognition and analysis have been an active focus of research for %
around two decades~\cite{Gavrila:CVIU1999,Moeslund:CVIU06,Poppe:CVIU07}. 
As a preparatory step, Human Motion Segmentation (HMS) aims to divide sequences of frames in a video into distinct, non-overlapping segments, each representing a specific human motion~\cite{Jhuang:ICCV13}. 
Due to the labor-intensive nature of manually annotating sequences in video, researchers often regard HMS as an unsupervised time-series clustering task.

Roughly, the existing methods for HMS %
typically %
assume that the frames in a video capturing consecutive motions lie on a Union of low-dimensional Subspaces (UoS) embedded in high-dimensional data. %
Thus, subspace clustering methods %
have emerged as a dominated research line %
for the HMS task~\cite{Elhamifar:CVPR09,Liu:ICML10,Lu:ECCV12,You:CVPR16-EnSC,Li:TIP17}.
However, %
an important prior for the HMS task is that the temporally neighboring frames in a video are %
more likely %
belonging to the same human motion. 
To incorporate the temporal continuity between frames in a video, various temporal regularizer %
is introduced to encourage the temporally neighboring frames to be clustered into the same subspace~\cite{Tierney:CVPR14-OSC,Li:ICCV15-TSC}. 
More recently, transfer learning-based %
subspace clustering methods, \eg,~\cite{Wang:AAAI18-TSS,Wang:TIP18-LTS,Zhou:CVPR20-MTS,Zhou:TPAMI22}, have been proposed to further improve the performance %
of HMS. 
Despite the flourish of developing %
these methods for the HMS task in the past decade, %
the clustering performance of HMS still faces a bottleneck.  

For human activities recognition task, as explored in the prior works~\cite{Jiang:TPAMI12-Keck,Ryoo:ICCV09-Ut}, the frames in videos capture both complex human motions and cluttered backgrounds. Thus, it turns out to be more likely that the features of the frames in video can hardly align well with the UoS distribution at all.
As a consequence, it is %
necessary to amend the representation of the frames in video to align with the UoS distribution while performing the motion segmentation.

In this paper, we attempt to jointly learn structured representations that align with the UoS distribution and simultaneously perform motion segmentation. 
To be specific, we propose a novel and effective approach for HMS, termed Temporal Rate Reduction Clustering (\trc), which integrates the Maximal Coding Rate Reduction (\mcr2) principle~\cite{Yu:NIPS20} and a temporal continuity regularization to jointly learn the temporally consistent representations that align with a UoS distribution and the affinity simultaneously.
We solve the problem efficiently by introducing a neural network and leveraging differential programming. 
Extensive experiments are conducted on five HMS benchmark datasets and superior performance confirms the effectiveness of the proposed approach.  %

The contributions of the paper are highlighted as follows. 
\begin{enumerate}
    \item We propose a novel approach, named Temporal Rate Reduction Clustering (\trc), which is able to jointly learn temporally consistent structured representations and affinity to segment the sequences of frames in video.

    {\mc
    }

    \item We demonstrate the effectiveness of our proposed \trc{} with extensive experiments on five benchmark datasets and different feature extractors, achieving state-of-the-art performance. 

\end{enumerate}

To the best of our knowledge, it is for the first time to exploit the \mcr2 principle for clustering %
temporal sequences. 

\section{Related Work}
\label{Sec:Related Work}

In this section, we will review the previous works for HMS at first, and then introduce the relevant work on the principle of maximal coding rate reduction.  

\myparagraph{Probabilistic methods for HMS}
Early human motion segmentation algorithms primarily relied on probabilistic models to model time series data, \eg, Hidden Markov Models~\cite{Smyth:AISworkshop1999}, Dynamic Bayesian Networks~\cite{Murphy:2002} and Auto-regressive Moving Average Models~\cite{Xiong:ICDM02}.
These methods typically employ the Expectation Maximization (EM) algorithm for effective optimization.
Besides, there are also several effective frameworks which extend classical clustering algorithms (\eg, $k$-means) by combining Dynamic Time Warping~\cite{Zhou:CSCVPR10,Zhou:TPAMI12}.

\myparagraph{Subspace clustering based methods for HMS}
Under the assumption that human motion data lie on a UoS, each motion corresponding to a subspace, it is appealing to apply subspace clustering methods to address the HMS task.
To date, various temporal subspace clustering methods are proposed, \eg, Ordered Subspace Clustering (OSC)~\cite{Tierney:CVPR14-OSC} and Temporal Subspace Clustering (TSC)~\cite{Li:ICCV15-TSC}, in which the temporal continuity information is exploited.  
In OSC, the $\|\cdot\|_{1,2}$ norm is introduced as a temporal continuity regularization; in TSC, the temporal continuity graph Laplacian is introduced to encourage neighboring frames to be grouped into the same subspace.
Then, in~\cite{Gholami:CVPR17}, Gaussian Process %
is incorporated to handle missing data to enhance the robustness; 
in \cite{Wang:PR22}, minimum spanning tree is introduced to characterize the affinity between neighboring frames with less redundancy.
In addition, transfer learning is also introduced to align the source domain and target domain by optimizing a projection~\cite{Wang:AAAI18-TSS,Wang:TIP18-LTS} or learning multi-mutual consistency and diversity across different domains~\cite{Zhou:CVPR20-MTS,Zhou:TPAMI22}. 
However, the performance of the methods mentioned above is still unsatisfactory due to the data deviating from the UoS distribution.

\myparagraph{Representation learning based subspace clustering methods for HMS}
To learn effective temporal representations for HMS, %
in \cite{Bai:ICDM20}, a dual-side auto-encoder is introduced 
to learn representations assisted with temporal consistency constraints; after that, in \cite{Bai:TIP22}, a velocity guidance mechanism is leveraged %
for the better capturing of changes between different motions; 
in \cite{Dimiccoli:ICIP19}, non-local self-similarity is introduced to form the representations of each frame; %
in \cite{Dimiccoli:TIP20}, graph consistency is introduced to regularize the learned representations.
More recently, %
in \cite{Dimiccoli:ICCV21-GCTSC}, an approach termed Graph Constraint Temporal Subspace Clustering (GCTSC) is developed in which graph consistency-based representation learning is combined with temporal subspace clustering (TSC). %
Unfortunately, %
in these aforementioned methods, there is no evidence to demonstrate that the learned representations are suitable or well aligned with the UoS distribution.

\myparagraph{\mcr2 principle}
In supervised learning, a so-called Maximal Coding Rate Reduction (\mcr2) principle is proposed %
to learn discriminative and diverse features that conform to a UoS distribution~\cite{Yu:NIPS20,Wang:ICML24}.
In the unsupervised learning field, %
the \mcr2 principle is leveraged to perform image clustering, assisted with contrastive learning in \cite{Li:Arxiv22}.
Then, an approach called Manifold Linearizing and Clustering (MLC) is presented in \cite{Ding:ICCV23}, which incorporates a doubly stochastic affinity into the \mcr2 framework, and in~\cite{Chu:ICLR24} %
MLC is further evaluated %
on visual datasets with pretrained CLIP features~\cite{Radford:ICML21-CLIP}, achieving excellent performance. 
Nonetheless, there is no prior work to address the HMS task with the \mcr2 framework to date.

\section{Our Method}
\label{Sec:Method}

\begin{figure*}[t]
    \centering
    \includegraphics[trim=50pt 0pt 50pt 0pt, clip,width=0.95\linewidth]{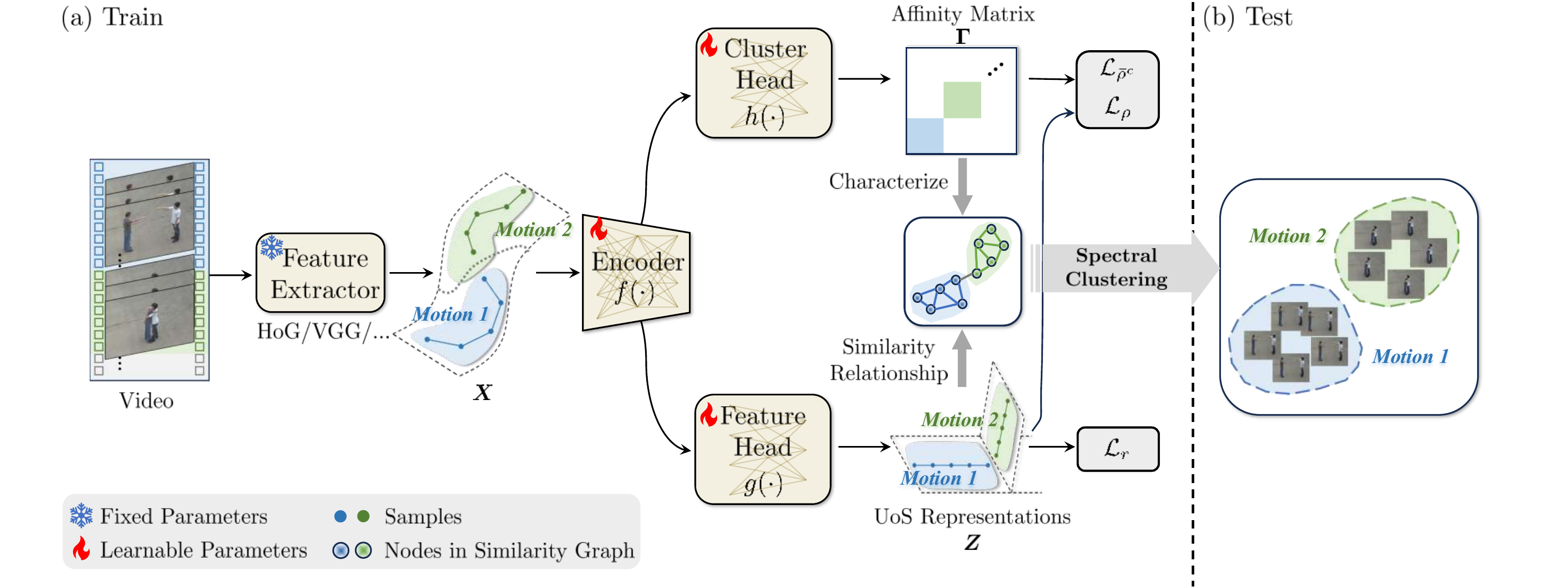}
    \caption{\textbf{The Framework of \trc.} Structured representations %
    and affinity are jointly learned in \trc{} to %
    facilitate motion segmentation.}
    \label{fig:framework}
\end{figure*}

In this section, we will first formulate 
a novel optimization problem for the HMS %
task %
to learn simultaneously efficient representations and affinity for segmentation. %
Then, we will develop a differential programming approach 
to solve the problem efficiently. 

\subsection{Problem Formulation}
\label{Sec:Problem Formulation}

Given a video consisting of $N$ frames $\mathcal{D}=\{{\mathcal{I}}_i\}_{i=1}^N$, HMS aims to %
group each frame into one of %
prescribed human motions.
Denote $\X=[\x_i,\ldots,\x_N]$ as the sequence of extracted features from each frame, which are typically used as the input data of HMS frameworks.

In the HMS task, the extracted features from the sequence of frames corresponding to different motions are typically assumed to approximately lie on a Union of Subspaces (UoS), where each subspace is spanned by the features of the frames belonging to a specific motion.
Based on such a UoS assumption, %
subspace clustering methods have emerged as the dominant approaches for the HMS task~\cite{Elhamifar:CVPR09,Liu:ICML10,Lu:ECCV12,Tierney:CVPR14-OSC,Li:ICCV15-TSC}.
However, the segmentation performance of these methods seems to be stuck, %
primarily due to the misalignment between the data and the UoS assumption, %
especially in the scenarios involving complex motions and cluttered background~\cite{Jiang:TPAMI12-Keck,Ryoo:ICCV09-Ut}.
To address this limitation, it is crucial to learn structured representation of the data, \ie, a mapping function $\mathcal{F}:\X\rightarrow \Z$ that transforms the %
input $\X$ into representation $\Z$ with more favorable distribution, thereby enhancing segmentation performance. %

\myparagraph{Given the partition $\boldPi$, learning UoS representations}
The principle of Maximal Coding Rate Reduction (\mcr2) %
guarantees to learn representations $\Z$ which align with UoS structure in supervised setting, where the coding rate quantifies the minimum average coding length required to compress the representations which are drawn from a mixture of Gaussian distributions in lossy compression scenario~\cite{Ma:PAMI07}.
To be specific, given a set of diagonal matrices $\boldPi= \{\boldPi_j\in\left\{0,1\right\}^{N\times N}\}_{j=1}^K$, where the $m$-th diagonal element of $\boldPi_j$ indicates whether the $m$-th sample belongs to the $j$-th class,
the \mcr2 principle aims to optimize:
\begin{align}
\label{eq:mcr2}
\begin{aligned}
      \mathop{\max}_{\Z}\quad&\Delta \rho(\Z,\boldPi,\epsilon) :=\rho(\Z,\epsilon)-\rho^c(\Z,\epsilon~|~\boldPi),\\
    \mathrm{s.t.}\quad &\|\z_i\|_2^2=1, \quad \text{for}~i=1,\cdots,N,
\end{aligned}
\end{align}
where 
\begin{equation}
    \rho(\Z,\epsilon)\coloneqq\frac{1}{2}\log\det(\boldsymbol{I}+\frac{d}{N\epsilon^2}\Z\Z^\top)
\end{equation}
is the coding rate of the representations $\Z\in\mathbb{R}^{d\times N}$ with respect to precision $\epsilon>0$ and 
\begin{equation}
    \rho^c(\Z,\epsilon~|~\boldPi)\coloneqq\sum_{j=1}^K\frac{\tr(\boldPi_j)}{2N}\log\det(\boldsymbol{I}+\frac{d}{\tr(\boldPi_j)\epsilon^2}\Z\boldPi_j\Z^\top)
\end{equation}
is the sum of coding rate of the representations $\Z_j$ from each class indicated by $\boldPi_j$.

From a geometric perspective, the $\log\det(\cdot)$-based function, which serves as a concave relaxation of $\mathrm{rank}(\cdot)$, measures the volume of the representations. 
By jointly maximizing the holistic volume of the representations while minimizing their intra-class volumes, the representations naturally conform to a union of orthogonal subspaces distribution.\footnote{Please refer to~\cite{Yu:NIPS20,Wang:ICML24} for rigorous proofs.}

\myparagraph{Given $\Z$, learning the partition $\boldPi$}
For the unsupervised HMS task, %
the representations are fixed and assumed lying on a UoS %
and we aim to find the assignment matrices set $\boldPi$. In such case, the correct assignment matrices set $\boldPi$ would sort these data into its own subspace and thus make the coding rate minimized~\cite{Ma:PAMI07}. Therefore, the task of learning $\boldPi$ can be formulated into an optimization problem as follows:
\begin{align}
\label{eq:min-c-r2-for-PI}
\begin{aligned}
      \mathop{\min}_{\boldPi}\quad \rho^c(\Z,\epsilon~|~\boldPi). 
\end{aligned}
\end{align}
However, for an unsupervised HMS task, the representations $\Z$ might not well align with a UoS distribution. 
\myparagraph{Jointly learning $\Z$ and $\boldPi$}
When both $\Z$ and 
$\boldPi$ are jointly learned, we have an optimization problem as follows:
\begin{align}
\label{eq:min-c-r2-for-both-Z-PI}
\begin{aligned}
    \mathop{\min}_{\Z, \boldPi}\quad & \rho^c(\Z,\epsilon~|~\boldPi), \\
    \mathrm{s.t.}\quad &\|\z_i\|_2^2=1, \quad \text{for}~i=1,\cdots,N. 
\end{aligned}
\end{align}
An important prior in the HMS task is that %
the temporally neighboring frames in the video are more likely %
belonging to the same motion.
Therefore, it is helpful to introduce a temporal continuity regularizer, which %
encourages learning the representations that are of temporal consistency between neighboring frames and thus facilitates the segmentation task~\cite{Tierney:CVPR14-OSC,Li:ICCV15-TSC}.
However, the temporal consistency among frames in a video is ignored in problem \eqref{eq:min-c-r2-for-both-Z-PI}. 

\myparagraph{Jointly learning temporally consistent representation $\Z$ and $\boldPi$}
Analog to~\cite{Li:ICCV15-TSC}, we introduce a temporal Laplacian regularizer, which is defined as follows:
\begin{equation}
    r(\Z)\coloneqq\frac{1}{2}\sum_{i=1}^N\sum_{j=1}^Nw_{ij}\|\z_i-\z_j\|_2^2=\tr(\Z\boldsymbol{L}\Z^\top),
\end{equation}
where $\boldsymbol{L}=\Diag(\W\boldsymbol{1}_N)-\W$ is the graph Laplacian matrix, $\boldsymbol{1}_N$ is a column vector of dimension $N$ consisting of 1, 
and the affinity $\W=\{w_{ij}\}_{i,j=1}^N$ is defined as:
\begin{equation}
    w_{ij}\coloneqq\left\{\begin{array}{lr}
    1,     &  \text{if }|i-j|\leq\frac{s}{2},\\
    0,     & \text{otherwise,}
    \end{array}\right.
\end{equation}
where $s$ is the size of a sliding window. %
The temporal Laplacian regularizer $r(\Z)$ conforms the similarity relationship among the learned representations to the pre-defined affinity $\W$, which geometrically governs the smoothness of learned representations along the temporal dimension.

By taking into account the temporal continuity prior, we formulate an optimization problem as follows:
\begin{align}
\label{eq:min-c-r2-for-both-Z-PI-temporal}
\begin{aligned}
    \mathop{\min}_{\Z, \boldPi}\quad & \rho^c(\Z,\epsilon~|~\boldPi) + \lambda r(\Z), \\
    \mathrm{s.t.}\quad &\|\z_i\|_2^2=1, \quad \text{for}~i=1,\cdots,N, 
\end{aligned}
\end{align}
where $\lambda >0$ is a hyper-parameter. 
While problem \eqref{eq:min-c-r2-for-both-Z-PI-temporal} looks appealing, unfortunately, there exists undesired trivial solutions $(\Z_\star, \boldPi_\star)$ that all embeddings are collapsed.\footnote{ %
The existence of collapsed solutions often leads to an over-smoothing issue. For example, the over-smoothing issue in graph neural networks results in indistinguishable node embeddings~\cite{Li:AAAI18}; 
the over-smoothing issue in deep subspace clustering causes catastrophically collapsed representations~\cite{Haeffele:ICLR21}. 
}

To prevent the collapsed solution, %
inspired by \cite{Yu:NIPS20, Meng:ICLR25}, we add %
a maximizing total coding rate based regularization term 
into problem \eqref{eq:min-c-r2-for-both-Z-PI-temporal}. Thus we have an optimization problem as follows:
\begin{align}
\label{Eq:TR2C}
\begin{aligned}
    \mathop{\min}_{\Z,\boldPi}\quad& - \rho(\Z,\epsilon) + \lambda_1 \rho^c(\Z,\epsilon~|~\boldPi) + \lambda_2 r(\Z),\\
    \mathrm{s.t.}\quad &\|\z_i\|_2^2=1, \quad \text{for}~i=1,\cdots,N,
\end{aligned}
\end{align}
where $\lambda_1,\lambda_2>0$ are two hyper-parameters.
We call this framework in \eqref{Eq:TR2C} a Temporal Rate Reduction Clustering (\trc).

\myparagraph{Remarks} 
The total coding rate term $\rho(\Z,\epsilon)$ in our \trc{} offers a %
tighter approximation to the $\mathrm{rank}(\Z)$~\cite{Fazel:ACC03-logdet,Ma:PAMI07,Yu:NIPS20,Liu:NIPS22-MEC}.
Minimizing $- \rho(\Z,\epsilon)$ together with $\rho^c(\Z,\epsilon~|~\boldPi)+r(\Z)$ can help prevent over-compressing the learned representations.
Although the \trc{} problem appears to be rational, it is still %
quite challenging to solve %
due to the combinatorial nature.

\subsection{Optimization}
\label{Sec:Optimization}

Rather than directly optimizing $\boldPi$, in this paper, following~\cite{Ding:ICCV23}, we introduce a doubly stochastic affinity matrix $\boldGamma \in \boldsymbol\Xi$, where $\boldsymbol\Xi :=\{\boldGamma\in\mathbb{R}_{+}^{N\times N}~|~\boldGamma\boldsymbol{1}=\boldsymbol{1},\boldGamma^\top\boldsymbol{1}=\boldsymbol{1}\}$, then the term $\rho^c$ is relaxed to:
\begin{equation}
    \bar \rho^c(\Z,\epsilon~|~\boldGamma)\coloneqq\frac{1}{N}\sum_{j=1}^N\log\det(\boldsymbol{I}+\frac{d}{\epsilon^2}\Z\Diag(\boldGamma_j)\Z^\top)
\end{equation}
where $\boldGamma_j$ being the $j$-th column of $\boldGamma$.

Similar to~\cite{Zhang:CVPR21-SENet, Ding:ICCV23}, we consult the differential programming approach to solve the continuously relaxed problem. 
We re-parameterize $\Z$ and $\boldGamma$ through properly designed neural networks and optimize over the parameters of the neural networks. %
To be specific, we introduce an encoder $f(\cdot)$, a feature head $g(\cdot)$ and a cluster head $h(\cdot)$ to form our implementation framework. 
Formally, the outputs of feature head and cluster head are computed by:
\begin{align}
\label{Eq:Z and Y}
\begin{aligned}
    \z_i &= g\left(f\left(\boldsymbol{x}_i\right)\right),\\
    \y_i &= h\left(f\left(\boldsymbol{x}_i\right)\right),   
\end{aligned}
\end{align}
for all $i\in\{1,\ldots,N\}$.
Then, after the normalization of the outputs $\Tilde{\z}_i=\z_i/\|\z_i\|_2$, $\Tilde{\y}_i=\y_i/\|\y_i\|_2$, 
we compute the affinity $\boldGamma$ by:
\begin{equation}
\label{Eq:Gamma}
    \boldGamma=\mathcal{P}_\Xi(\Tilde{\Y}^\top\Tilde{\Y}),
\end{equation}
where $\mathcal{P}_\Xi(\cdot)$ is a sinkhorn projection~\cite{Cuturi:NIPS13-sinkhorn}, which is a differentiable projection to doubly stochastic matrix. Note that owing to the normalization for $\z_i$ and the sinkhorn projection, the constraints in \eqref{Eq:TR2C} and for defining the doubly stochastic $\Gamma$ can be (automatically) satisfied.

Equipped with the reparameterization, rather than directly optimizing over $\Z$ and $\boldGamma$, we instead update the parameters of the networks by back-propagation. Specifically, we denote the parameters in networks $f(\cdot)$, $g(\cdot)$ and $h(\cdot)$ as  $\boldsymbol \theta$. Then %
the parameters $\boldsymbol \theta$ can be updated by minimizing the following loss functions:

\begin{align}
\label{Eq:loss}
\mathcal{L}=-\mathcal{L}_{\rho}+ \lambda_1 \mathcal{L}_{\bar\rho^c}+ \lambda_2 \mathcal{L}_{r},
\end{align}
where 
\begin{align}
\begin{aligned}
    \mathcal{L}_{\rho}&\coloneqq\frac{1}{2}\log\det(\boldsymbol{I}+\frac{d}{N\epsilon^2}\Z(\boldsymbol \theta)\Z(\boldsymbol \theta)^\top),\\
    \mathcal{L}_{\bar\rho^c}&\coloneqq \sum_{j=1}^N\frac{1}{N}\log\det(\boldsymbol{I}+\frac{d}{\epsilon^2}\Z(\boldsymbol \theta)\Diag(\boldGamma_j(\boldsymbol \theta))\Z(\boldsymbol \theta)^\top),\\
    \mathcal{L}_{r}&\coloneqq \tr(\Z(\boldsymbol \theta)\boldsymbol{L}\Z(\boldsymbol \theta)^\top).
\end{aligned}
\end{align}

Finally, %
having the affinity $\boldGamma(\boldsymbol \theta)$, 
we apply spectral clustering \cite{Shi:TPAMI00} to yield HMS results %
as in~\cite{Li:ICCV15-TSC,Dimiccoli:ICCV21-GCTSC}. %
For clarity, we illustrate the overall framework of our \trc{} %
in Figure~\ref{fig:framework} and summarize the whole training procedure in Algorithm~\ref{alg:algorithm}.

\begin{figure}[h]
\centering
\resizebox{\linewidth}{!}{
\begin{minipage}{\linewidth}
\begin{algorithm}[H]
\caption{Temporal Rate Reduction Clustering (\trc)}
\label{alg:algorithm}
\begin{algorithmic}[1] %
\linespread{1.1} %
\item[\textbf{Input:}] Input Features $\boldsymbol{X}\in \mathbb{R}^{D\times N}$, hyper-parameters $\lambda_1,\lambda_2$, number of iterations $T$, network parameters $\boldsymbol \theta$, %
learning rate $\eta$
\item[\textbf{Initialization:}] Randomly initialize parameters $\boldsymbol \theta$

\For{$t=1,\dots,T$} 
    \State\textit{\# Forward propagation}
    \State Compute $\boldsymbol{Z(\boldsymbol \theta)}$ and  $\boldsymbol{Y(\boldsymbol \theta)}$ by (\ref{Eq:Z and Y})

    \State Compute affinty $\boldsymbol{\Gamma(\boldsymbol \theta)}$ by (\ref{Eq:Gamma})
    
    \State\textit{\# Backward propagation}
    \State Compute loss $\mathcal{L}$ by (\ref{Eq:loss})
    \State Compute $\nabla_{\boldsymbol \theta} \doteq \frac{\partial \mathcal{L}}{\partial {\boldsymbol \theta}} $

    \State Set ${\boldsymbol \theta} \gets {\boldsymbol \theta} - \eta \cdot  \nabla_{\boldsymbol \theta}$

\EndFor
\item[\textbf{Test:}]  Apply spectral clustering on $\boldGamma(\boldsymbol \theta)$. %
\end{algorithmic}
\end{algorithm}
\end{minipage}
}
\end{figure}

\section{Experiments}
\label{Sec:Experiments}
To evaluate the effectiveness of our proposed approach, following~\cite{Li:ICCV15-TSC,Wang:AAAI18-TSS,Wang:TIP18-LTS,Zhou:CVPR20-MTS,Zhou:TPAMI22,Bai:TIP22,Dimiccoli:ICCV21-GCTSC}, we conduct experiments on five benchmark datasets, including Weizmann action dataset (Weiz)~\cite{Gorelick:TPAMI07-Weiz}, Keck gesture dataset (Keck)~\cite{Jiang:TPAMI12-Keck}, UT interaction dataset (UT)~\cite{Ryoo:ICCV09-Ut}, Multi-model Action Detection dataset (MAD)~\cite{Huang:ECCV14-MAD}, and UCF-11 YouTube action dataset (YouTube) \cite{Liu:CVPR09-youtube}. 
We defer the description of these datasets to Appendix~\ref{Sec:Datasets}.

\subsection{Experimental Setups}

For datasets Weiz, Keck, UT, and MAD, following the baselines, we conduct experiments based on 324-dimensional HoG features~\cite{Zhu:CVPR06-HoG} of each frame.\footnote{The HoG features are available at \url{https://github.com/wanglichenxj/Low-Rank-Transfer-Human-Motion-Segmentation}.}
For the YouTube dataset, following \cite{Zhou:TPAMI22}, we conduct experiments on 1000-dimensional pretrained VGG-16 features~\cite{Simonyan:ICLR15}.
To explore the limit of \trc, we also evaluate the performance with the features extracted from the image encoder of pretrained CLIP model~\cite{Radford:ICML21-CLIP}. Key information about datasets is summarized in Table~\ref{tab:dataset-information}.

\begin{table}[h]
  \centering
  \caption{\textbf{Key information about datasets used by training.} We show the number of sequences, the number of motions, the maximal number of frames of all the sequences, dimension of HoG, VGG and CLIP features.}
  \resizebox{0.95\linewidth}{!}{
    \begin{tabular}{lP{1.cm}P{1.2cm}P{1.2cm}P{1.8cm}P{1.1cm}}
    \toprule
    \rowcolor{myGray} &  &  &  & Dim & Dim \\
    \rowcolor{myGray} \multirow{-2}{*}{Datasets} & \multirow{-2}{*}{\#Seq}& \multirow{-2}{*}{\#Motions} & \multirow{-2}{*}{\#Frames} & (HoG/VGG) & (CLIP) \\
    \midrule
    Weiz & 9     & 10  & 826  & 324 (HoG) & 768 \\
    Keck  & 4     & 10  & 1245  & 324 (HoG) & 768 \\
    UT    & 10    & 6  &  650  & 324 (HoG) & - \\
    MAD   & 40    & 10  & 1379  & 324 (HoG) & - \\
    YouTube   & 4    & 10  &  2572  & 1000 (VGG) & 768\\
    \bottomrule
    \end{tabular}}
  \label{tab:dataset-information}%
\end{table}%

\begin{table*}[tbp]
  \centering
  \caption{\textbf{The performance of \trc{} comparing to state-of-the-art algorithms.}}
    \vskip -0.1 in
  \resizebox{0.95\textwidth}{!}{
    \begin{tabular}{lcccccccccc}
    \toprule
     & \multicolumn{2}{c}{\textbf{Weiz}} & \multicolumn{2}{c}{\textbf{Keck}} & \multicolumn{2}{c}{\textbf{UT}} & \multicolumn{2}{c}{\textbf{MAD}} & \multicolumn{2}{c}{\textbf{YouTube}}\\
\cmidrule(lr){2-3}\cmidrule(lr){4-5}\cmidrule(lr){6-7}\cmidrule(lr){8-9} \cmidrule(lr){10-11}
\multirow{-2}{*}{\textbf{Method}} & ACC   & NMI   & ACC   & NMI   & ACC   & NMI   & ACC   & NMI & ACC   & NMI \\
    \midrule
    LRR~\cite{Liu:ICML10}   & 43.82  & 36.38  & 48.62  & 42.97  & 40.51  & 41.62  & 22.49  & 23.97  & - & -\\
    RSC~\cite{Li:CVPR19-RSC}   & 41.12  & 48.94  & 34.85  & 32.52  & 36.64  & 18.81  & 37.30  & 34.18  & - & -\\
    SSC~\cite{Elhamifar:CVPR09}   & 60.09  & 45.76  & 38.58  & 31.37  & 49.98  & 43.89  & 47.58  & 38.17  & - & -\\
    LSR~\cite{Lu:ECCV12}   & 50.93  & 50.91  & 45.48  & 48.94  & 43.22  & 51.83  & 36.67  & 39.79 & 93.16 & 96.64 \\
    \midrule
    OSC~\cite{Tierney:CVPR14-OSC}   & 70.47  & 52.16  & 59.31  & 43.93  & 68.77  & 58.46  & 55.89  & 43.27  & - & -\\
    TSC~\cite{Li:ICCV15-TSC}   & 61.11  & 81.99  & 47.81  & 71.29  & 53.40  & 75.93  & 55.56  & 77.21 & 90.40 & 95.00 \\
    \midrule
    TSS~\cite{Wang:AAAI18-TSS}   & 62.08  & 85.09  & 53.95  & 80.49  & 59.44  & 78.78  & 57.92  & 82.86  & 62.94 & 88.20 \\
    LTS~\cite{Wang:TIP18-LTS}   & 63.91  & 85.99  & 55.09  & 82.26  & 62.99  & 81.28  & 59.80  & 82.11 & 62.26 & 88.98 \\
    MTS~\cite{Zhou:CVPR20-MTS}   & 64.36  & 83.71  & 60.10  & 82.70  & 64.33  & 82.39  & 61.63  & 83.14  & 64.40 & 81.41\\
    CDMS~\cite{Zhou:TPAMI22} & 65.05 & 83.75 & 62.07 & 80.40 & 66.43 & 83.06 & 65.36 & 82.51 & 67.98 & 91.33\\
    \midrule
    MLC~\cite{Ding:ICCV23}   & 37.30  & 45.86  & 47.29  & 49.78  & 45.79  & 35.30  & 30.27  & 29.40  & 94.82 & 97.30 \\
    DGE~\cite{Dimiccoli:TIP20} & - & - & 72.00 & 83.00 & - & -& 67.00 & 82.00 & - & - \\
    DSAE~\cite{Bai:ICDM20} & 61.99 & 78.79 & 57.53 & 74.07 & 60.06 & 79.50 & 55.48 & 77.34 & - & - \\
    VSDA~\cite{Bai:TIP22} &62.87 & 79.92 & 58.04 & 73.97 & 62.03 & 82.26 & 56.06 & 77.70 & - & - \\
    GCTSC~\cite{Dimiccoli:ICCV21-GCTSC} & 85.01  & 90.53  & 78.64  & 83.25  & 87.00  & 82.56  & 82.97  & 84.71  & 95.79 & 96.30 \\
    {\bf Our \trc{}}  & $\textbf{94.12}{\scriptstyle\pm1.20}$ & $\textbf{95.91}{\scriptstyle\pm0.66}$ & $ \textbf{83.50}{\scriptstyle\pm1.98}$ & $\textbf{85.63}{\scriptstyle\pm0.86}$ & $\textbf{93.54}{\scriptstyle\pm1.05}$ & $\textbf{91.83}{\scriptstyle\pm0.65}$ & $\textbf{83.08}{\scriptstyle\pm0.62}$ & $\textbf{86.86}{\scriptstyle\pm0.37}$ & $\textbf{97.96}{\scriptstyle\pm1.53}$ & $\textbf{98.96}{\scriptstyle\pm0.83}$ \\
    \bottomrule
    \end{tabular}%
    }
    \vskip -0.1 in
  \label{tab:benchmark-performance}%
\end{table*}%

We %
use clustering accuracy (ACC) and normalized mutual information (NMI) as the evaluation metrics.
The performance is %
reported by taking the mean and standard deviation after running the experiments with 5 different random seeds.

We choose a lightweight neural network architecture, where
the encoder is a two-layer Multi-Layer Perceptron (MLP), with the clustering head and feature head being Fully Connected (FC) layers.
The %
hyper-parameters $\lambda_1$ and $\lambda_2$ are tuned independently for each dataset.
Sliding window size $s$ is fixed as $s=2$ for all datasets.
The sensitivity to hyper-parameters is reported in Figure~\ref{fig:sensitivity} and the hyper-parameter settings are summarized in the Appendix~\ref{sec:hyperpara}.

\subsection{Comparative Results}
We compare the performance of \trc{} on HoG features to subspace clustering algorithms, \eg, LRR~\cite{Liu:ICML10}, RSC~\cite{Li:CVPR19-RSC}, SSC~\cite{Elhamifar:CVPR09}, LSR~\cite{Lu:ECCV12}, temporal regularized subspace clustering algorithms, \eg, OSC~\cite{Tierney:CVPR14-OSC}, TSC~\cite{Li:ICCV15-TSC}, transferable subspace clustering algorithms, \eg, TSS~\cite{Wang:AAAI18-TSS}, LTS~\cite{Wang:TIP18-LTS}, MTS~\cite{Zhou:CVPR20-MTS}, CDMS~\cite{Zhou:TPAMI22}, and representation learning assisted temporal clustering algorithms, \eg, DGE~\cite{Dimiccoli:TIP20}, DSAE~\cite{Bai:ICDM20}, VSDA~\cite{Bai:TIP22}, GCTSC~\cite{Dimiccoli:ICCV21-GCTSC}. The results other than our \trc{} are cited from~\cite{Dimiccoli:ICCV21-GCTSC,Zhou:TPAMI22}.

As shown in Table~\ref{tab:benchmark-performance}, although \trc{} does not utilize additional data through a transfer learning strategy, yet it achieves a clustering accuracy that is 20\% higher than that of the transfer learning-based approach.
\trc{} also outperforms other representation learning assisted HMS algorithms, namely, DGE \cite{Dimiccoli:TIP20}, DSAE \cite{Bai:ICDM20}, VSDA \cite{Bai:TIP22}, GCTSC \cite{Dimiccoli:ICCV21-GCTSC}.
This may stem from the fact that these %
representation learning methods using self-similarity, auto-encoder, or graph consistency cannot substantially improve the structure of the data distribution. %
In contrast, \trc{} explicitly ensures that the learned representations exhibit an desirable %
distribution, contributing to improved separability.
It is noteworthy that the conventional subspace clustering algorithms without using temporal continuity information %
perform well on the YouTube dataset, since the background of YouTube dataset shifts significantly among different motions and the pretrained VGG model extracts semantic meaningful features. 
\trc{} additionally integrates temporally consistency and yields satisfying segmentation result. 
Please refer to Appendix~\ref{sec:Segmentation Results Visualization} for the visualization of segmentation results of \trc{}.

\subsection{More Evaluations}

\myparagraph{Qualitative evaluation of representations}
To have an intuitive comparison, we conduct experiments to visualize the input HoG data and the learned \trc{} representations on the Keck, UT and MAD datasets. %
The visualization results are displayed in Figure~\ref{fig:PCA} (see Appendix~\ref{fig:PCA-supp} for more results).
For each dataset, we use a subset containing 3 different motions for better clarity.
We apply Principal Component Analysis (PCA) for dimension reduction because it performs a linear projection on the input data, well preserving its structure.
 
As can be observed, the input data with raw HoG features (in the first row) %
lie %
on approximately %
one-dimensional manifolds without clear and separable %
a union of subspace structure.
This observation accounts for the reason why the previous HMS approaches %
did not achieve satisfactory %
performance with the raw HoG feature. %
On the contrary, the output \trc{} representation (in the second row) exhibits a clear union-of-orthogonal-subspaces structure, making it easier to segment different motions.
The clear contrast in PCA visualization reveals that the union-of-orthogonal-subspaces distribution of features is a key factor contributing to the state-of-the-art performance of \trc.

\begin{figure}[tbp]
    \centering
    \resizebox{0.5\textwidth}{!}{%
        \begin{tabular}{ccc}
    \includegraphics[trim=90pt 100pt 90pt 100pt, clip,width=0.15\textwidth]{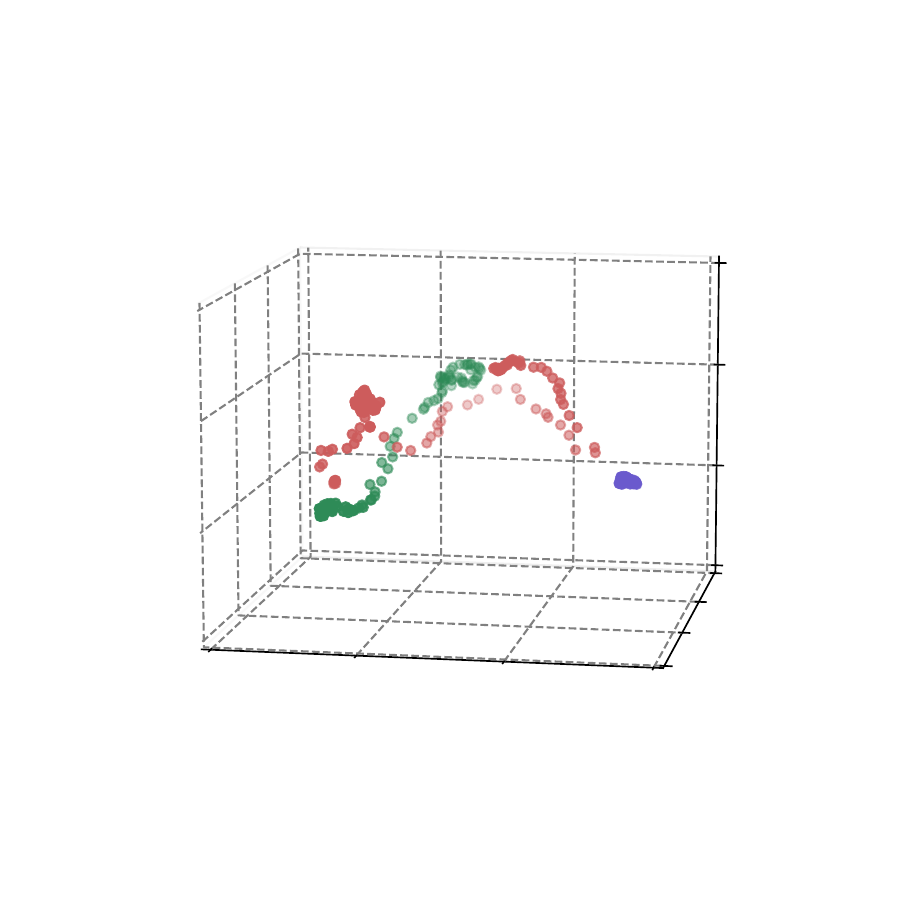}&
    \includegraphics[trim=90pt 100pt 90pt 100pt, clip,width=0.15\textwidth]{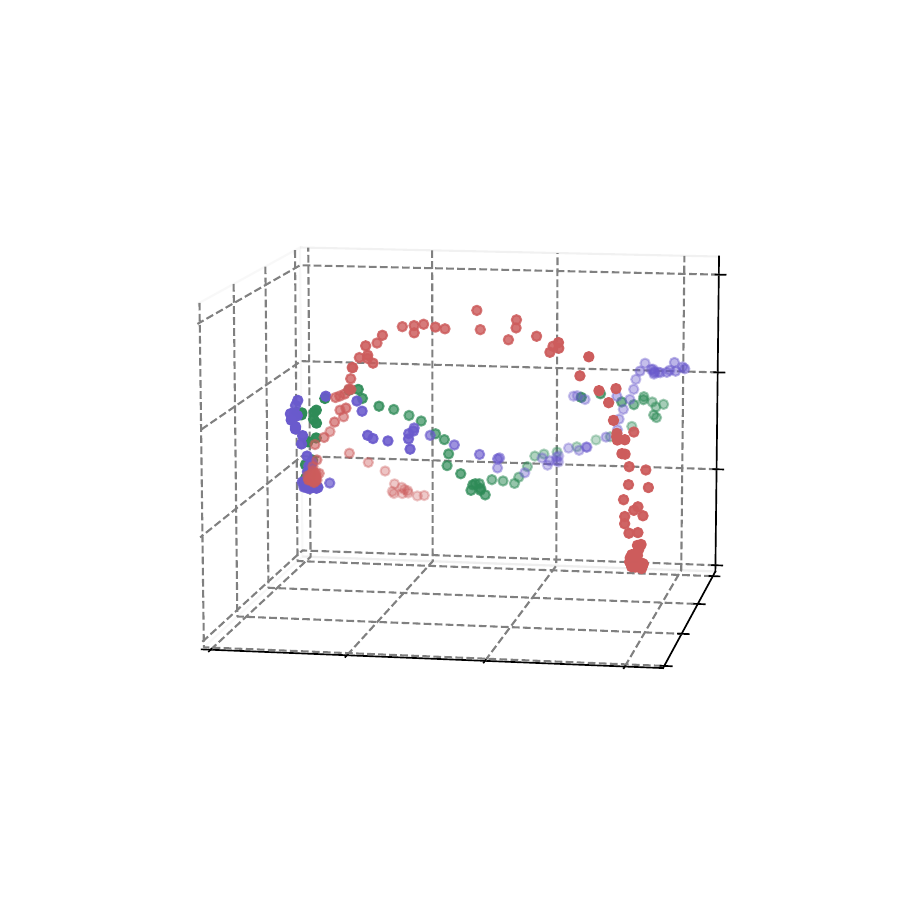}&
    \includegraphics[trim=90pt 100pt 90pt 100pt, clip,width=0.15\textwidth]{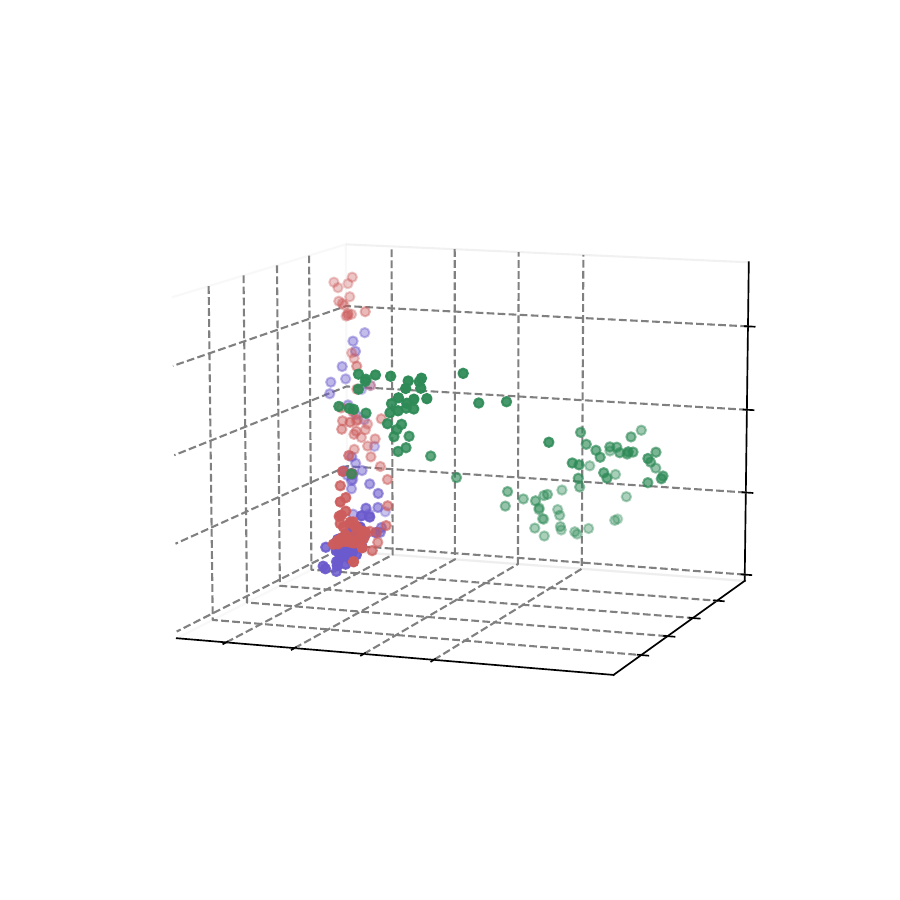}\\
   \subcaptionbox{Keck}{
            \includegraphics[trim=90pt 100pt 90pt 100pt, clip,width=0.15\textwidth]{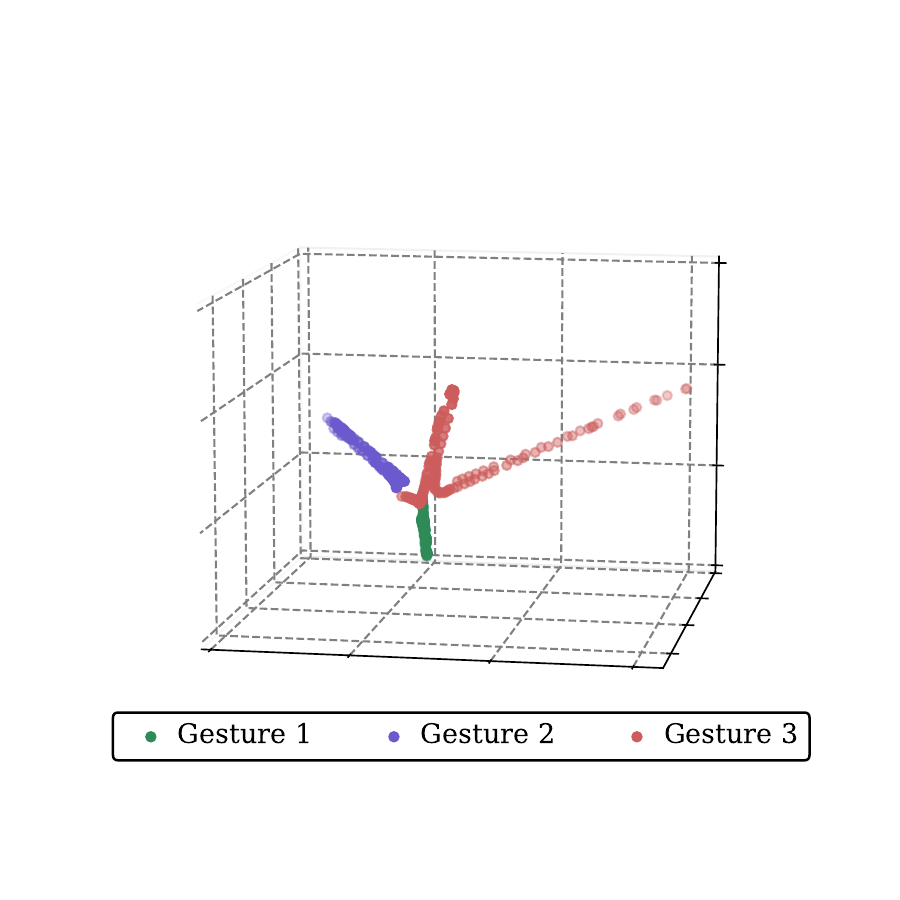}}&
    \subcaptionbox{UT}{
            \includegraphics[trim=90pt 100pt 90pt 100pt, clip,width=0.15\textwidth]{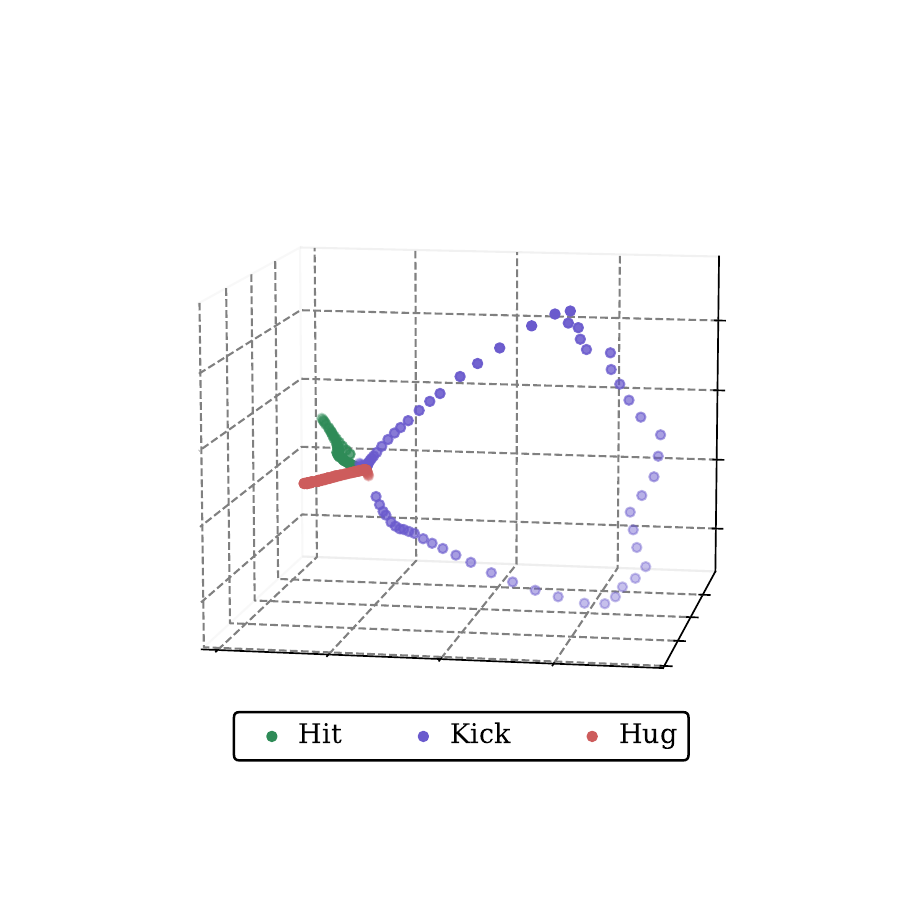}}&
    \subcaptionbox{MAD}{
           \includegraphics[trim=90pt 100pt 90pt 100pt, clip,width=0.15\textwidth]{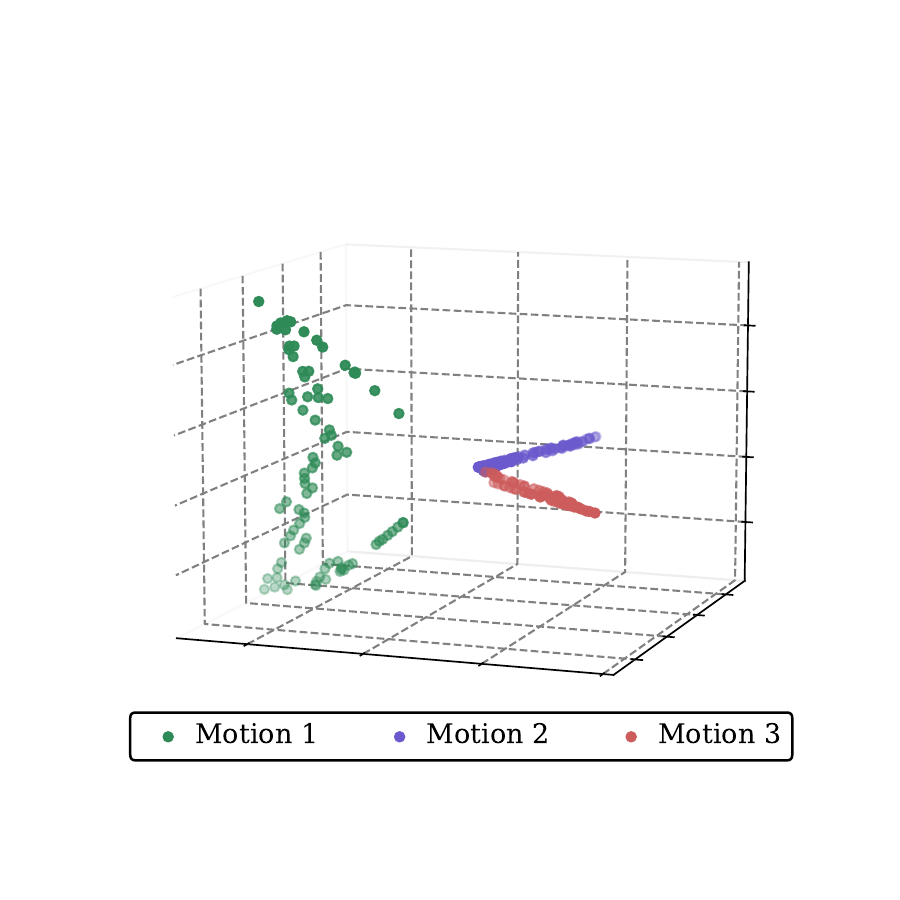}}
    \end{tabular}
    }
\caption{\textbf{Visualization of features via PCA.} First row: input HoG features. Second row: learned \trc{} representations. %
Experiments are conducted on the first sequence of each dataset.}
\label{fig:PCA}
\vskip -0.15in
\end{figure}

\begin{figure}[htb]
    \centering
    \begin{subfigure}[b]{0.5\linewidth}
           \includegraphics[width=\textwidth]{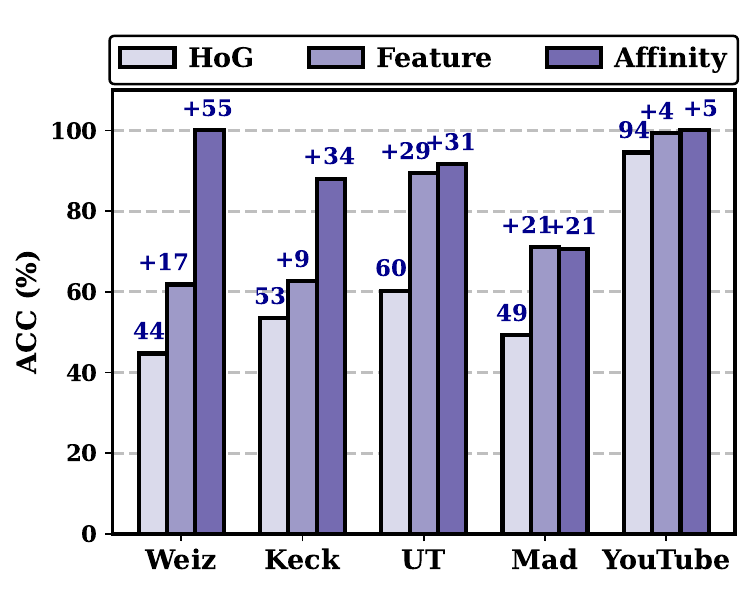}
            \caption{SC}
            \label{fig:feature_acc-sc}
    \end{subfigure}\hfill
    \begin{subfigure}[b]{0.5\linewidth}
           \includegraphics[width=\textwidth]{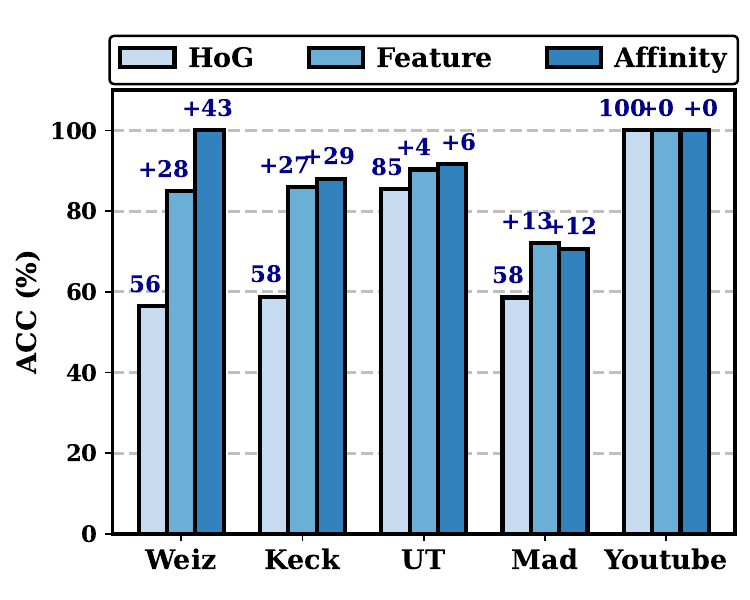}
            \caption{EnSC}
            \label{fig:feature_acc-ensc}
    \end{subfigure}
\caption{\textbf{Clustering accuracy of using HoG features, learned features, and learned affinity.} 
Experiments are conducted on the first sequence of each dataset.}  
\label{fig:feature_acc}
  \vskip -0.1in
\end{figure}

\myparagraph{Quantitative evaluation of representations}
To quantitatively evaluate the effectiveness of the learned representation of \trc, we illustrate the clustering performance of using HoG feature and the learned representation in Figure~\ref{fig:feature_acc}.
We perform %
spectral clustering (SC)~\cite{Shi:TPAMI00} and Elastic Net Subspace Clustering (EnSC)~\cite{You:CVPR16-EnSC} algorithms, and report the clustering accuracy in Figures~\ref{fig:feature_acc-sc} and~\ref{fig:feature_acc-ensc}, respectively.\footnote{For EnSC, we tune the 
hyper-parameter %
$\gamma\in\{1,2,5,10,20,50,100,200,400,800,1600,3200\}$ and the %
hyper-parameter $\tau\in\{0.9,0.95,1\}$ and report the best clustering result.}
 
Comparing to the HoG features, the clustering accuracy of using the learned representations improves significantly across all datasets and clustering algorithms, with particularly notable gains on the UT dataset using spectral clustering (a 29\% improvement) and the Weizmann dataset using EnSC (a 28\% improvement).
Since the performance of clustering algorithms heavily depends on the underlying data distribution, these significant improvements highlight the enhanced quality of the learned representations.
Assisted with these features, the clustering accuracy of the affinity matrix $\boldGamma$ is further improved, suggesting that the clustering head in \trc{} is more effective in revealing clusters than other classical clustering approaches.
The superior performance of the clustering head may be attributed to the fact that the optimal $\boldGamma$ in Problem (\ref{Eq:TR2C}) is better at capturing the subspace membership of the learned representations.

\begin{figure}[htb]
    \centering
    \begin{subfigure}[b]{0.5\linewidth}
           \includegraphics[trim=0pt 0pt 25pt 5pt, clip, width=\textwidth]{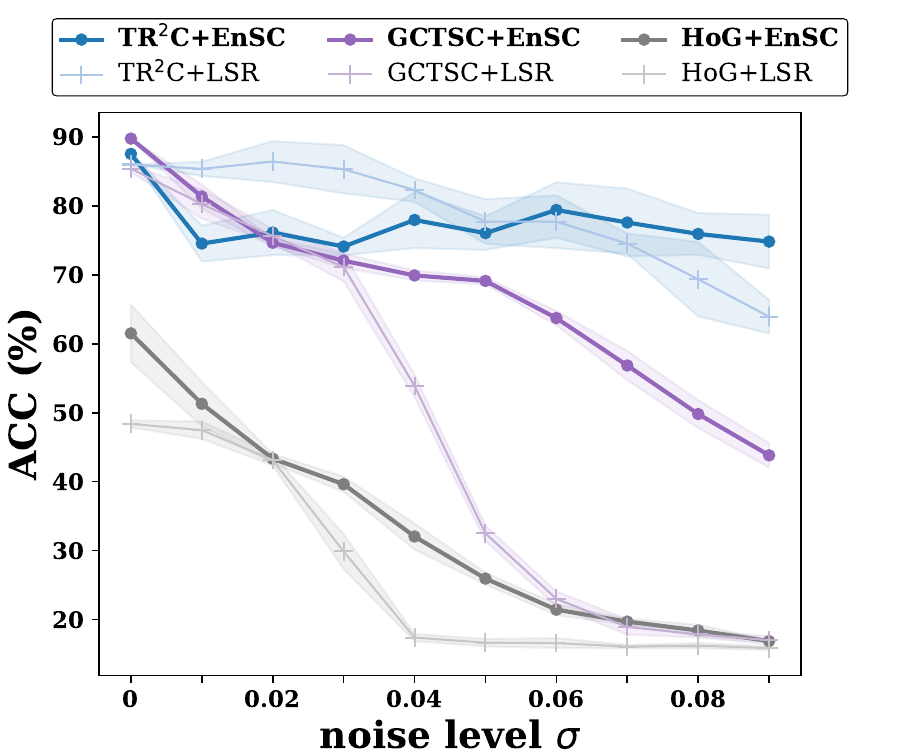}
            \caption{Weiz}
            \label{fig:noise-weiz}
    \end{subfigure}\hfill
    \begin{subfigure}[b]{0.5\linewidth}
           \includegraphics[trim=0pt 0pt 25pt 5pt, clip, width=\textwidth]{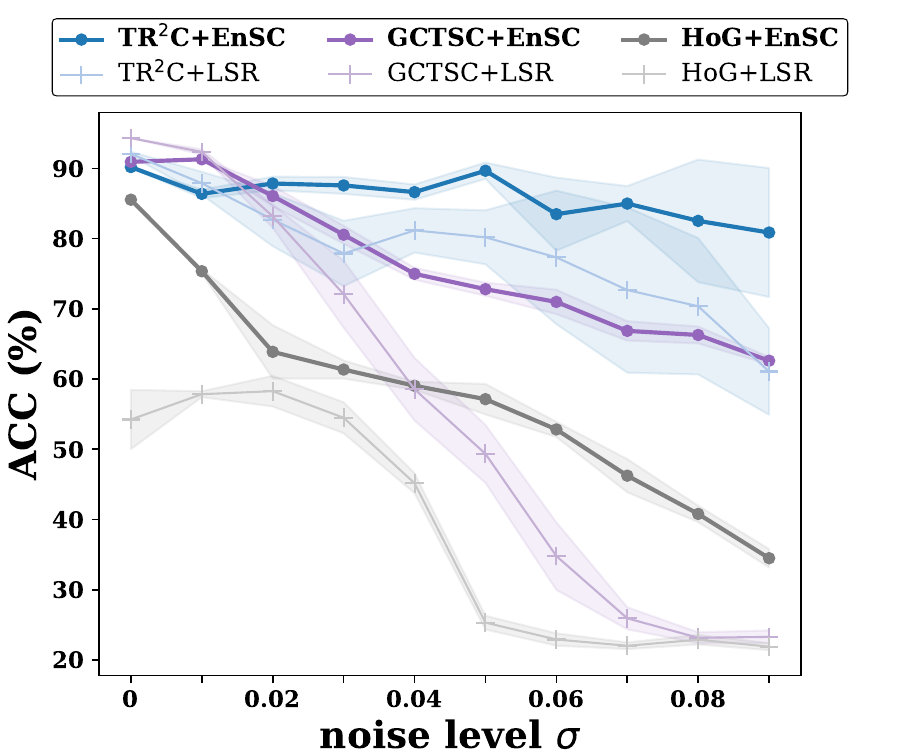}
            \caption{UT}
            \label{fig:noise-ut}
    \end{subfigure}
\caption{\textbf{Clustering accuracy of features under noise corruption.} We test on features learned by \trc{}, GCTSC, and HoG features, using EnSC (thick) and LSC (thin) for clustering.}  
\label{fig:noise}
\end{figure}

\myparagraph{Robustness evaluation of representations}
Intuitively, if the representations align with a UoS structure, they will enjoy strong robustness to the random noise corruption. 
To verify this, we corrupt the learned representations of \trc{}, GCTSC~\cite{Dimiccoli:ICCV21-GCTSC} and HoG features by the additive isotropic Gaussian noise $\mathcal{N}(\boldsymbol{0},\sigma\boldsymbol{I})$, where $\sigma>0$ is the noise level.
We apply EnSC~\cite{You:CVPR16-EnSC} and LSR~\cite{Lu:ECCV12} to cluster the corrupted features and plot the clustering accuracy along with the standard deviation after running with 5 different random seeds.
As shown in Figure~\ref{fig:noise}, although the clustering performance of GCTSC is highly competitive without noise ($\sigma=0$), it decreases significantly regardless of the clustering algorithms on both datasets.
In contrast, the representations of \trc{} is more robust to the noise corruption.
When clustering with EnSC (thick line), the average clustering accuracy of \trc{}'s representations drops at most 15\% and 10\% on Weiz and UT dataset, respectively, which are significantly less than the 45\% and 30\% drop of GCTSC.

\myparagraph{Ablation study}
To study the effect of each component in the loss function, we conduct ablation study and report the results in Table~\ref{tab:ablation} (see Appendix for more results).
Clearly, we can read that both $\mathcal{L}_\rho$ and $\mathcal{L}_{\bar\rho^c}$ are indispensable for our \trc{} to learn structured representations %
to facilitate HMS. 
As illustrated, the absence of either $\mathcal{L}_\rho$ or $\mathcal{L}_{\bar\rho^c}$ will seriously detriment the performance, resulting in over-segmentation (line 3) or over-compactness (line 2) of representations.
Additionally, the temporal regularizer $\mathcal{L}_r$ also contributes significantly to the clustering performance (line 1), which validates that temporal consistency of representations is an indispensable prior for HMS problem.

\begin{table}[htbp]
  \centering
  \caption{\textbf{Ablation study.} We report the average performance of all the sequences in the Weiz, Keck and UT dataset.}
  \resizebox{\linewidth}{!}{
    \begin{tabular}{P{0.6cm}P{0.6cm}P{0.6cm}|P{0.9cm}P{0.9cm}P{0.9cm}P{0.9cm}P{0.9cm}P{0.9cm}}
    \toprule
    \rowcolor{myGray}\multicolumn{3}{c|}{Loss} & \multicolumn{2}{c}{Weiz} & \multicolumn{2}{c}{Keck} & \multicolumn{2}{c}{UT}\\
    \rowcolor{myGray}$\mathcal{L}_{\rho}$& $\mathcal{L}_{\bar\rho^c}$& $\mathcal{L}_{r}$ & ACC   & NMI & ACC   & NMI & ACC   & NMI \\
    \midrule
    $\checkmark$& $\checkmark$&   & 37.30  & 45.86   & 47.29 & 49.78    & 45.79 & 35.30 \\
          & $\checkmark$& $\checkmark$& 53.14  & 61.51 & 47.91 & 51.39 & 63.13 & 59.51 \\
    $\checkmark$&       & $\checkmark$& 64.68  & 74.67& 58.60 & 65.21 &       65.67&  66.09\\
    $\checkmark$ &     &  & 41.21  & 44.57 & 44.01 & 41.46   &  46.80 & 37.49\\
         &  $\checkmark$ &  & 56.03  & 64.19  & 47.50 & 52.11    &  76.39 & 72.41\\
       &      & $\checkmark$ & 52.59  & 60.33& 48.35 & 50.87 & 62.13 & 58.29\\
        $\checkmark$& $\checkmark$& $\checkmark$& \textbf{94.07}  & \textbf{96.08}& \textbf{86.78} & \textbf{86.93}& \textbf{94.05} & \textbf{92.34}\\ 
    \bottomrule
    \end{tabular}%
    }
  \label{tab:ablation}%
\end{table}%

\begin{figure*}[ht]
\centering
   \begin{subfigure}[b]{0.24\linewidth}
           \includegraphics[trim=30pt 55pt 25pt 60pt, clip,width=\textwidth]{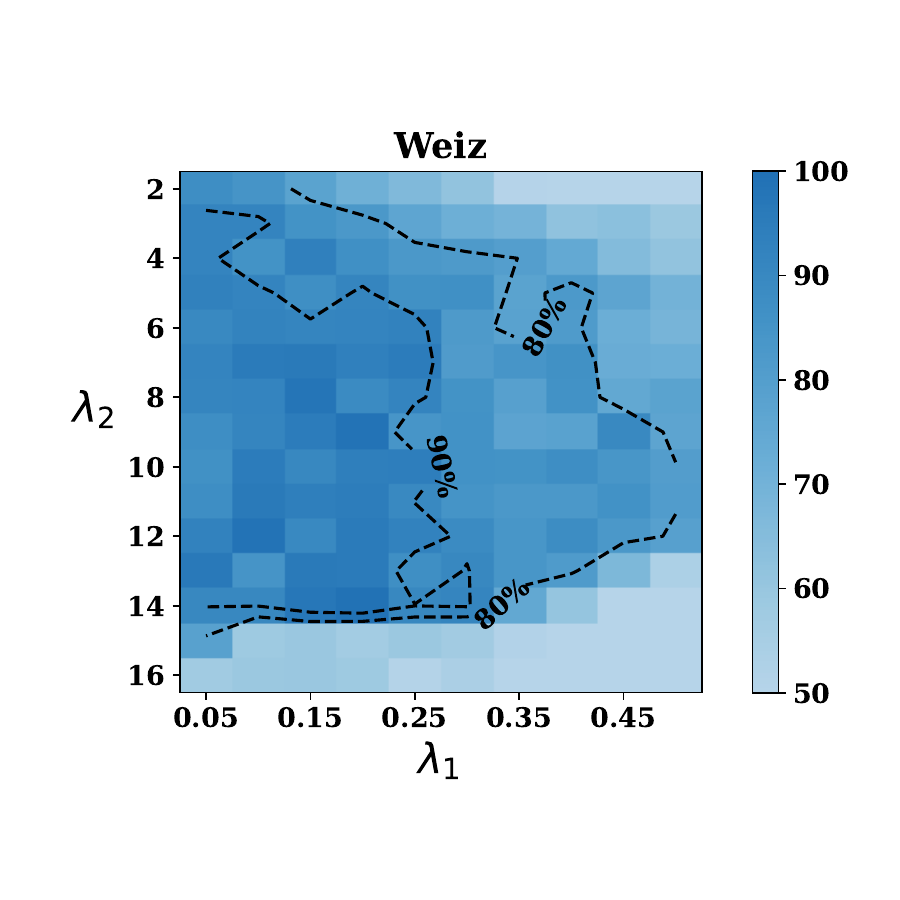}
    \end{subfigure}\hfill
    \begin{subfigure}[b]{0.24\linewidth}
           \includegraphics[trim=30pt 55pt 25pt 60pt, clip,width=\textwidth]{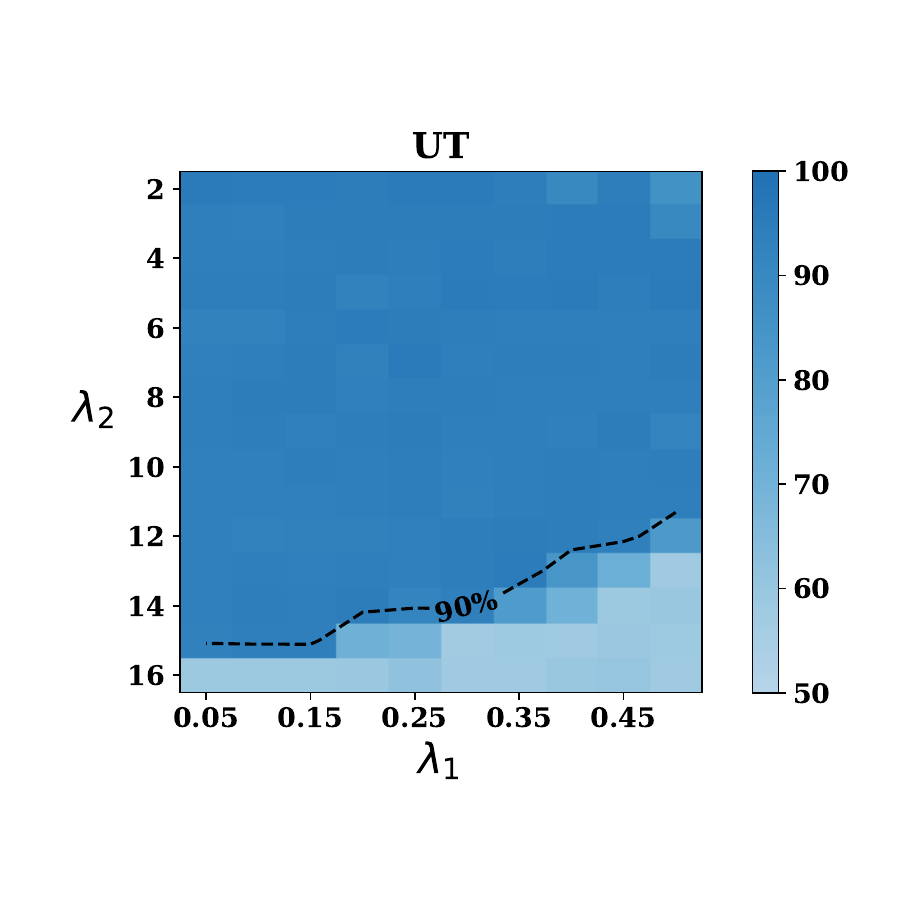}
    \end{subfigure}\hfill
    \begin{subfigure}[b]{0.24\linewidth}
           \includegraphics[trim=30pt 55pt 25pt 60pt, clip,width=\textwidth]{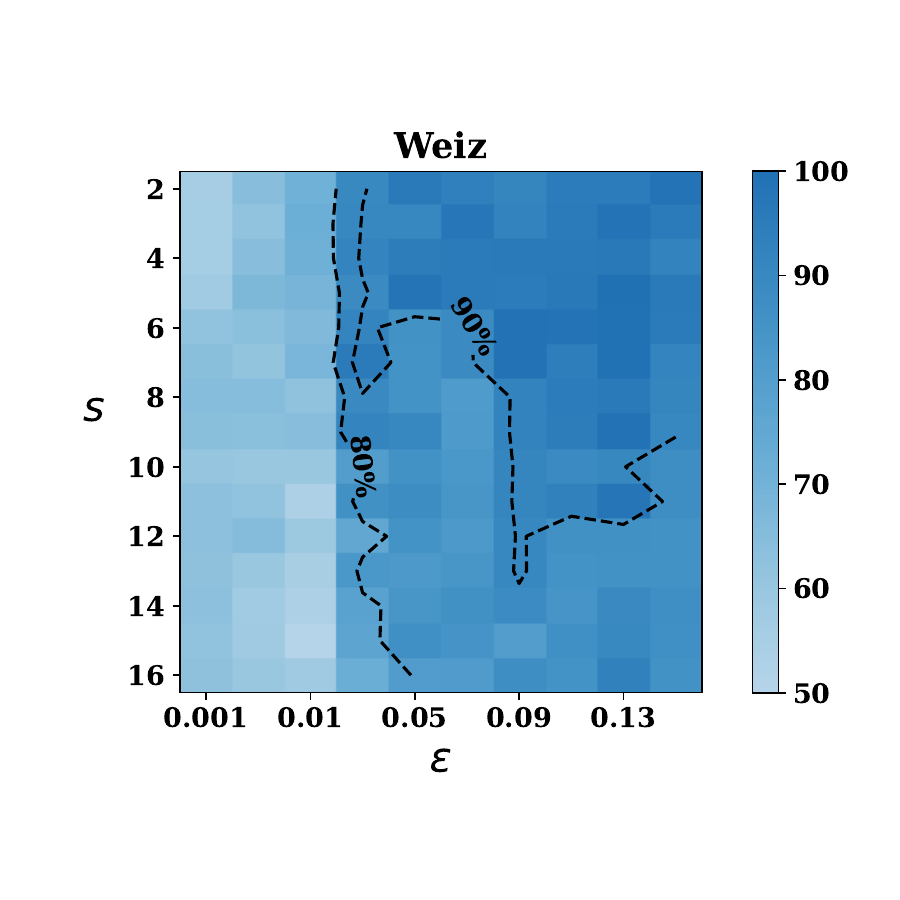}
    \end{subfigure}\hfill
    \begin{subfigure}[b]{0.24\linewidth}
           \includegraphics[trim=30pt 55pt 25pt 60pt, clip,width=\textwidth]{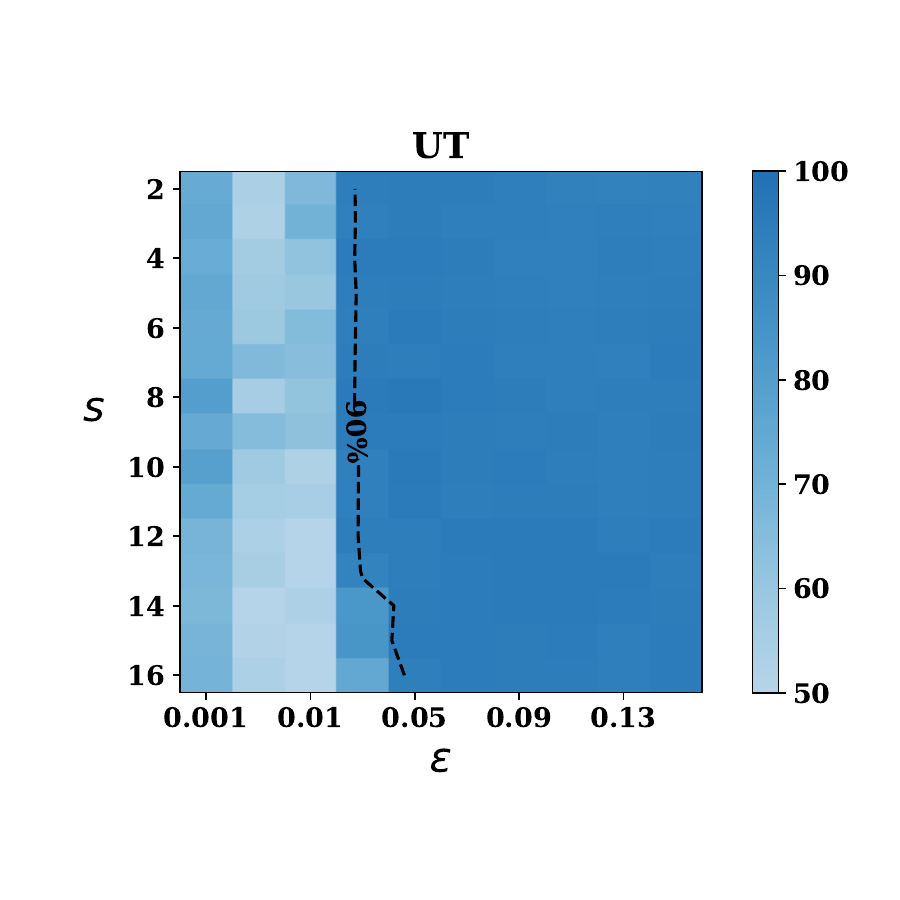}
    \end{subfigure}
\caption{\textbf{Sensitivity to hyper-parameters.} The sensitivity of \trc{} with respect to $\lambda_1$, $\lambda_2$, $s$ and $\epsilon$ is studied through experiments on the first sequence of the Weiz and UT dataset with three different random seeds.}
\label{fig:sensitivity}
\vskip -0.15in
\end{figure*}

\myparagraph{Sensitivity to hyper-parameters}
We study the sensitivity of our model with respect to the %
parameters $\lambda_1$ and %
$\lambda_2$, 
the window size $s$ and the coding precision $\epsilon$.
As shown in Figure~\ref{fig:sensitivity}, $\lambda_1$ is recommended to be smaller than $0.35$, as an over-large $\lambda_1$ %
might lead to over-compactness of the learned representations.
In contrast, $\lambda_2$ and $s$ can be selected from a wide range while maintaining optimal performance. 
The coding precision $\epsilon$ determines the level of distortion in data compression\footnote{Please kindly refer to \cite{Ma:PAMI07} for more details.}, which is recommended to be larger than $0.1$.
In general, our \trc{} framework is insensitive to these %
hyper-parameters.

\myparagraph{Time cost comparison}
To test the time consuming of \trc{}, we report the training time of the first sequence across different benchmarks and computing devices.\footnote{Since spectral clustering is adopted by all baselines for segmentation, our evaluation specifically focuses on the training time cost.}
We defer the complexity analysis of \trc{} to Appendix~\ref{sec:Complexity_Analysis}.
For TSC and GCTSC along with its GPU implementation, we use the code provided by~\cite{Dimiccoli:ICCV21-GCTSC} and train for $T=15$ and $T=100$ iterations, respectively.
For \trc{}, we train for $T=500$ iterations.
We conduct all the experiments with a single NVIDIA RTX 3090 GPU and Intel Xeon Platinum 8255C CPU.
As shown in Table~\ref{tab:Time cost}, %
the time cost of \trc{} is comparable to TSC, but significantly less than GCTSC, which is quite time-consuming. 
Since the parallel computing empowered by GPU speeds up particularly for large $N$, \trc{}$+$GPU outperforms TSC in terms of speed with a greater performance margin on the YouTube dataset.

\begin{table}[htbp]
    \centering
\caption{\textbf{Total training time (s) comparison.} The %
best time cost is marked in \textbf{bold} and the second best result %
is %
\underline{underlined}.}
\label{tab:Time cost}
\resizebox{\linewidth}{!}{
    \begin{tabular}{l|c|ccccc}
    \toprule
     \rowcolor{myGray} & $T$ &  Weiz&  Keck&  UT& MAD & YouTube\\
     \midrule
         TSC& 15 & \underline{20.0} & \underline{20.4} & \textbf{5.6} & \textbf{9.2} & \underline{116.5}\\
         GCTSC& 100 & 1551.7 & 1554.1 & 415.4 & 810.3 & 12677.8\\
 GCTSC$+$GPU& 100 & 1122.2 &1142.4 & 374.8 & 622.4 & 8474.7\\
         \trc& 500 & 91.0 & 228.2 & 82.2 & 138.4 & 929.1\\
         \trc$+$GPU& 500 &\textbf{10.7} & \textbf{16.9} & \underline{10.4} & \underline{12.8} & \textbf{41.0}\\
     \bottomrule
    \end{tabular}
    }
\end{table}

\myparagraph{\trc{} based on CLIP}
The performance of \trc{} based on CLIP pretrained features (denoted as ``\trc+CLIP'') comparing to state-of-the-art approaches is shown in Table~\ref{tab:CLIP-performance}.
As illustrated, the clustering accuracy of \trc$+$CLIP surpasses that of \trc{} on the Weizmann, Keck and YouTube dataset, with improvements of 2\%, 7\% and 1\%, respectively.  
This performance enhancement is attributed to the superior representation capability of the CLIP pretrained model.
Specifically, the pretrained CLIP image encoder captures more 
high-level semantic information from each frame, which is crucial for distinguishing different human motions. In contrast, HoG features primarily capture low-level information from each frame.

Besides, we also conduct experiments on the zero-shot learning of CLIP (please refer to Appendix~\ref{sec:zero-shot}), which demonstrates that vanilla zero-shot classification is not suitable for HMS; whereas \trc$+$CLIP succeeds by learning temporally consistent representations that align with a union of orthogonal subspaces.

\begin{table}[tbp]
  \centering
  \caption{\textbf{The performance of \trc{} based on CLIP features comparing to state-of-the-art algorithms.} 
  We denote ``$\boldsymbol{A}+$CLIP'' as the algorithm $\boldsymbol{A}$ based on CLIP features.
  }
  \resizebox{0.95\linewidth}{!}{
    \begin{tabular}{lcccccc}
    \toprule
    \rowcolor{myGray} & \multicolumn{2}{c}{Weiz} & \multicolumn{2}{c}{Keck} & \multicolumn{2}{c}{YouTube}\\
\rowcolor{myGray} \multirow{-2}{*}{Method} & ACC   & NMI   & ACC   & NMI & ACC   & NMI \\
\midrule
    TSC   & 61.11  & 81.99  & 47.81  & 71.29 & 90.40 & 95.00 \\
 TSC$+$CLIP& 89.61& 93.35& 78.81&83.81 & 93.86 & 94.91\\
    GCTSC & 85.01  & 90.53  & 78.64  & 83.25 & 95.79 & 96.30 \\
 GCTSC$+$CLIP& 89.39& 89.90& 83.31&84.55 & 96.64 & 97.41 \\
    \trc  & 94.12 & 95.91 &  83.50 & 85.63 & 97.96 & 98.96\\
    \trc$+$CLIP & \textbf{96.21} & \textbf{97.12} & \textbf{90.10} & \textbf{89.63} & \textbf{99.35} & \textbf{99.48}\\
    \bottomrule
    \end{tabular}%
    }
  \label{tab:CLIP-performance}%
  \vskip -0.17in
\end{table}%

\myparagraph{Comparison to Temporal Action Segmentation}
Temporal Action Segmentation (TAS) is an unsupervised task closely related to HMS, as both aim to partition videos into non-overlapping segments~\cite{Ding:TPAMI23}.
The key difference lies in the scale and the nature of the actions: HMS typically involves \emph{macro-scale} motions (\eg, running, jumping) characterized by global and easily distinguishable movements, whereas TAS focuses on \emph{micro-scale} manipulative actions (\eg, grasping a cup, pouring milk), which are more subtle and fine-grained.
We conduct experiments on %
three benchmark dataset on TAS, including the Breakfast dataset~\cite{Kuehne:CVPR14-Breakfast}, 
the YouTube Instructional dataset~\cite{Alayrac:CVPR16-Inria}, and the 50 Salads dataset~\cite{Stein:UC13-Salad}.
For the feature extractor selection of each dataset, we follow the baselines.
Experimental details are described in Appendix~\ref{sec:action seg}.
As can be seen in Table~\ref{tab:action seg}, \trc{} performs comparable with the state-of-the-art TAS algorithms on the three benchmark datasets for TAS. %

\begin{table}[htbp]
  \centering
  \caption{\textbf{The action segmentation performance of \trc{}.} Other results %
  are directly cited from their %
  papers.}
  \vskip -0.1 in
  \label{tab:action seg}%
  \resizebox{\linewidth}{!}{
    \begin{tabular}{lccccccccc}
    \toprule
      \rowcolor{myGray} & \multicolumn{3}{c}{Breakfast} & \multicolumn{3}{c}{YouTube Instr.} & \multicolumn{3}{c}{50 Salads} \\
    \rowcolor{myGray} \multirow{-2}{*}{Method} & MoF   & F1    & mIoU  & MoF   & F1    & mIoU  & MoF   & F1    & mIoU \\
    \midrule
    LSTM+AL~\cite{Aakur:CVPR19} & 42.9  & -     & -     & -     & 39.7  & -     & -     & -     & - \\
    TWF~\cite{Sarfraz:CVPR21}   & 62.7  & 49.8  & \textbf{42.3}  & 56.7  & 48.2  & -     & \underline{66.8}  & \underline{56.4}  & \textbf{48.7} \\
    ABD~\cite{Du:CVPR22}   & \textbf{64.0}    & 52.3  & -     & 67.2  & 49.2  & -     & \textbf{71.8}  & -     & - \\
CoSeg~\cite{Wang:TNNLS23} & 53.1  & \textbf{54.7}  & -     & 47.9  & -     & \underline{53.7}  & -     & -     & - \\
    ASOT~\cite{Xu:CVPR24}  & \underline{63.3}  & \underline{53.5}  & 35.9  & \underline{71.2}  & \textbf{63.3}  & 47.8  & 64.3  & 51.1  & 33.4 \\
    Our \trc  & 59.9  & 47.1  & \underline{39.9}  & \textbf{71.8}  & \underline{59.1}  & \textbf{57.8}  & 65.8  & \textbf{58.4}  & \underline{48.2} \\
    \bottomrule
    \end{tabular}%
    }
\vskip -0.1in
\end{table}%

Conversely, we also evaluate the state-of-the-art TAS algorithms on Weizmann and Keck datasets with HoG, CLIP and DINOv2 \cite{oquab:TMLR24-dinov2} features.
As shown in Table~\ref{tab:TAS_sota_on_HMS}, our \trc{} achieves the best performance across different datasets and different features.
Furthermore, we argue that the performance of our \trc{} faithfully reflects the quality of the input features, \ie, DINOv2$ > $CLIP$ > $HoG.
\vskip -0.1 in

\begin{table}[htbp]
  \centering
  \caption{\textbf{Evaluating state-of-the-art TAS algorithms} on Weiz and Keck datasets with HoG, CLIP and DINOv2 features.}
  \resizebox{0.9\linewidth}{!}{
    \begin{tabular}{cccccccc}
    \toprule
    \rowcolor{myGray} &       & \multicolumn{2}{c}{Weiz} & \multicolumn{2}{c}{Keck} & \multicolumn{2}{c}{AVG} \\
    \rowcolor{myGray} \multirow{-2}{*}{Feature} &  \multirow{-2}{*}{Methods} & ACC   & NMI   & ACC   & NMI   & ACC   & NMI \\
    \midrule
    \multirow{5}[2]{*}{HoG} & TWF~\cite{Sarfraz:CVPR21}   & 66.1  & 84.3  & 48.8  & 63.7  & 57.5  & 74.0  \\
          & ASOT~\cite{Xu:CVPR24}  & 68.0  & 77.6  & 66.4  & 76.0  & 67.2  & 76.8  \\
          & HVQ~\cite{Spurio:AAAI25} & 66.7  & 61.5  & 63.5  & 75.2  &    65.1  & 68.4  \\
          & GCTSC~\cite{Dimiccoli:ICCV21-GCTSC} & 85.0  & 90.5  & 78.6  & 83.3  & 81.8  & 86.9  \\
          & Our \trc{}  & \textbf{94.1}  & \textbf{95.9}  & \textbf{83.5}  & \textbf{85.6}  & \textbf{88.8}  & \textbf{90.8}  \\
    \midrule
    \multirow{5}{*}{CLIP} & TWF~\cite{Sarfraz:CVPR21}   & 76.8  & 89.3  & 70.4  & 79.4  & 73.6  & 84.4  \\
    \multirow{5}{*}{(ViT-L/14)} & ASOT~\cite{Xu:CVPR24}  & 71.1  & 79.4  & 67.0  & 76.4  & 69.1  & 77.9  \\
    & HVQ~\cite{Spurio:AAAI25} & 72.6  & 85.2  & 73.5  & 78.5    & 73.1  & 81.9  \\
          & GCTSC~\cite{Dimiccoli:ICCV21-GCTSC} & 89.4  & 89.9  & 83.3  & 84.6  & 86.4  & 87.3  \\
          & Our \trc{}  & \textbf{96.2}  & \textbf{97.1}  & \textbf{90.1}  & \textbf{89.6}  & \textbf{93.2}  & \textbf{93.4}  \\
    \midrule
    \multirow{5}{*}{DINOv2} & TWF~\cite{Sarfraz:CVPR21}   & 66.8  & 83.1  & 65.2  & 70.8  & 66.0  & 77.0  \\
    \multirow{5}{*}{(ViT-L/14)} & ASOT~\cite{Xu:CVPR24}  & 71.4  & 80.9  & 60.1  & 74.4  & 65.8  & 77.7  \\
    & HVQ~\cite{Spurio:AAAI25} & 73.0  & 85.5  & 67.2  & 77.1   & 70.1  & 81.3  \\
          & GCTSC~\cite{Dimiccoli:ICCV21-GCTSC} & 90.8  & 91.8  & 82.8  & 84.8  & 86.8  & 88.3  \\
          & Our \trc{}  & \textbf{98.6}  & \textbf{98.5}  & \textbf{90.2}  & \textbf{89.7}  & \textbf{94.4}  & \textbf{94.1}  \\
    \bottomrule
    \end{tabular}}
  \label{tab:TAS_sota_on_HMS}%
\end{table}%

\section{Conclusion}
\label{Sec:Conclusion}
We have presented a novel framework for the HMS task, called \trc, which jointly learns structured  representations and affinity to segment the frame sequences in video.
Specifically, the structured representations learned by \trc{} maintain temporally consistency and align well with a UoS structure, which is favorable for the HMS task.
We note that our \trc{} is an effective and efficient %
deep subspace clustering framework %
for HMS, where %
$\rho^c(\boldsymbol{Z},\epsilon~|~\boldPi)$ is for subspace detection, $r(\boldsymbol{Z})$ is a temporal continuity prior, and $-\rho(\boldsymbol{Z},\epsilon)$ is to prevent representation collapse. As future work, it would be attempting to explore more sophisticated %
implementation %
by customizing different components.

\section*{Acknowledgments}
The authors would like to thank the constructive comments from anonymous reviewers. 
This work is supported by the National Natural Science Foundation of China under Grants~61876022 and 62076031. 

{
\small
\bibliographystyle{ieeenat_fullname}
\bibliography{./biblio/xhmeng,./biblio/yc,./biblio/cgli,./biblio/zhjj,./biblio/learning,./biblio/temp,./biblio/temporal}
}

\onecolumn
\setcounter{page}{1}
\setcounter{section}{0}
\mymaketitlesupplementary
\appendix
\renewcommand{\thefigure}{\Alph{section}.\arabic{figure}}
\renewcommand{\thetable}{\Alph{section}.\arabic{table}}
\setcounter{figure}{0}
\setcounter{table}{0}
\renewcommand{\appendixname}{Appendix~\Alph{section}}

\section{Experimental Supplementary Material}
\subsection{Datasets Description}
\label{Sec:Datasets}

\myparagraph{Weizmann action dataset (Weiz)}
The Weizmann dataset~\cite{Gorelick:TPAMI07-Weiz} contains 90 motion sequences, with 9 individuals each completing 10 motions, \eg, running, jumping, skipping, waving and bending. 
The resolution of video is $180\times 144$ pixels with 50 FPS.\\
\myparagraph{Keck gesture dataset (Keck)}
The Keck dataset~\cite{Jiang:TPAMI12-Keck} contains 56 action sequences, with 4 individuals each performing 14 motions derived from military hand signals, \eg, turning left, turning right, starting, and speeding up.
The resolution of video is $640\times 480$ pixels with 15 FPS.\\
\myparagraph{UT interaction dataset (UT)}
The UT dataset~\cite{Ryoo:ICCV09-Ut} contains 10 video sequences, each of which consists of 2 people completing 6 different motions, \eg, shaking hands, hugging, pointing, and kicking.
The resolution of video is $720\times 480$ pixels with 30 FPS.\\
\myparagraph{Multi-model Action Detection dataset (MAD)}
The MAD dataset~\cite{Huang:ECCV14-MAD} contains 40 video sequences (20 people, 2 videos each) with 35 motions in each video.
The resolution of video is $320\times 240$ pixels with 30 FPS.
The dataset gives both depth data and skeleton data.\\
\myparagraph{UCF-11 YouTube action dataset (YouTube)}
The YouTube dataset \cite{Liu:CVPR09-youtube} contains 1168 video sequences with 11 motions, \eg, biking, diving, and golf swinging.
The resolution of video is $320\times 240$ pixels with 30 FPS.
Specifically, the human motions in the YouTube dataset are partially associated with objects such as horses, bikes, or dogs.

To have a fair comparison with the baselines, we cut down the number of human motions of Keck, MAD and YouTube datasets to $10$.
For Keck, Weiz and YouTube datasets in which each video captures only one human motion, we concatenate the original videos and conduct experiments on the resulting videos.

\subsection{List of Hyper-Parameters}
\label{sec:hyperpara}
The hyper-parameters of training \trc{} are summarized in Table~\ref{tab:hyperpara}.
We choose the same hidden dimension $d_{pre}$, output dimension $d$, window size $s$, coding precision $\epsilon$, and learning rate $\eta$ for all the experiments and tune the weights $\lambda_1$ and $\lambda_2$ for each dataset.
For training on CLIP features, we decrease the number of training iterations from $500$ to $100$ due to the faster convergence, while keeping all the other hyper-parameters unchanged. 

\begin{table*}[h]
\caption{\textbf{Detailed hyper-parameters configuration for training \trc{} with different feature extractors.}}
\label{tab:hyperpara}
\begin{center}
\resizebox{0.7\linewidth}{!}{
    \begin{tabular}{P{1.5cm}P{1.5cm}P{0.9cm}P{0.9cm}P{0.9cm}P{0.9cm}P{0.9cm}P{0.9cm}P{0.9cm}P{1.5cm}}
    \toprule
         \rowcolor{myGray} Features & Dataset &  $d_{pre}$&  $d$&  $T$& $\lambda_1$& $\lambda_2$&  $s$& $\epsilon$&$\eta$\\
         \midrule
         \multirow{4}{*}{HoG} &Weiz&  512&  64&  500& 0.1& 12&  2& 0.1&$5\times 10^{-3}$\\
         
         &Keck&  512&  64&  500& 0.1& 10&2& 0.1&$5\times 10^{-3}$\\
         
         &UT&  512&  64&  500& 0.1& 10&
         2&0.1&$5\times 10^{-3}$\\
         
         &MAD&  512&  64&  500& 0.15& 15&
         2& 0.1& $5\times 10^{-3}$\\
     \midrule
         \multirow{1}{*}{VGG}&YouTube & 512 & 64 & 500 & 1 & 2 & 2 & 0.1 & $5\times 10^{-3}$\\
    \midrule
         \multirow{3}{*}{CLIP}&Weiz&  512&  64&  100& 0.1& 12
&
         2& 0.1&$5\times 10^{-3}$\\
         &Keck&  512&  64&  100& 0.1& 10
&  2& 0.1&$5\times 10^{-3}$\\
 &YouTube& 512 & 64 & 100 & 1 & 2 & 2 & 0.1 & $5\times 10^{-3}$\\
 \bottomrule
    \end{tabular}
    }
\end{center}
\end{table*}

\subsection{Visualization of Representations by GCTSC and \trc{}}
\label{sec:supp PCA visualization}
\begin{figure*}[ht]
    \centering
    \begin{subfigure}[b]{0.18\linewidth}
           \includegraphics[trim=90pt 100pt 90pt 100pt, clip,width=\textwidth]{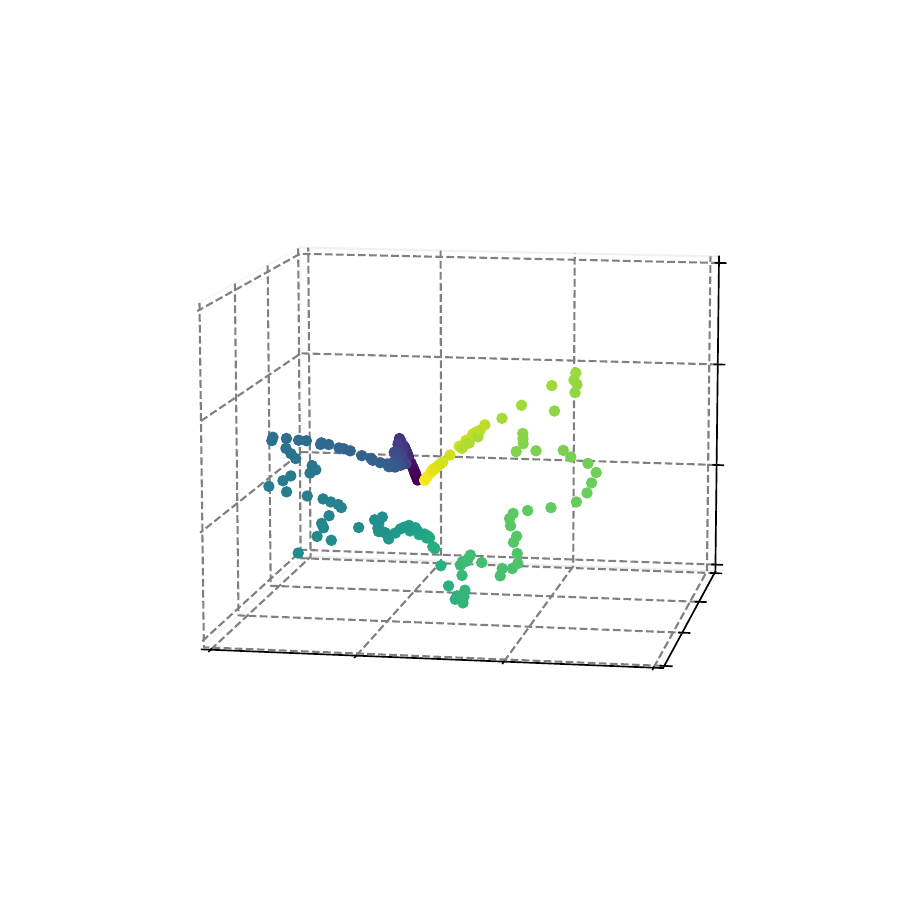}
    \end{subfigure}\hfill
    \begin{subfigure}[b]{0.18\linewidth}
            \includegraphics[trim=90pt 100pt 90pt 100pt, clip,width=\textwidth]{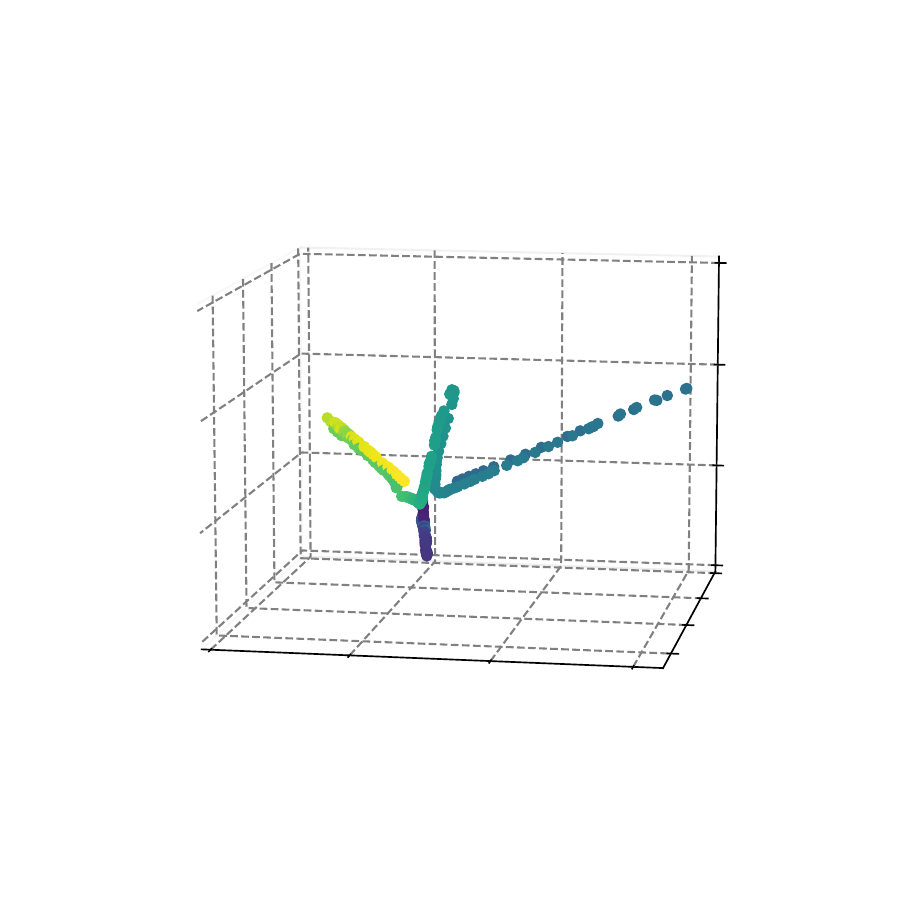}
    \end{subfigure}\hfill
    \begin{subfigure}[b]{0.18\linewidth}
           \includegraphics[trim=90pt 100pt 90pt 100pt, clip,width=\textwidth]{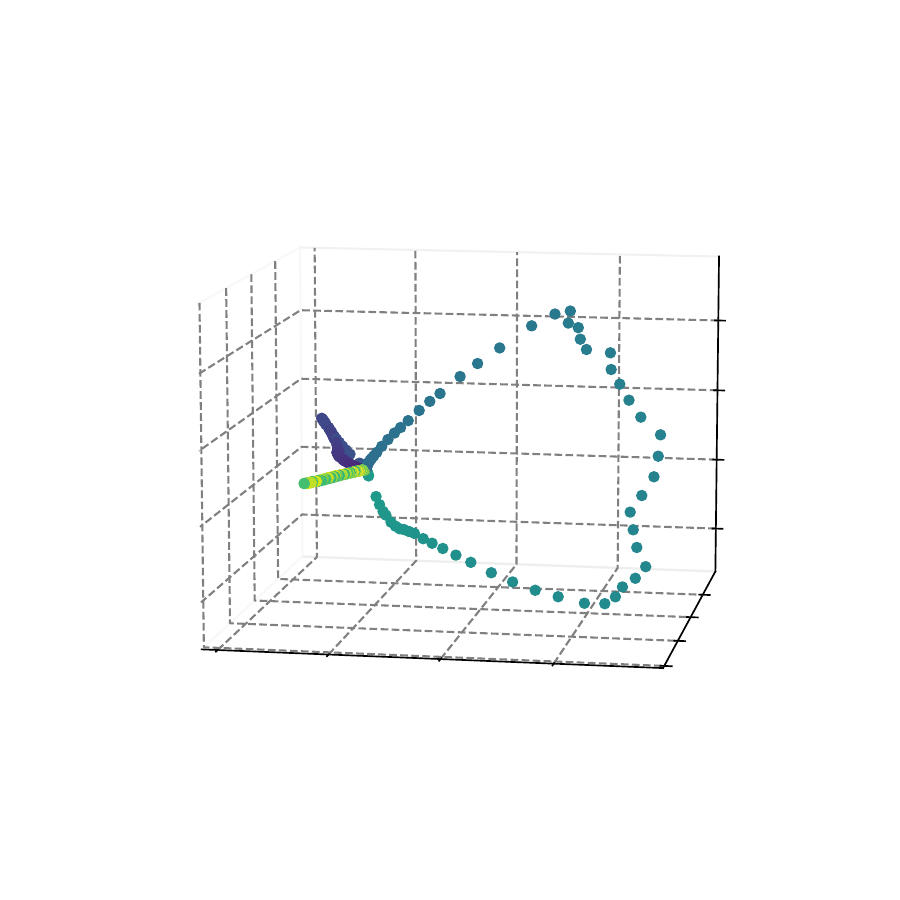}
    \end{subfigure}\hfill
    \begin{subfigure}[b]{0.18\linewidth}
           \includegraphics[trim=90pt 100pt 90pt 100pt, clip,width=\textwidth]{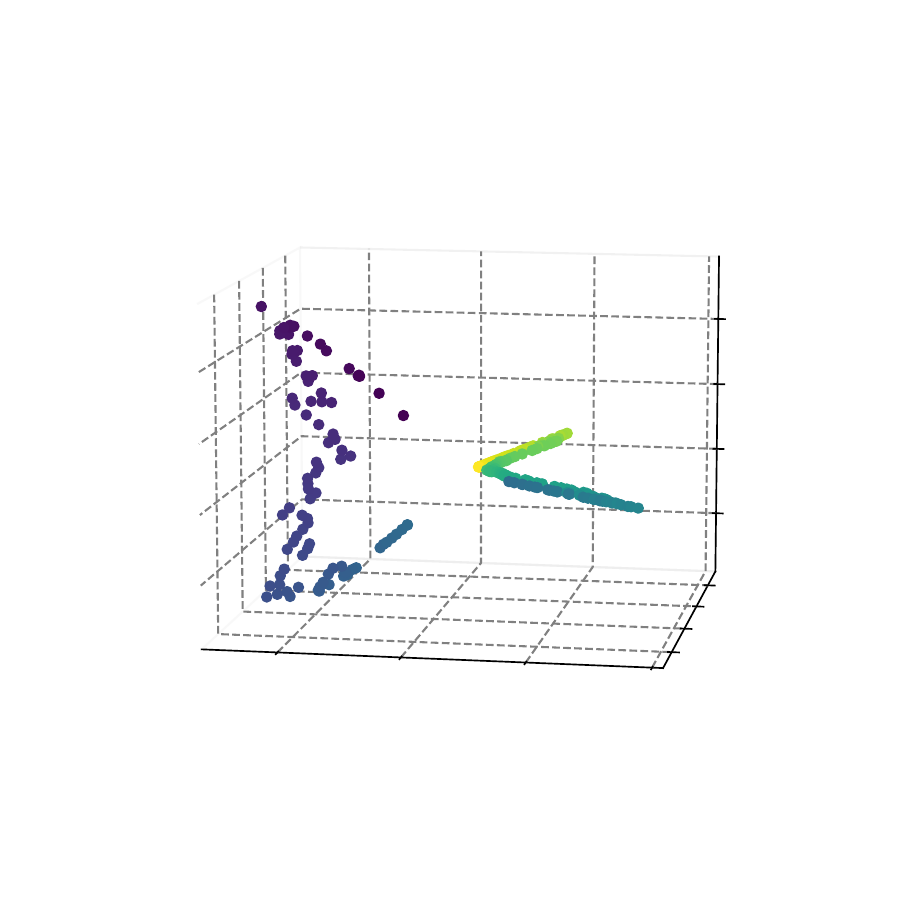}
    \end{subfigure}\hfill
    \begin{subfigure}[b]{0.18\linewidth}
           \includegraphics[trim=90pt 100pt 90pt 100pt, clip,width=\textwidth]{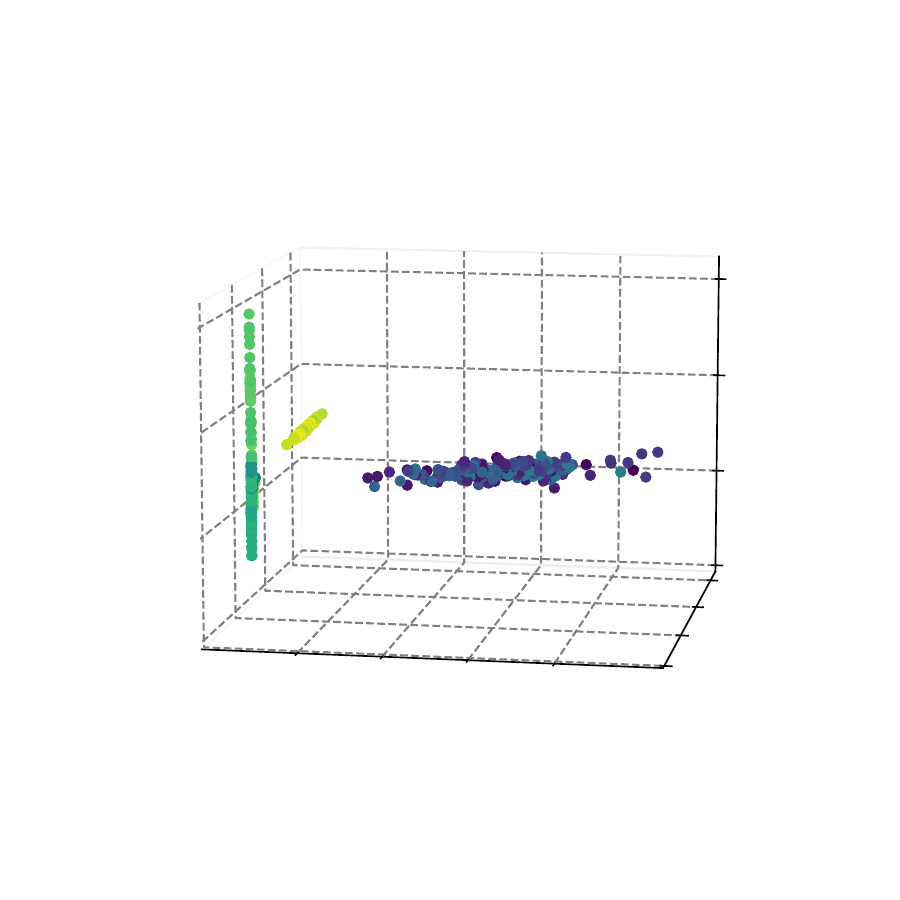}
    \end{subfigure}\\
        \begin{subfigure}[b]{0.18\linewidth}
           \includegraphics[trim=90pt 100pt 90pt 100pt, clip,width=\textwidth]{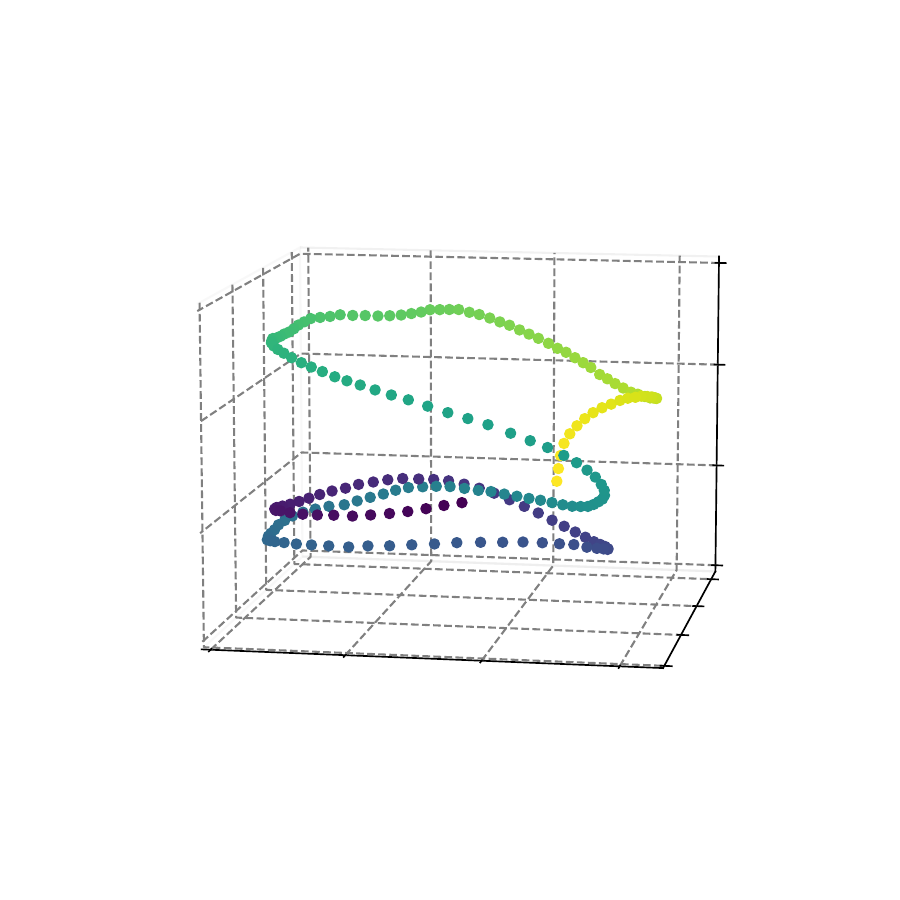}
            \caption{Weiz}
    \end{subfigure}\hfill
    \begin{subfigure}[b]{0.18\linewidth}
            \includegraphics[trim=90pt 100pt 90pt 100pt, clip,width=\textwidth]{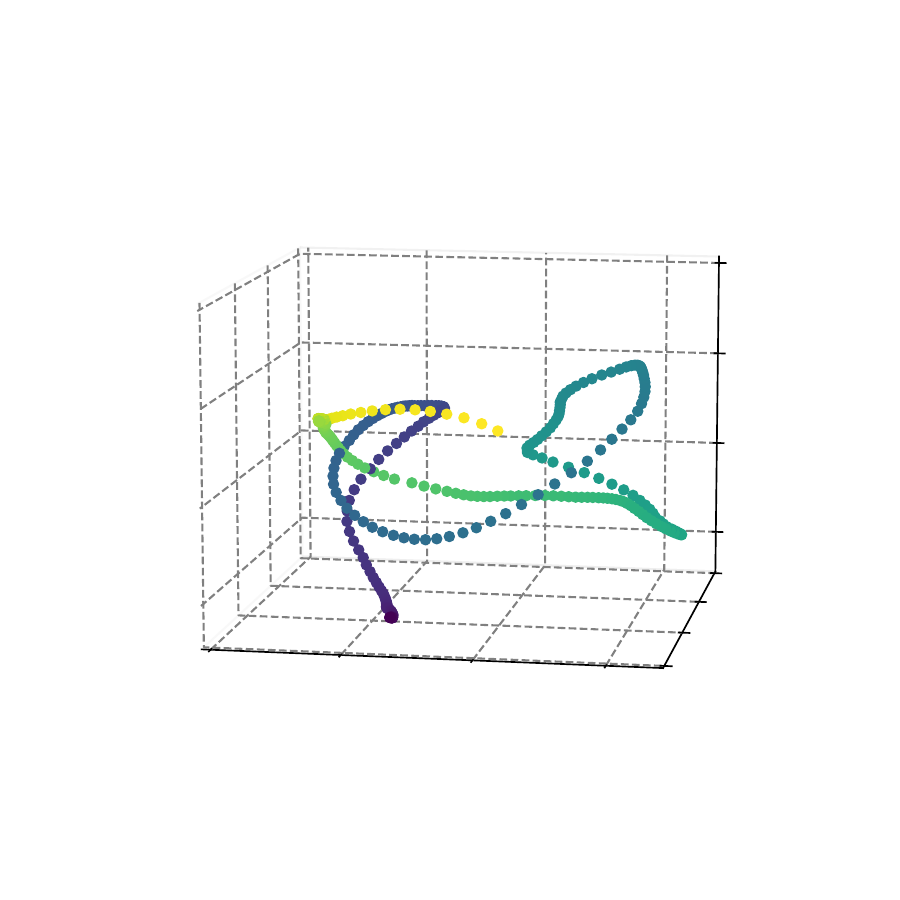}
            \caption{Keck}
    \end{subfigure}\hfill
    \begin{subfigure}[b]{0.18\linewidth}
           \includegraphics[trim=90pt 100pt 90pt 100pt, clip,width=\textwidth]{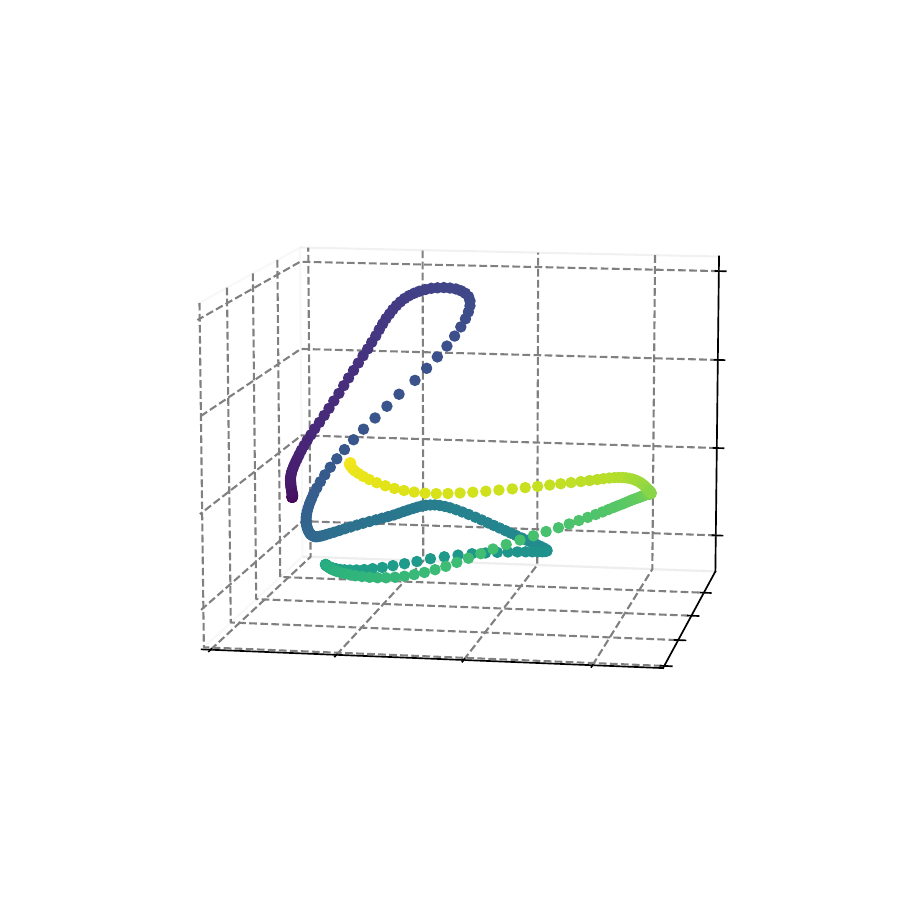}
            \caption{UT}
    \end{subfigure}\hfill
    \begin{subfigure}[b]{0.18\linewidth}
           \includegraphics[trim=90pt 100pt 90pt 100pt, clip,width=\textwidth]{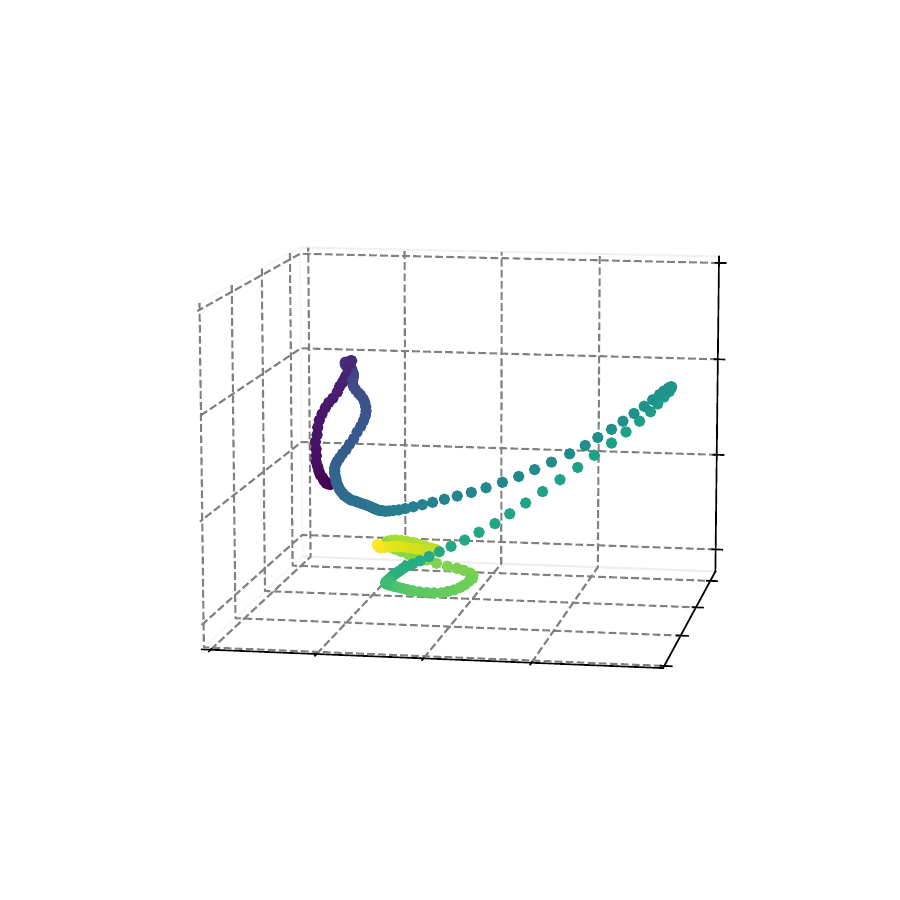}
            \caption{MAD}
    \end{subfigure}\hfill
    \begin{subfigure}[b]{0.18\linewidth}
           \includegraphics[trim=90pt 100pt 90pt 100pt, clip,width=\textwidth]{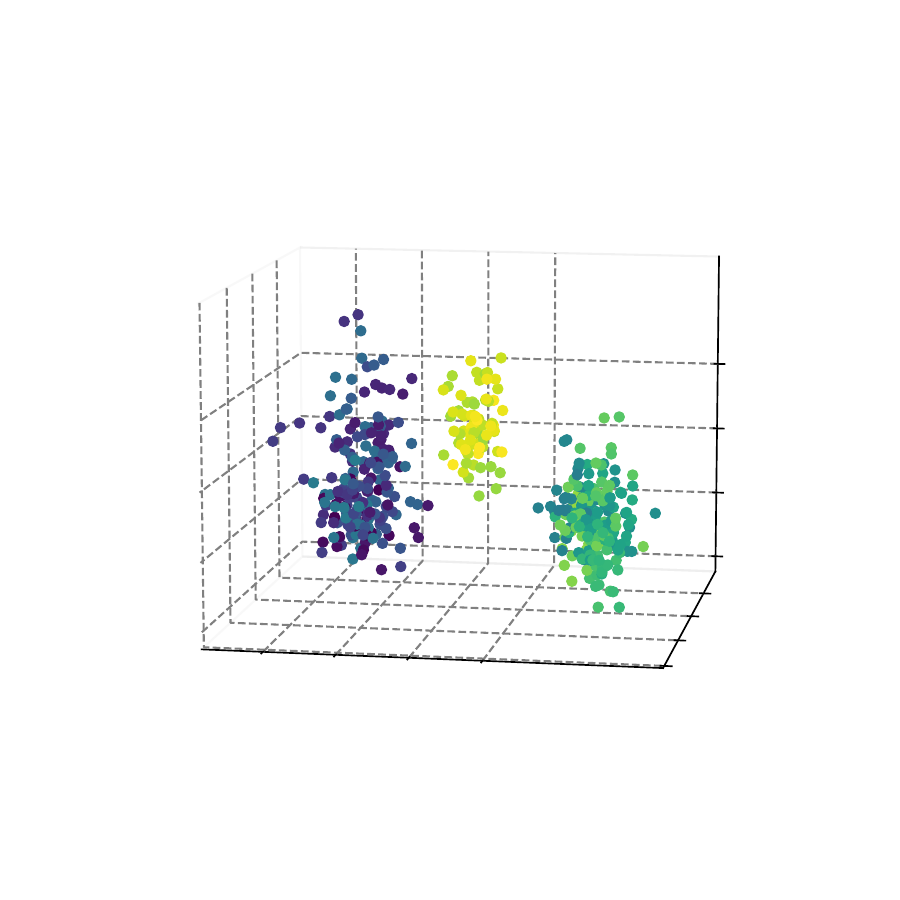}
            \caption{YouTube}
    \end{subfigure}
    
\caption{\textbf{PCA visualization of learned representations.} First row: representations learned by \trc{} . Second row: representations learned by GCTSC. We conduct experiments on the first sequence of each dataset.}
\label{fig:PCA-supp}
\end{figure*}

In the main text, we have visualized the representations from different motions by different colors to demonstrate the union-of-orthogonal-subspaces distribution of learned representations.
To further demonstrate the temporal continuity of learned representations, we visualize the data points (\ie, the feature vectors of frames in video) with a continuously varying colormap. %
As illustrated in Figure~\ref{fig:PCA-supp} (in the first row), while the temporal consistency is preserved, the distribution of the representations learned by \trc{} are ``compressed'' in a structured way; whereas  
the learned representations by GCTSC (in the second row) %
do preserve the temporal continuity very well, but lack of specific structures to facilitate the task of motion segmentation. 

\subsection{Clustering Performance Evaluation on Different Representations} %
\begin{figure}[h]
    \centering
    \begin{subfigure}[b]{0.42\linewidth}
            \includegraphics[width=\textwidth]{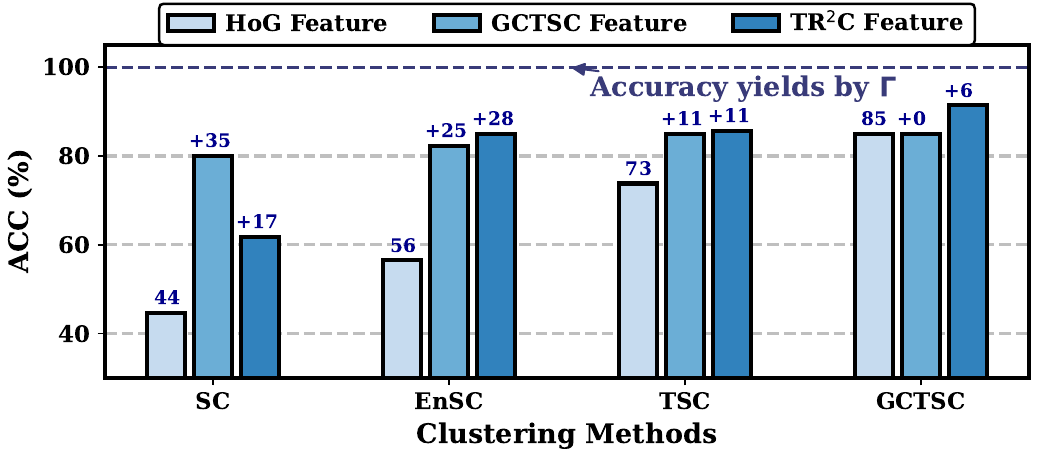}
            \caption{Weizmann}
    \end{subfigure}
    \begin{subfigure}[b]{0.42\linewidth}
            \includegraphics[width=\textwidth]{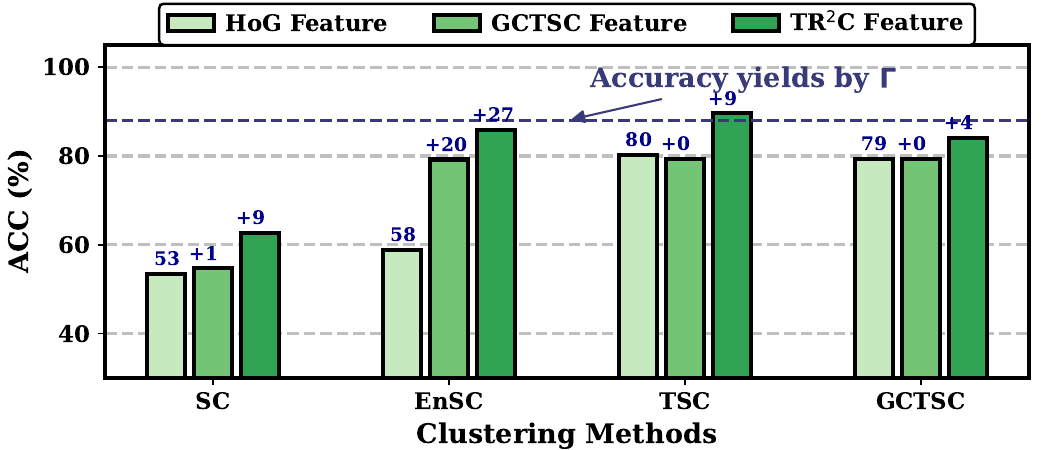}
            \caption{Keck}
    \end{subfigure}\\
    \begin{subfigure}[b]{0.42\linewidth}
            \includegraphics[width=\textwidth]{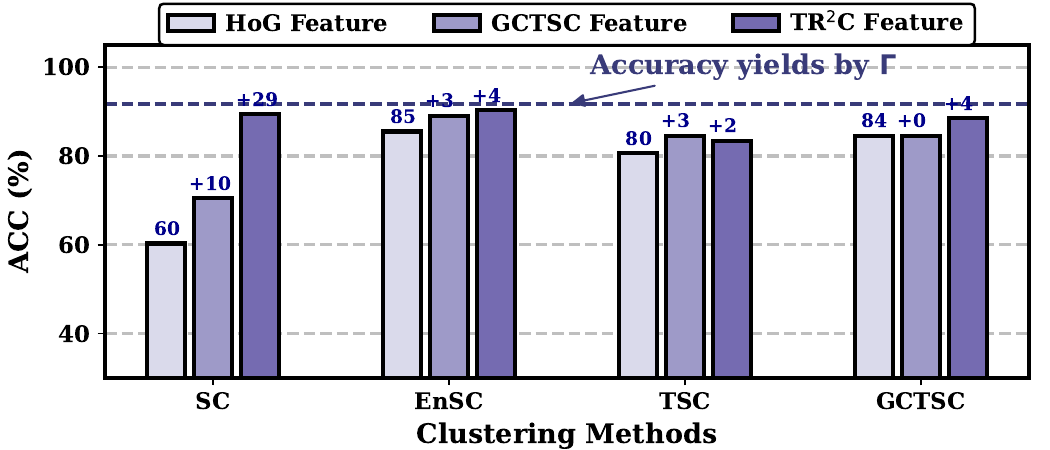}
            \caption{UT}
    \end{subfigure}
    \begin{subfigure}[b]{0.42\linewidth}
            \includegraphics[width=\textwidth]{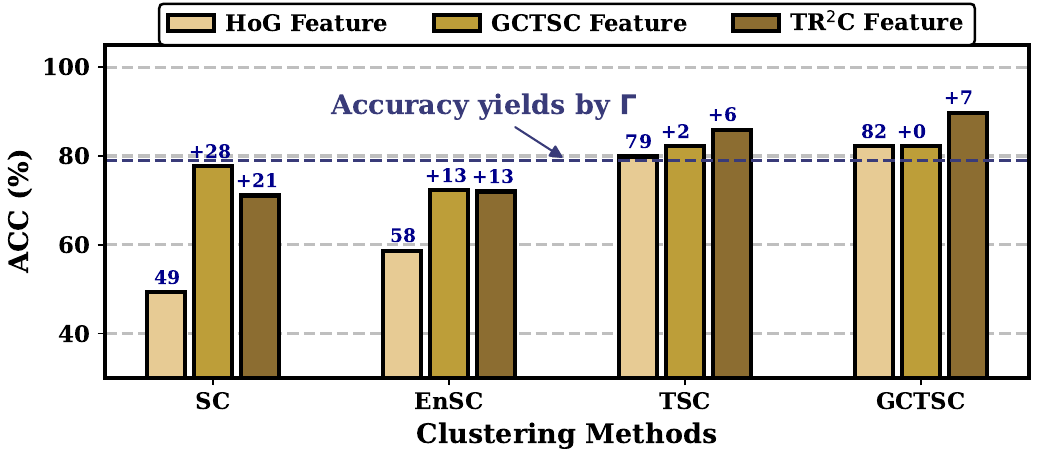}
            \caption{MAD}
    \end{subfigure}\\
    \begin{subfigure}[b]{0.42\linewidth}
            \includegraphics[trim=3pt 0pt 8pt 0pt, clip,width=\textwidth]{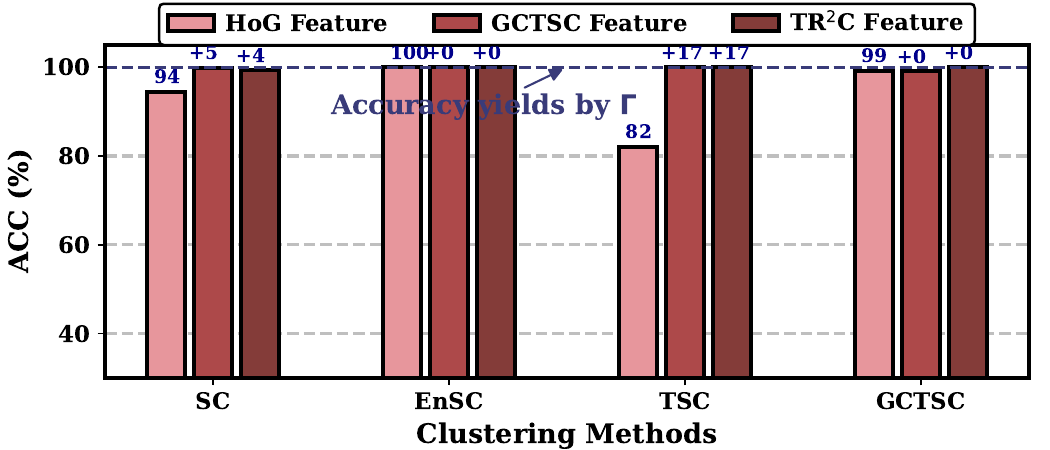}
            \caption{YouTube}
    \end{subfigure}
    \caption{\textbf{Clustering performance evaluation on different representations.}}
    \label{fig:feature_TSC_GCTSC}
\end{figure}

To further validate the effectiveness of \trc{}, we %
use the HoG features, the representations learned by GCTSC and \trc{} as the input and evaluate the performance of different methods, including 
Spectral Clustering (SC), Elastic Net Subspace Clustering (EnSC), TSC and GCTSC.
As shown in Figure~\ref{fig:feature_TSC_GCTSC}, the clustering performance of representations from \trc{} consistently surpasses that of HoG features, regardless of the datasets and clustering methods used.
The performance gap is notably larger when clustering with SC and EnSC, as these classical clustering approaches overlook the temporal consistency prior of HMS.
In contrast, the performance improvements in TSC and GCTSC, which incorporate temporal consistency regularizers, are largely driven by the union-of-orthogonal-subspaces distribution learned by \trc{}.
Notably, the representations learned by GCTSC also achieve satisfying clustering performance, though they do not outperform \trc{}, except for clustering with SC on datasets Weizmann and MAD.
It is surprising that the accuracy yields by $\boldGamma$ of \trc{} even outperforms ``\trc{} features$+$GCTSC'' on all the datasets except for the MAD, implying that the reparameterized affinity matrix is better at capturing the subspace membership.

\subsection{Ablation Study}
We report the ablation study results of all the benchmarks in Table~\ref{tab:ablation-supp}. The performances are averaged across all the sequences of each dataset.
As analyzed in the main text, each term in \trc{} is indispensable for the learning of temporally consistent representations that align with a union of orthogonal subspaces.

\begin{table}[h]
  \centering
  \caption{\textbf{Ablation study.} We report the average performance of all the sequences.}
  \resizebox{0.95\linewidth}{!}{
    \begin{tabular}{P{1.4cm}P{1.4cm}P{1.4cm}|P{1cm}P{1cm}|P{1cm}P{1cm}|P{1cm}P{1cm}|P{1cm}P{1cm}|P{1cm}P{1cm}}
    \toprule
    \rowcolor{myGray} \multicolumn{3}{c|}{Loss} & \multicolumn{2}{c|}{Weiz} &\multicolumn{2}{c|}{Keck} &\multicolumn{2}{c|}{UT} &\multicolumn{2}{c|}{MAD} &\multicolumn{2}{c}{YouTube}\\
    \rowcolor{myGray} $\mathcal{L}_{\rho}$& $\mathcal{L}_{\bar\rho^c}$& $\mathcal{L}_{r}$& ACC   & NMI & ACC   & NMI & ACC   & NMI & ACC   & NMI & ACC   & NMI\\
    \midrule
    $\checkmark$& $\checkmark$&      & 37.30 & 45.86 & 47.29 & 49.78 & 45.79 & 35.30 & 30.27& 29.40 & 94.82 & 97.30 \\
          & $\checkmark$& $\checkmark$& 53.14 & 61.51 & 47.91 & 51.39 & 63.13 & 59.51  & 50.54 & 53.23 & 96.07 & 97.77 \\
    $\checkmark$&       & $\checkmark$&  64.68  & 74.67 & 58.60 & 65.21 &    65.67&  66.09 & 64.91 & 72.37 & 48.16 & 53.36 \\
    $\checkmark$ &     &      &  41.21 & 44.57 & 44.01 & 41.46 & 46.80 & 37.49& 28.00 & 22.97& 58.87 & 54.08 \\
         &  $\checkmark$ &        & 56.03 & 64.19 & 47.50 & 52.11 &  76.39 & 72.41& 43.23 & 43.11 & 90.15 & 91.79 \\
       &      & $\checkmark$  & 52.59 & 60.33 & 48.35 & 50.87 & 62.13 & 58.29& 50.54 & 53.13 & 96.01 & 97.52 \\
        $\checkmark$& $\checkmark$& $\checkmark$& \textbf{94.07} & \textbf{96.08} & \textbf{86.78} & \textbf{86.93} & \textbf{94.05} & \textbf{92.34}& \textbf{83.99} & \textbf{87.32} & \textbf{96.40} & \textbf{98.50}\\ 
    \bottomrule
    \end{tabular}%
    }
  \label{tab:ablation-supp}%
\end{table}%

\subsection{Complexity Analysis}
\label{sec:Complexity_Analysis}

We analyze the time complexity of $\log\det(\cdot)$ operation, as it is the most computationally intensive component in \trc{}.
By the commutative property: $\log\det(\boldsymbol{I}+\Z\Z^\top)=\log\det(\boldsymbol{I}+\Z^\top\Z)$ (see~\cite{Ma:PAMI07}), we reduce the matrix size involved in $\log\det(\cdot)$ from $N\times N$ to $d\times d$, which significantly improves both time and memory efficiency, especially when $d \ll N$.
Since that $-\mathcal{L}_{\rho}+\mathcal{L}_{\bar\rho^c}$ requires computing $\log\det(\cdot)$ for $N+1$ times, the complexity of our loss becomes $\mathcal{O}(Nd^3)$, which can be further accelerated with GPU support.
We report the time cost (ms/iter) of \trc{} with varying $N$ on HoG features ($d=324$) in Table~\ref{tab:time_vary_N}.
As can be seen, both the time and memory cost of \trc{} are significantly reduced by exploiting the commutative property of $\log\det(\cdot)$ operation.

\begin{table}[htbp]
  \centering
  \caption{Time cost (ms/iter) with varying $N$ on HoG features. ``OOM'' refers to out-of-memory.}
  \label{tab:time_vary_N}
  \resizebox{0.7\linewidth}{!}{
    \begin{tabular}{lcccccccc|c}
    \toprule
   \rowcolor{myGray} $N$  & 200   & 400   & 600   & 800   & 1000  & 2000  & 3000  & 4000 & Complexity\\
    \midrule
    w/o Commutation & 33.2  & 97.1  & 229.1  & 546.8  & 1039.3  & OOM   & OOM   & OOM  & $\mathcal{O}(N^4)$\\
    Our \trc{} & 16.1  & 17.7  & 21.3  & 23.6  & 28.0  & 53.9  & 105.2  & 162.9   & $\mathcal{O}(Nd^3)$ \\
    \bottomrule
    \end{tabular}}
\end{table}%

\subsection{Learning Curves}
\begin{figure*}[bt]
    \centering
    \begin{subfigure}[b]{0.48\linewidth}
    \begin{subfigure}[b]{0.5\linewidth}
           \includegraphics[width=\textwidth, height=0.75\textwidth]{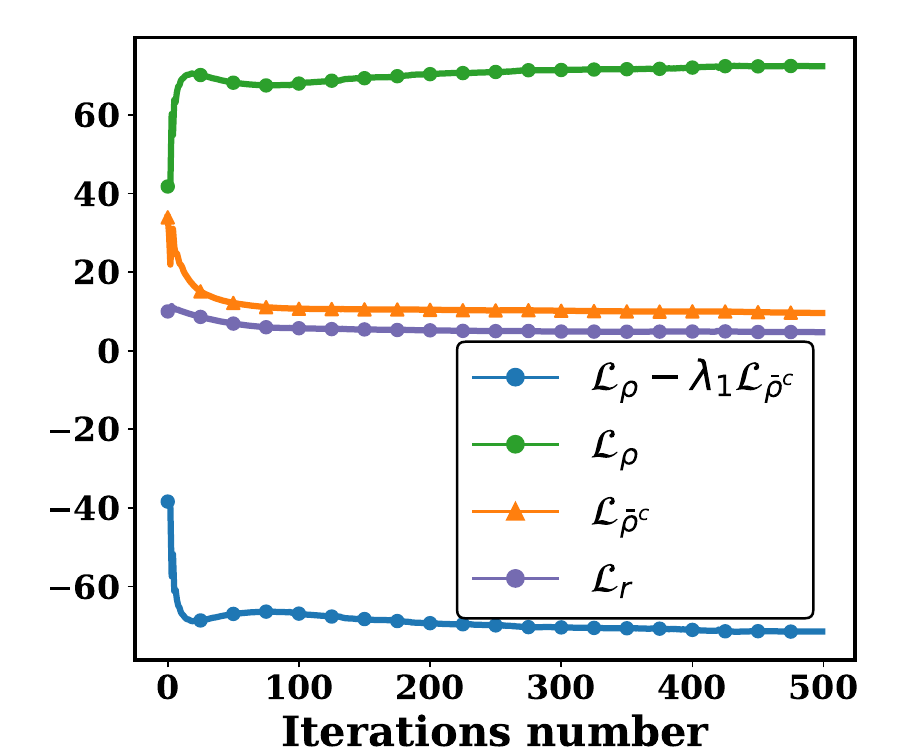}
    \end{subfigure}\hfill
    \begin{subfigure}[b]{0.5\linewidth}
           \includegraphics[width=\textwidth, height=0.75\textwidth]{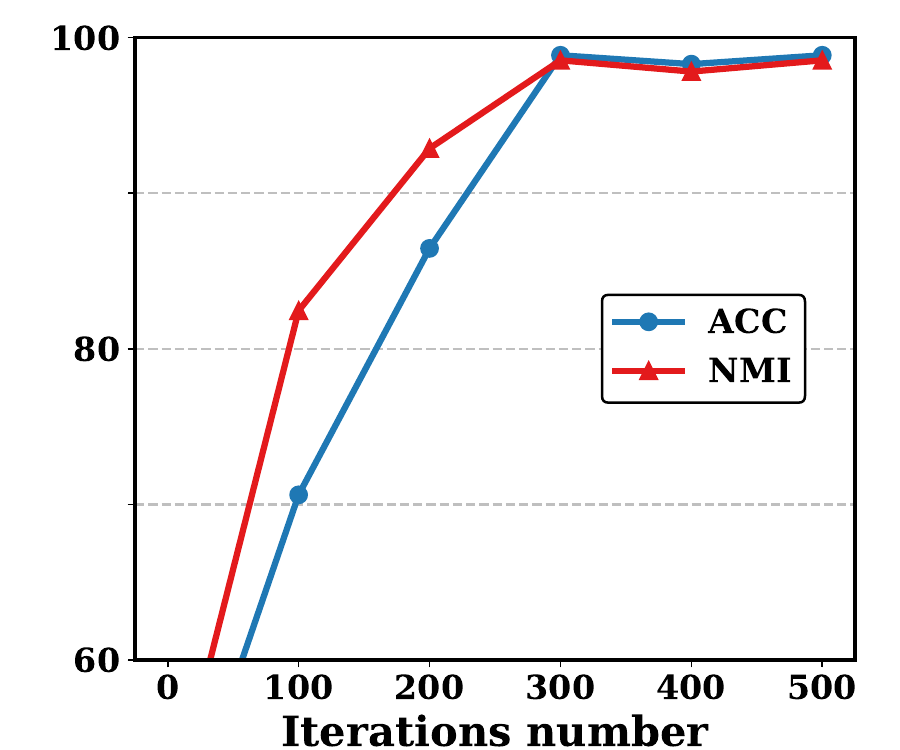}
    \end{subfigure}
    \caption{Weizmann Dataset}
    \end{subfigure}
    \begin{subfigure}[b]{0.48\linewidth}
    \begin{subfigure}[b]{0.5\linewidth}
           \includegraphics[width=\textwidth, height=0.75\textwidth]{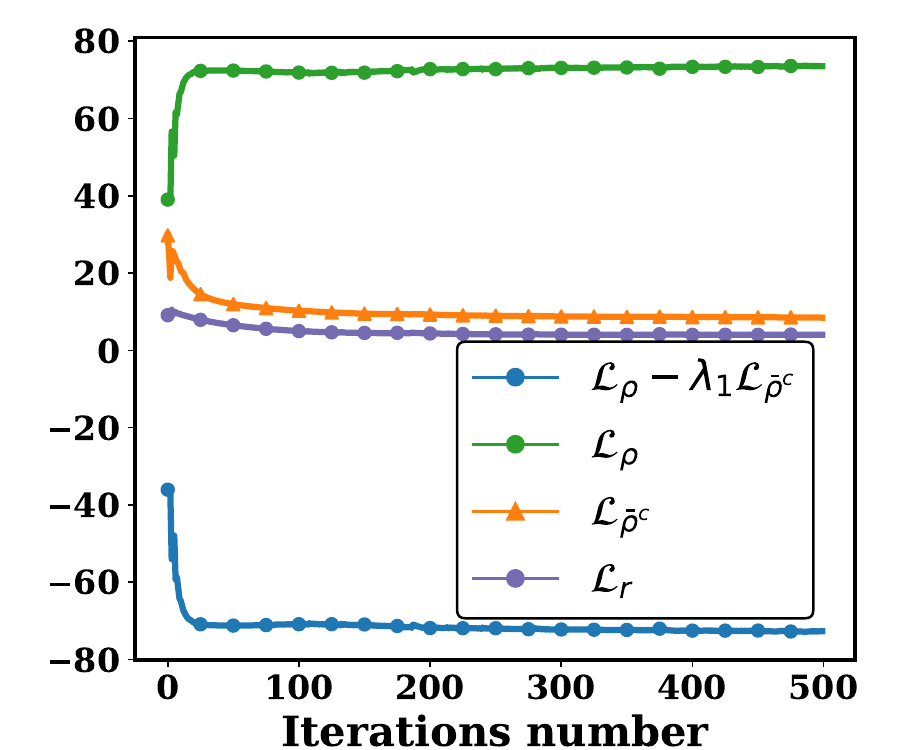}
    \end{subfigure}\hfill
    \begin{subfigure}[b]{0.5\linewidth}
           \includegraphics[width=\textwidth, height=0.75\textwidth]{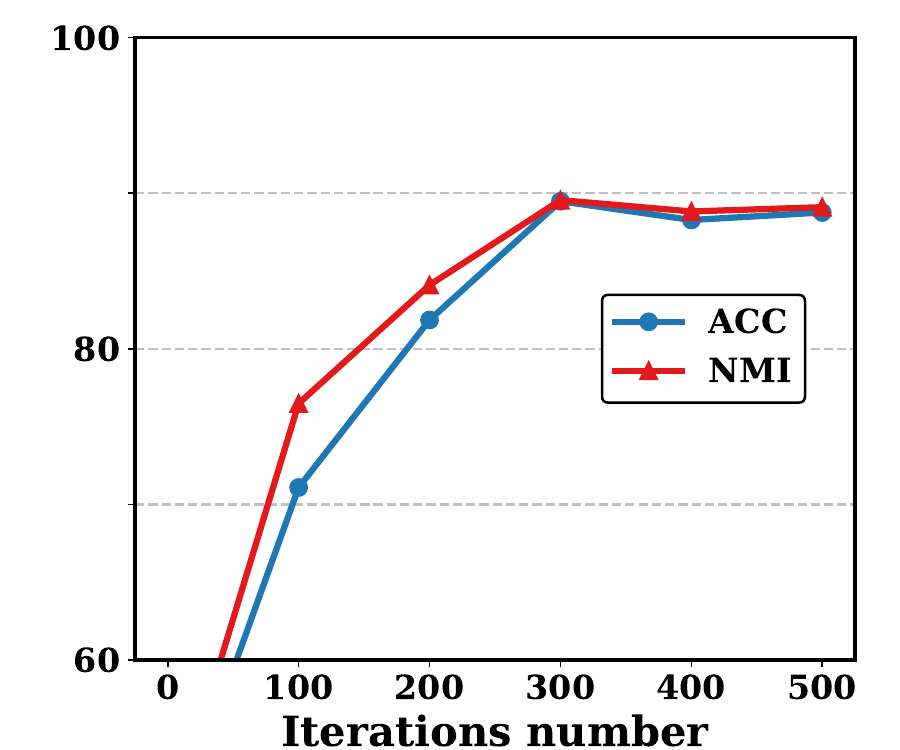}
    \end{subfigure}
    \caption{Keck Dataset}
    \end{subfigure}\\
    \begin{subfigure}[b]{0.48\linewidth}
    \begin{subfigure}[b]{0.5\linewidth}
           \includegraphics[width=\textwidth, height=0.75\textwidth]{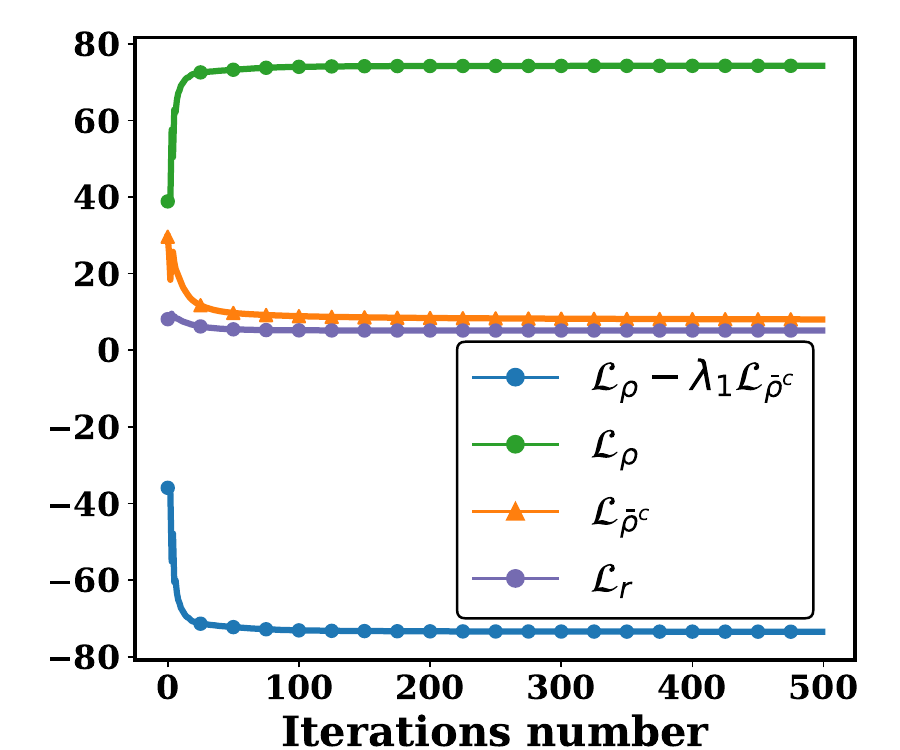}
    \end{subfigure}\hfill
    \begin{subfigure}[b]{0.5\linewidth}
           \includegraphics[width=\textwidth, height=0.75\textwidth]{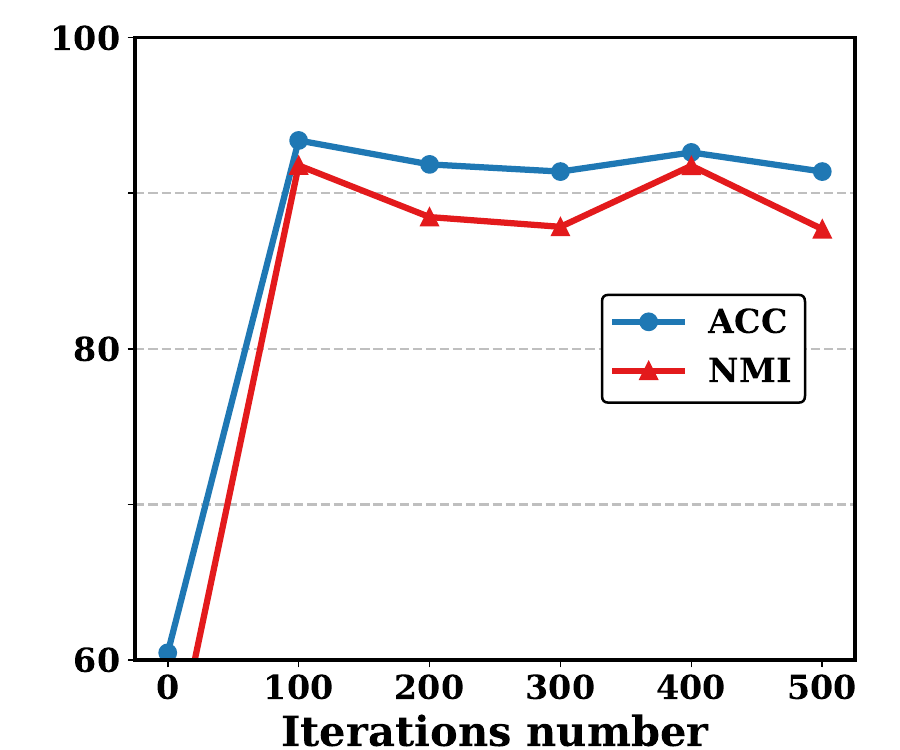}
    \end{subfigure}
    \caption{UT Dataset}
    \end{subfigure}
    \begin{subfigure}[b]{0.48\linewidth}
    \begin{subfigure}[b]{0.5\linewidth}
           \includegraphics[width=\textwidth, height=0.75\textwidth]{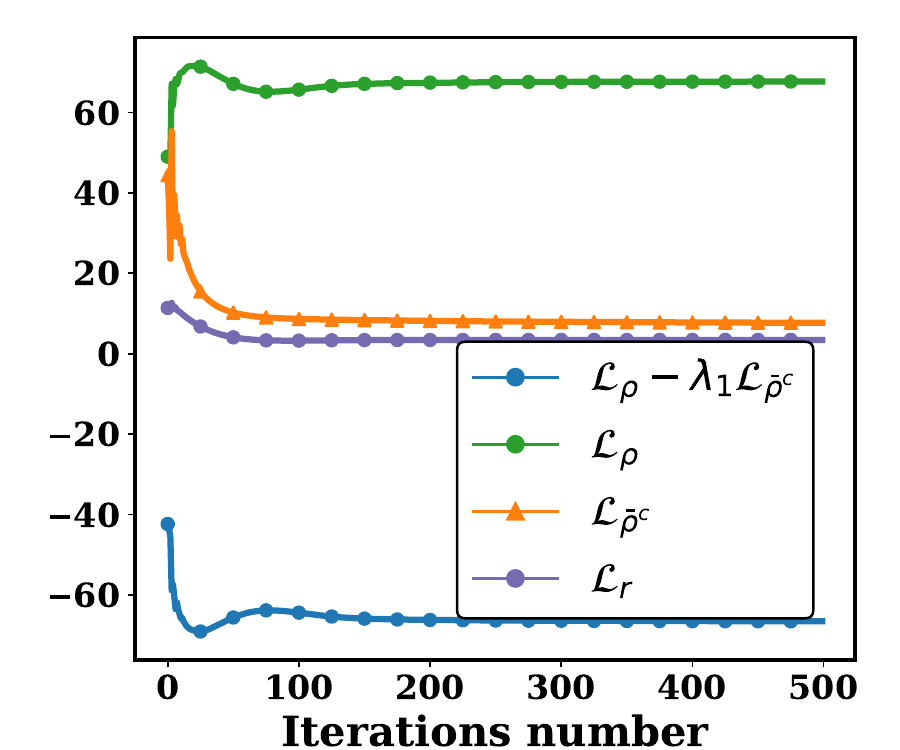}
    \end{subfigure}\hfill
    \begin{subfigure}[b]{0.5\linewidth}
           \includegraphics[width=\textwidth, height=0.75\textwidth]{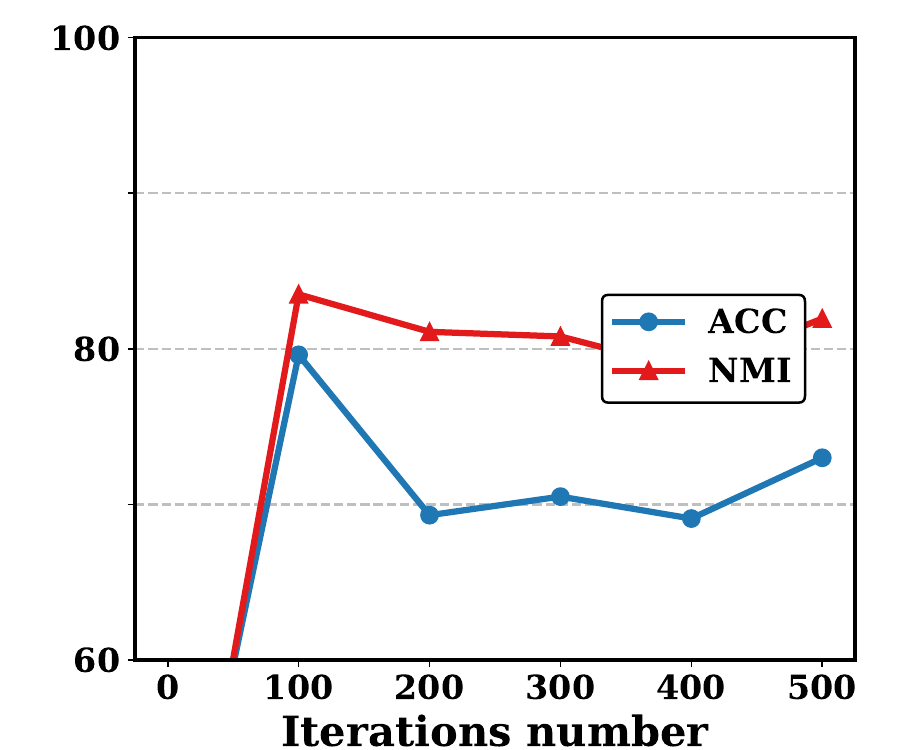}
    \end{subfigure}
    \caption{MAD Dataset}
    \end{subfigure}\\
    \begin{subfigure}[b]{0.48\linewidth}
    \begin{subfigure}[b]{0.5\linewidth}
           \includegraphics[width=\textwidth, height=0.75\textwidth]{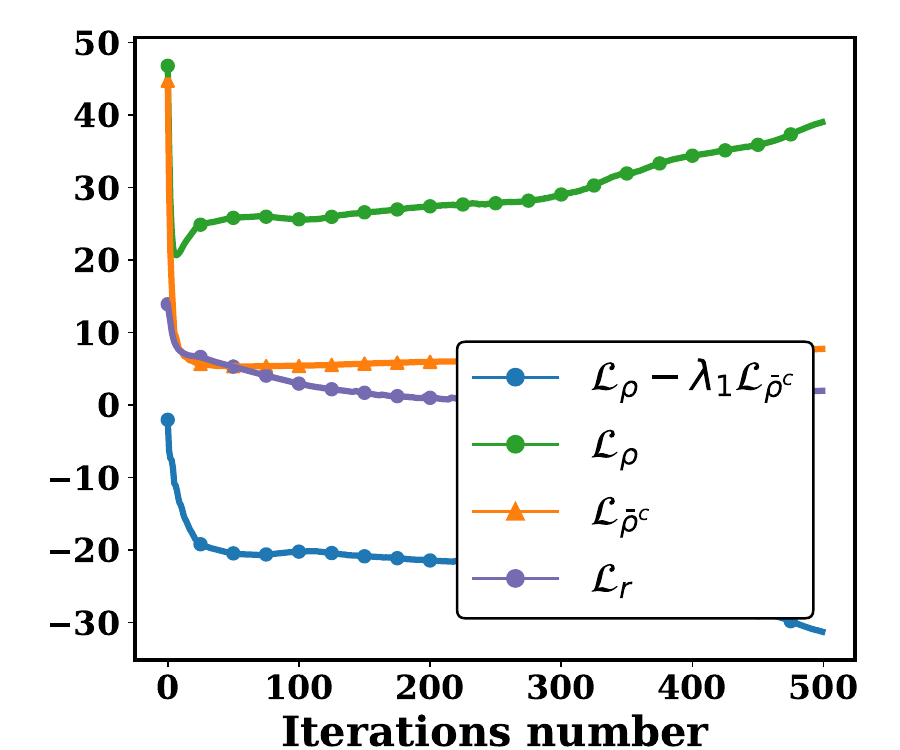}
    \end{subfigure}\hfill
    \begin{subfigure}[b]{0.5\linewidth}
           \includegraphics[width=\textwidth, height=0.75\textwidth]{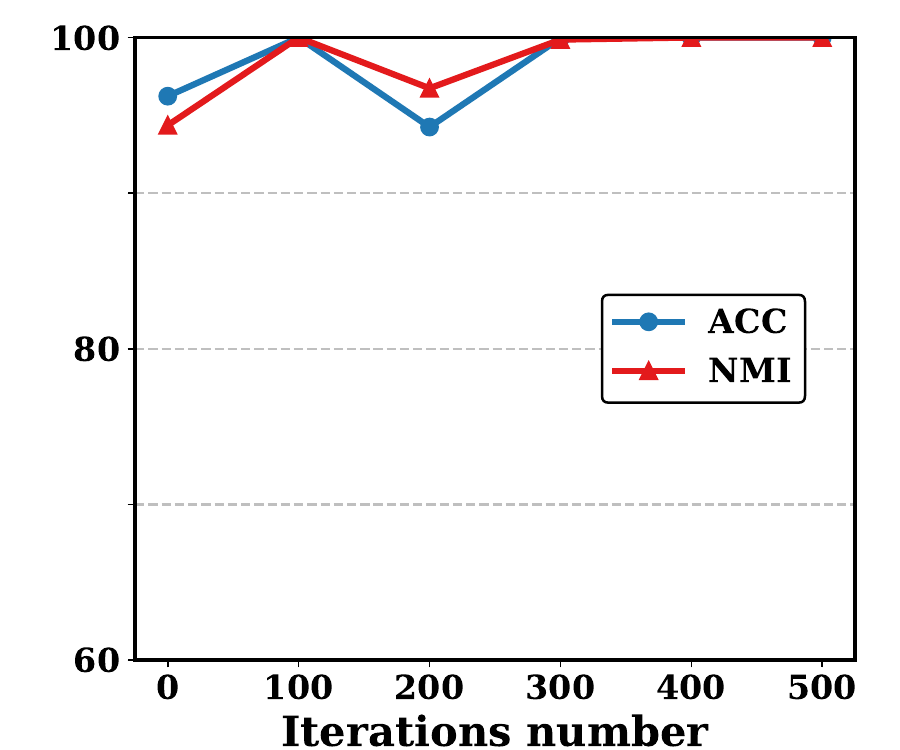}
    \end{subfigure}
    \caption{YouTube Dataset}
    \end{subfigure}
\caption{\textbf{Learning curves of the \trc{} framework on HoG features.}}
\label{fig:learning curves}
\end{figure*}

We plot the learning curves with respect to $\mathcal{L}_{\rho}-\lambda_1\mathcal{L}_{\bar\rho^c}$, $\mathcal{L}_{\rho}$, $\mathcal{L}_{\bar\rho^c}$, $\mathcal{L}_{r}$ and the clustering performance in Figure~\ref{fig:learning curves}.
As illustrated, the gap between $\mathcal{L}_{\rho}$ and $\mathcal{L}_{\bar\rho^c}$ increases rapidly as the $\mathcal{L}_{\rho}-\lambda_1\mathcal{L}_{\bar\rho^c}$ decreases, encouraging the UoS structure of learned representations.
The $\mathcal{L}_{r}$ decreases, promoting the temporal continuity of learned representations.
Consequently, the clustering results gradually converge to state-of-the-art performances.

\subsection{Segmentation Results Visualization}
\label{sec:Segmentation Results Visualization}
To qualitatively demonstrate the effectiveness of \trc{}, we visualize the video segmentation results along with the ground-truth labels for the first three sequences on the five benchmark datasets.
Notably, our unsupervised \trc{} produces segmentation results that closely match the manually annotated ground-truth labels on the Weizmann, UT, and YouTube datasets.
For the Keck and MAD datasets, segmentation errors primarily occur in frames capturing transitions between different human motions.
For instance, in the Keck dataset, these frames often show individuals adjusting their standing positions, making it inherently difficult to determine whether they belong to the preceding or the subsequent motion motion.

\begin{figure*}[tbp]
    \centering
    \begin{subfigure}[b]{0.33\linewidth}
    \begin{subfigure}[b]{\linewidth}
           \includegraphics[width=\textwidth]{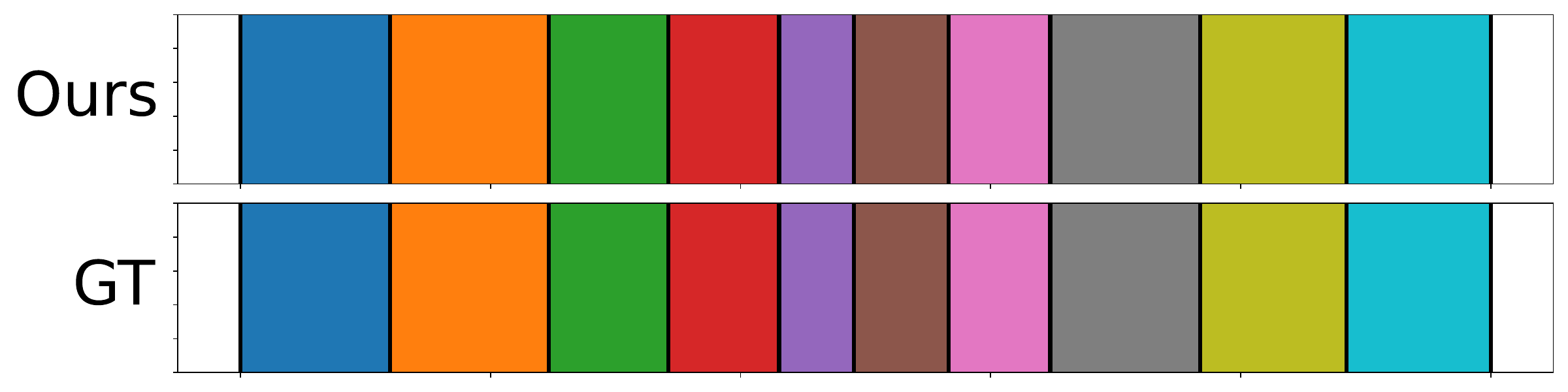}
    \end{subfigure}\\
    \begin{subfigure}[b]{\linewidth}
           \includegraphics[width=\textwidth]{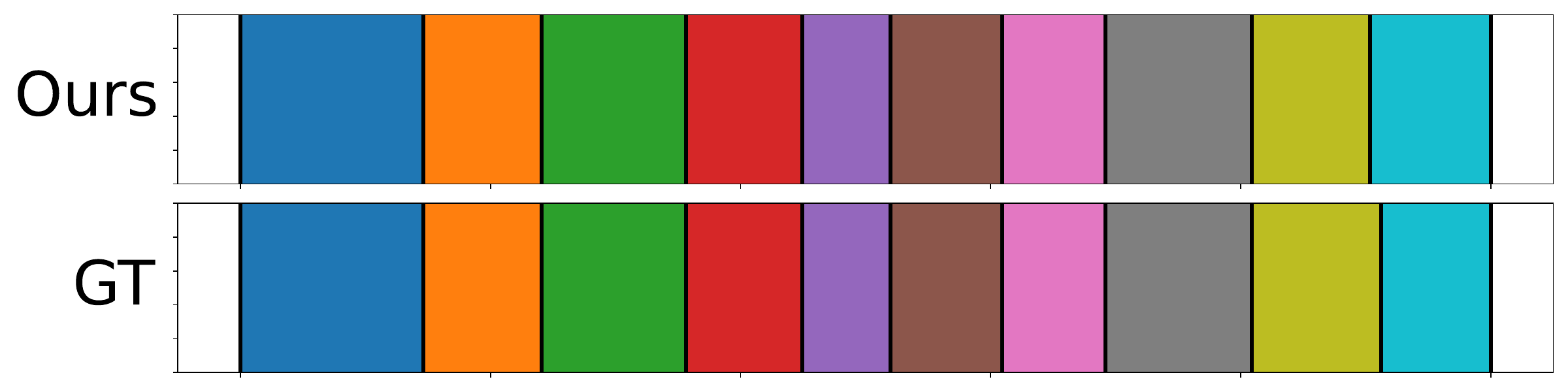}
    \end{subfigure}\\
    \begin{subfigure}[b]{\linewidth}
           \includegraphics[width=\textwidth]{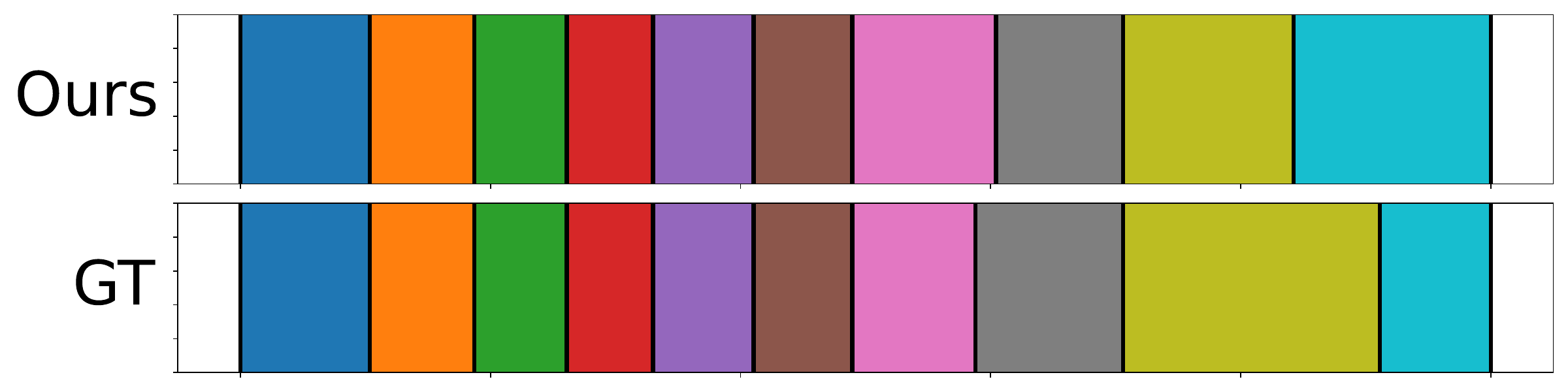}
    \end{subfigure}
    \caption{Weizmann Dataset}
    \end{subfigure}
    \begin{subfigure}[b]{0.33\linewidth}
    \begin{subfigure}[b]{\linewidth}
           \includegraphics[width=\textwidth]{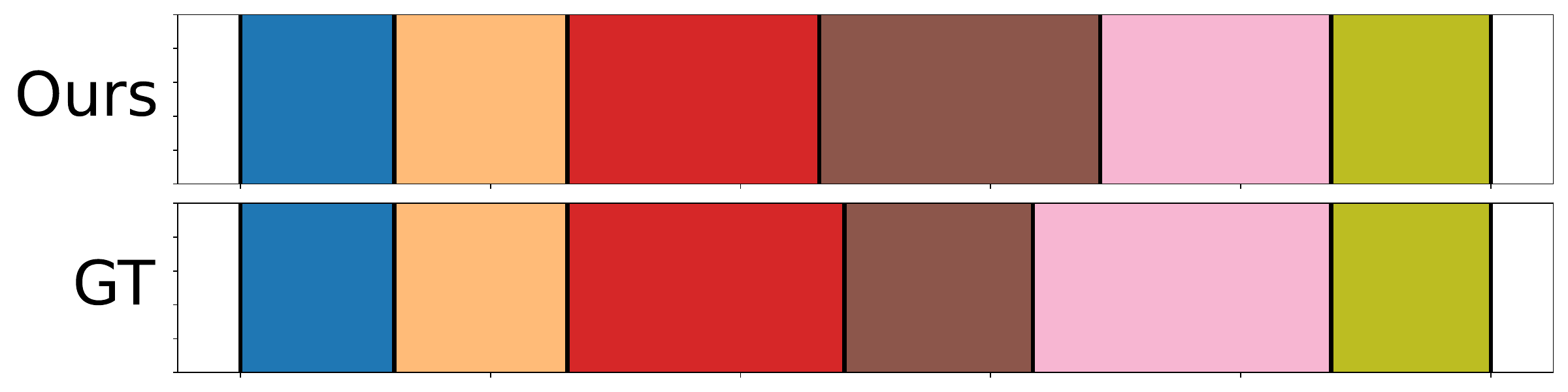}
    \end{subfigure}\\
    \begin{subfigure}[b]{\linewidth}
           \includegraphics[width=\textwidth]{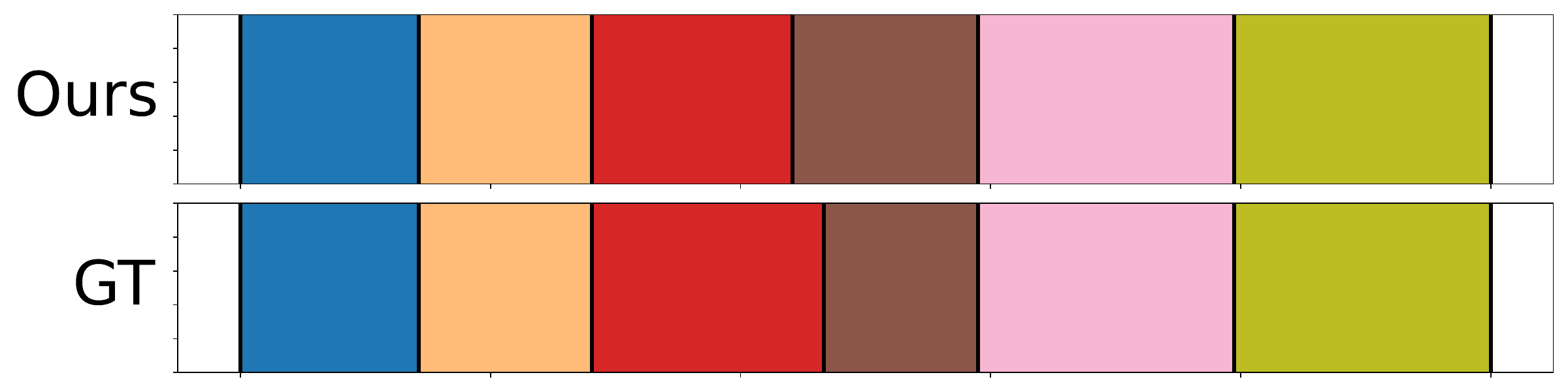}
    \end{subfigure}\\
    \begin{subfigure}[b]{\linewidth}
           \includegraphics[width=\textwidth]{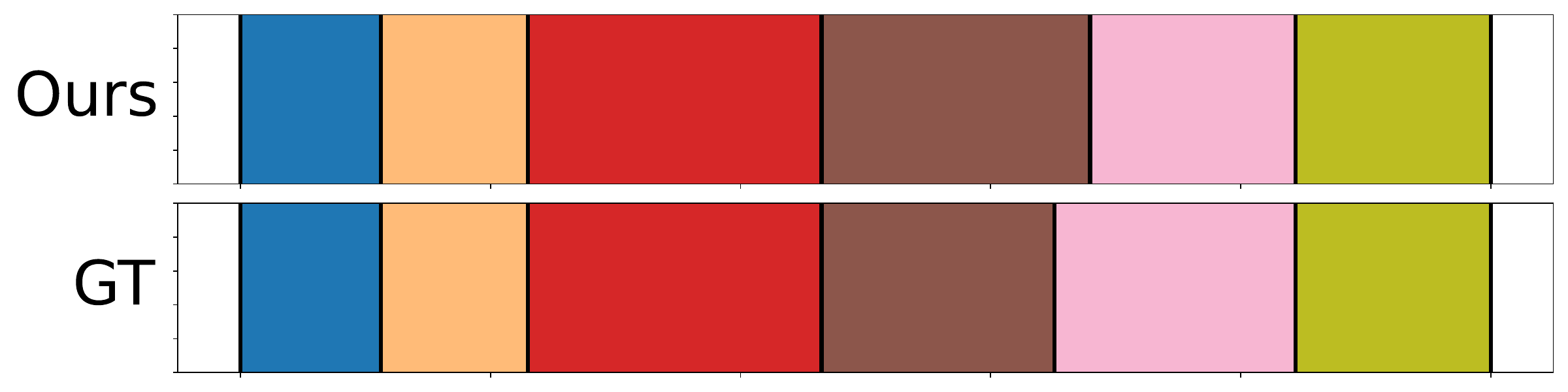}
    \end{subfigure}
    \caption{UT Dataset}
    \end{subfigure}
    \begin{subfigure}[b]{0.33\linewidth}
    \begin{subfigure}[b]{\linewidth}
           \includegraphics[width=\textwidth]{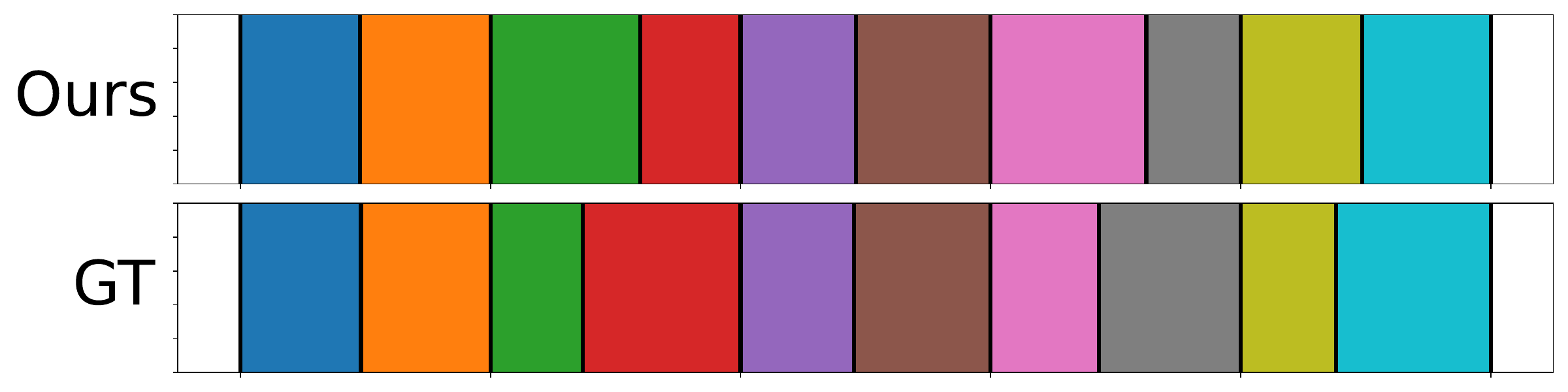}
    \end{subfigure}\\
    \begin{subfigure}[b]{\linewidth}
           \includegraphics[width=\textwidth]{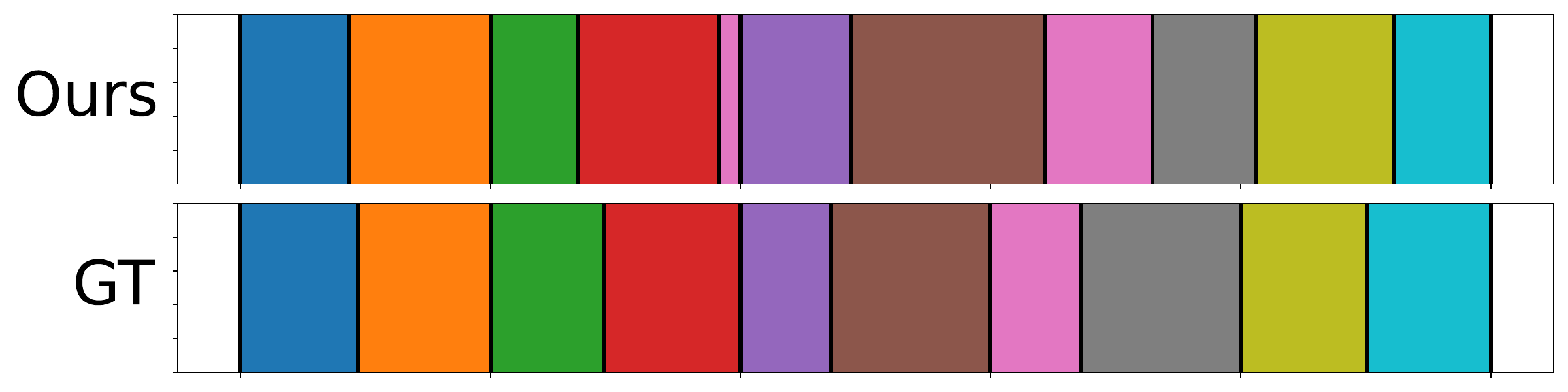}
    \end{subfigure}\\
    \begin{subfigure}[b]{\linewidth}
           \includegraphics[width=\textwidth]{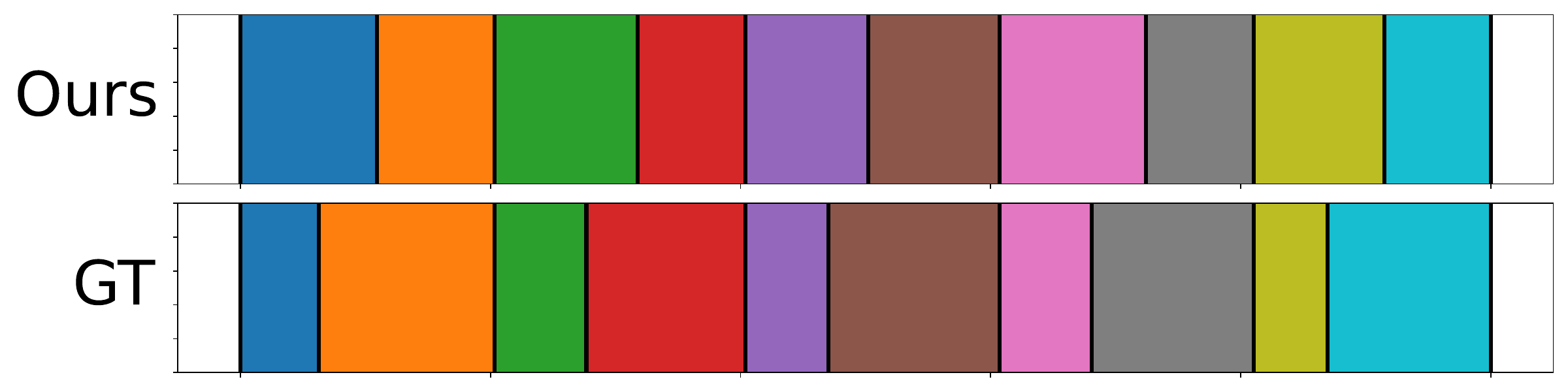}
    \end{subfigure}
    \caption{Keck Dataset}
    \end{subfigure}\\
    \begin{subfigure}[b]{0.33\linewidth}
    \begin{subfigure}[b]{\linewidth}
           \includegraphics[width=\textwidth]{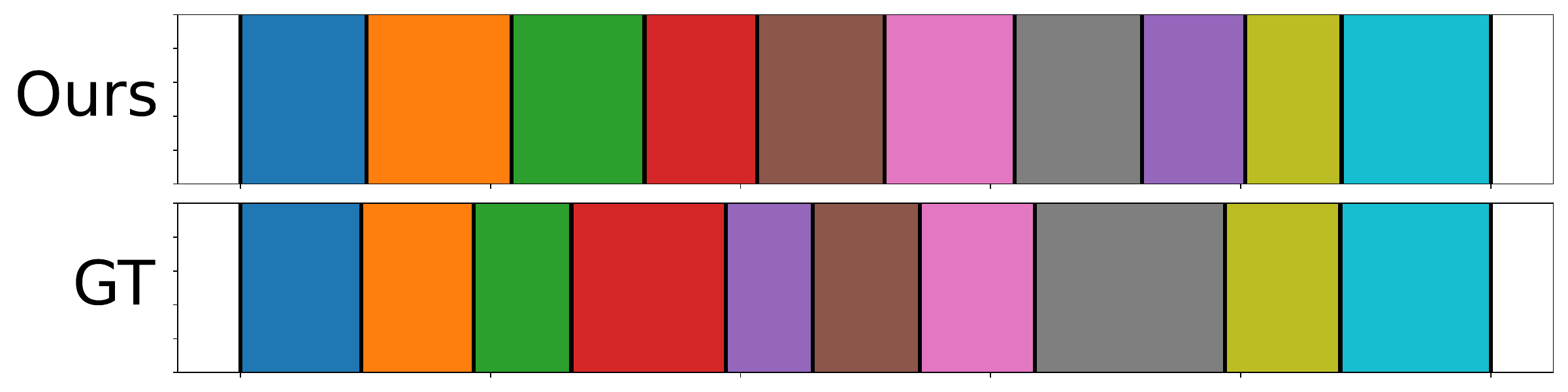}
    \end{subfigure}\\
    \begin{subfigure}[b]{\linewidth}
           \includegraphics[width=\textwidth]{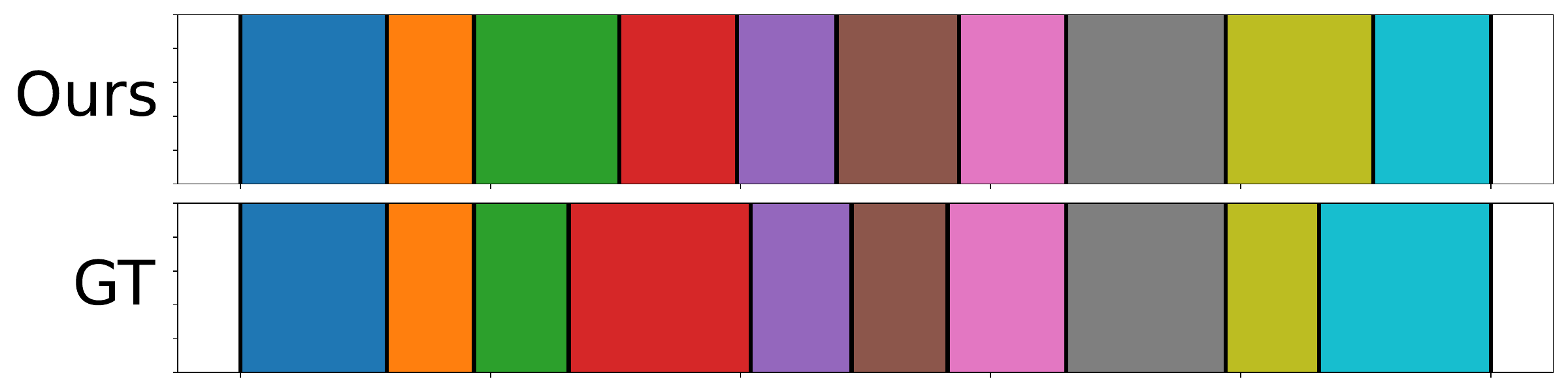}
    \end{subfigure}\\
    \begin{subfigure}[b]{\linewidth}
           \includegraphics[width=\textwidth]{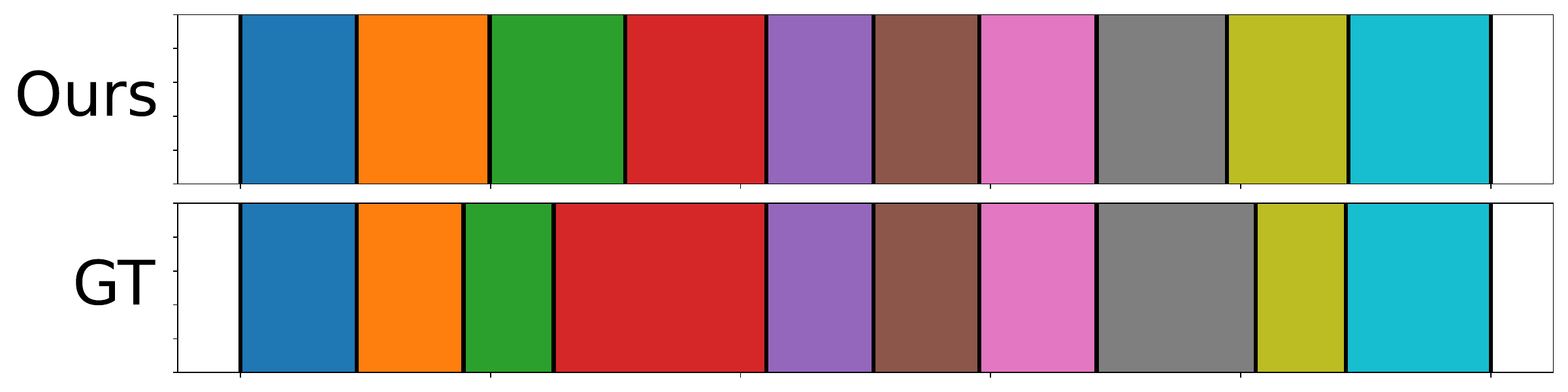}
    \end{subfigure}
    \caption{MAD Dataset}
    \end{subfigure}
    \begin{subfigure}[b]{0.33\linewidth}
    \begin{subfigure}[b]{\linewidth}
           \includegraphics[width=\textwidth]{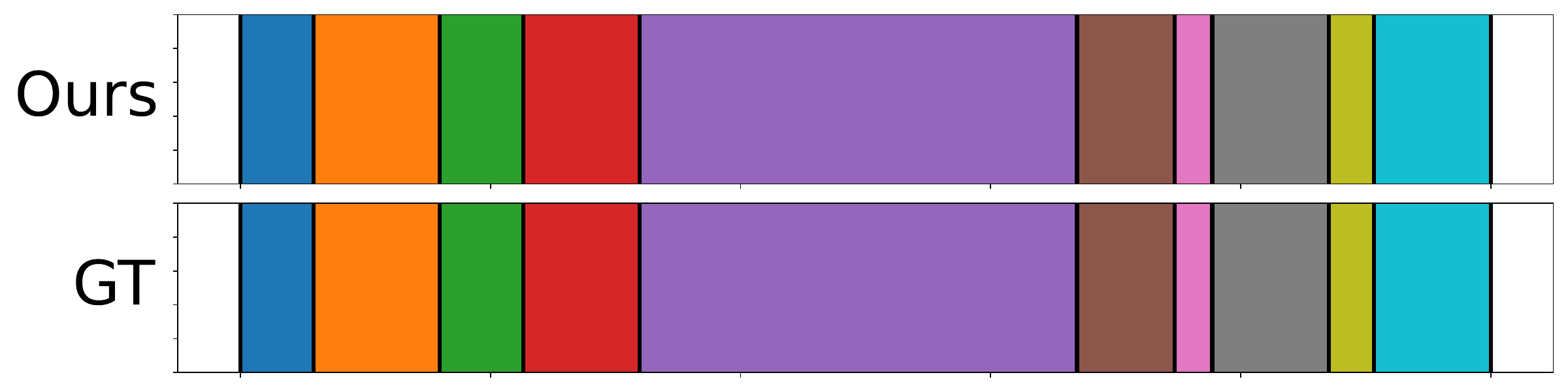}
    \end{subfigure}\\
    \begin{subfigure}[b]{\linewidth}
           \includegraphics[width=\textwidth]{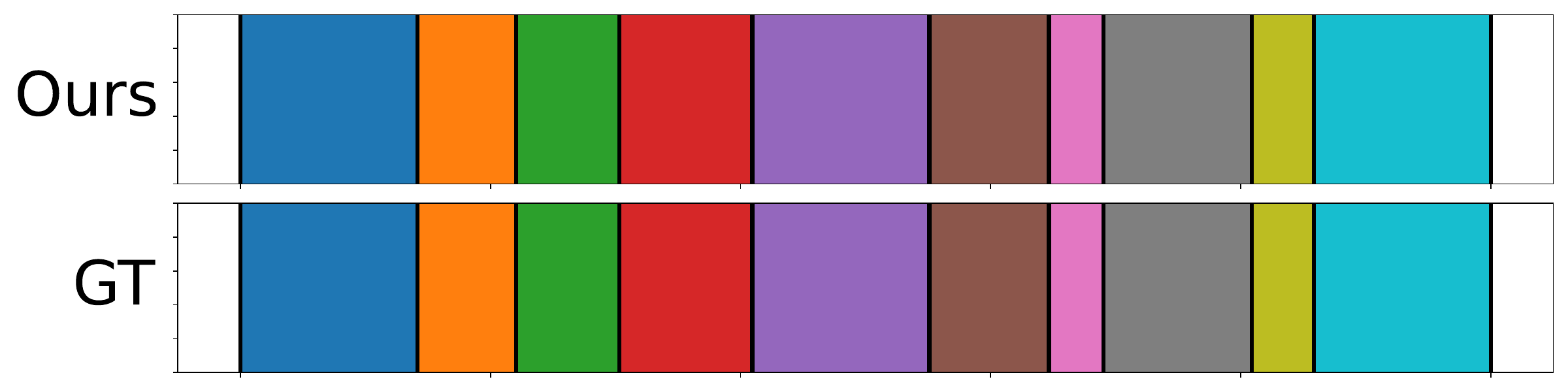}
    \end{subfigure}\\
    \begin{subfigure}[b]{\linewidth}
           \includegraphics[width=\textwidth]{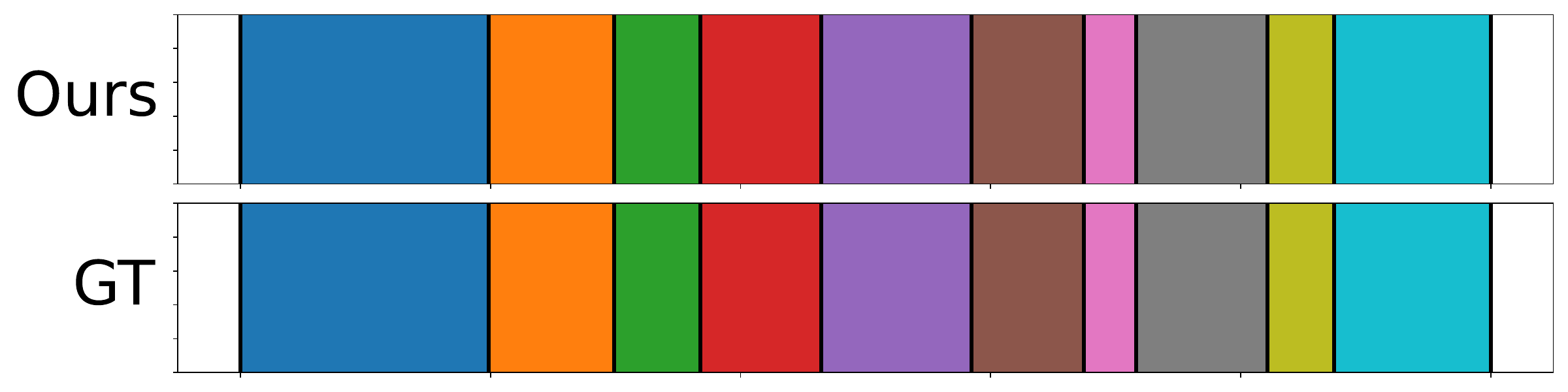}
    \end{subfigure}
    \caption{YouTube Dataset}
    \end{subfigure}
\caption{\textbf{Segmentation results visualization.} }
\label{fig:mask-visualization}
\end{figure*}

\subsection{Compared to CLIP Zero-Shot}
\label{sec:zero-shot}
Next, using the pretrained CLIP model, we explore the performance of zero-shot learning in the HMS task.
For the Weizmann dataset, we first convert the ground-truth labels of the dataset into textual descriptions of each motion.
For instance, the motion ``Wave1'' is described as ``A photo of people waving one hand.'' (see Table \ref{tab:Zero-shot text desc} for all the descriptions).
Then, we extract text embeddings for all the descriptions using a pretrained text encoder of CLIP.
For each frame in the dataset, we match its image embedding to the text embedding with the highest cosine similarity and assign the corresponding description as the zero-shot classification result for that frame.

\definecolor{myblue}{rgb}{0.29, 0.6, 0.78}

\begin{figure}[tbp]
\centering
\begin{minipage}[t]{0.4\textwidth}
    \centering
    \includegraphics[trim=0pt 50pt 0pt 30pt, clip, width=\linewidth]{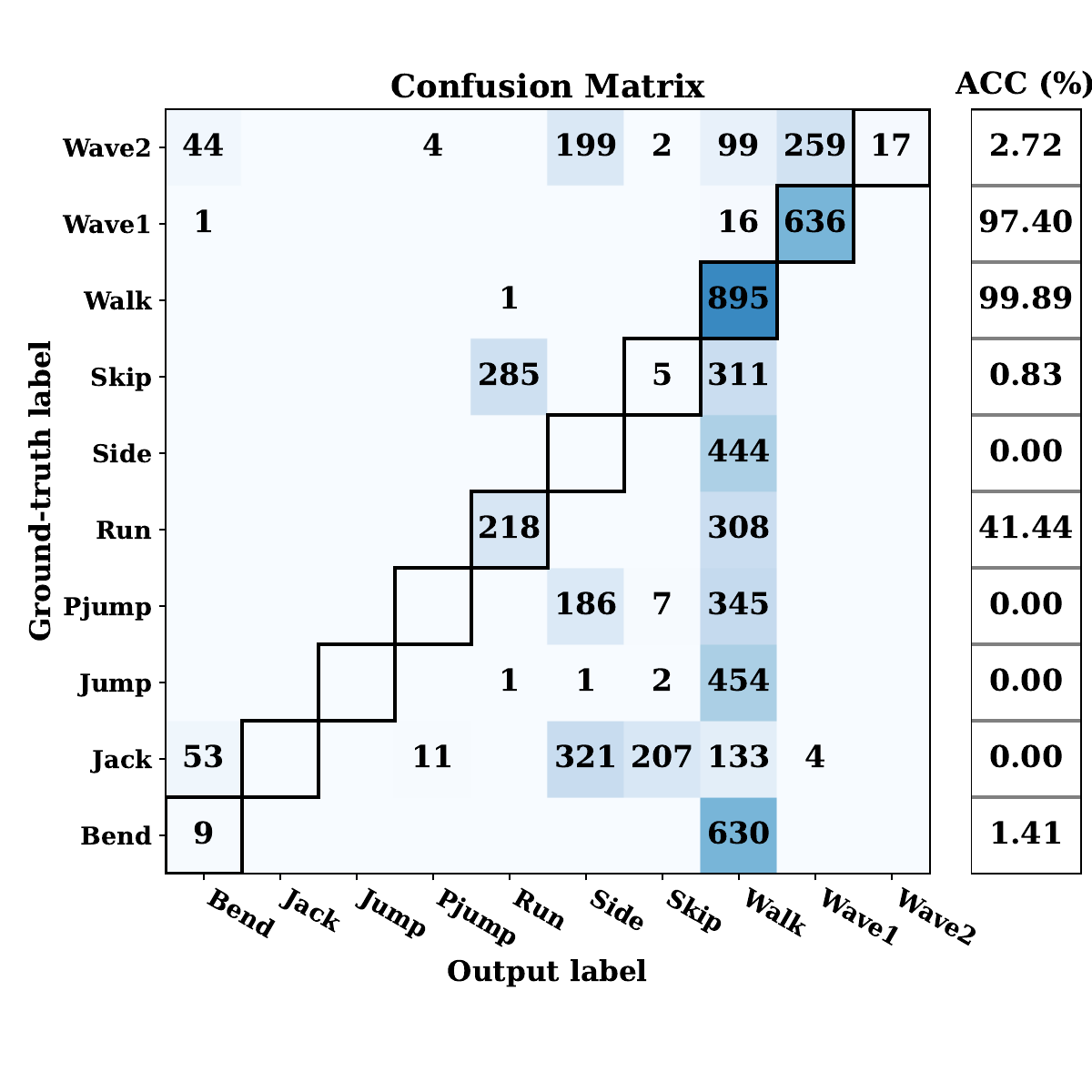}
    \caption{\textbf{Confusion matrix of zero-shot classification result for HMS task.}}
    \label{fig:confusion}
\end{minipage}%
\hfill
\begin{minipage}[t]{0.55\textwidth}
    \vspace{-2.2 in}
    \centering
    \captionof{table}{\textbf{Zero-shot classification with coarse label for HMS task.} The coarse ground-truth label is marked in \colorbox{myblue!40}{blue}.}
    \label{tab:zero-shot coarse label}
    \resizebox{0.9\linewidth}{!}{
    \begin{tabular}{>{\columncolor{myGray}}P{0.9cm}|P{1.0cm}P{1.0cm}P{1.0cm}P{1.0cm}P{1.0cm}|P{1.0cm}}
    \toprule
     \rowcolor{myGray} & \multicolumn{5}{c|}{Target} & ACC\\
    \rowcolor{myGray} \multirow{-2}{*}{GT} & Bend& Jump& Run& Walk&Wave & (\%)\\
    \midrule
         Bend&  \cellcolor{myblue!40}10&  &  &  629& & 1.56\\
         Jack&  83&  \cellcolor{myblue!40}270&  7&  327& 42 & 37.04\\
         Jump&  &  \cellcolor{myblue!40}&  &  458& & 0.00\\
         Pjump&  & \cellcolor{myblue!40} 107&  &  431& & 19.89\\
         Run&  &  & \cellcolor{myblue!40} 218&  308& & 41.44\\
         Side&  & \cellcolor{myblue!40} &  &  444& & 0.00\\
         Skip&  & \cellcolor{myblue!40} &  290&  311& & 0.00\\
         Walk&  &  &  1& \cellcolor{myblue!40} 895& & 99.89\\
         Wave1&  4&  &  &  44& \cellcolor{myblue!40}605 & 92.65\\
         Wave2&  77&  &  &  217& \cellcolor{myblue!40}330 & 52.88\\
    \bottomrule
    \end{tabular}
    }
\end{minipage}
\end{figure}

As shown in Figure~\ref{fig:confusion}, the classification accuracy for ``Walk'' and ``Wave1'' is 99.89\% and 97.40\%, respectively, making them two of the best-performing classes.
However, the overall classification accuracy is only 29.14\%, which is significantly lower than the performance of \trc$+$CLIP (96.21\%).
Notably, 63.31\% of frames are misclassified as ``Walk'' while no samples from the ``Jack'', ``Jump'', ``PJump'' and ``Side'' classes are correctly identified.
If we reduce the difficulty by using coarse labels consisting only of ``bend'', ``jump'', ``run'', ``walk'' and ``wave'' (Table~\ref{tab:zero-shot coarse label}), the accuracy of zero-shot classification for HMS is 39.87\%, which is still significantly lower than the performance of \trc$+$CLIP (96.21\%).

These results demonstrate that vanilla zero-shot classification is not suitable for HMS,  which due to the fact that zero-shot learning classifies frames individually, failing to capture in-context information.
In contrast, \trc$+$CLIP succeeds by learning temporally consistent representations that align with a union of orthogonal subspaces.

\begin{table}[h]
    \centering
\caption{\textbf{Textual description for zero-shot classification of Weizmann dataset.}}
\label{tab:Zero-shot text desc}
  \resizebox{0.9\linewidth}{!}{
    \begin{tabular}{lccc|cccc}
    \toprule
          \rowcolor{myGray} \#&Label&  Icon& Textual Description&\#&Label&  Icon& Textual Description\\
    \midrule
          1&Bend
&  \begin{minipage}[c][0.7cm][c]{1.5cm}
\centering
\includegraphics[width=0.7cm, height=0.7cm]{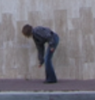}%
\includegraphics[width=0.7cm, height=0.7cm]{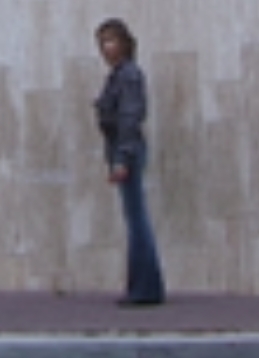}
\end{minipage}& A photo of people bending.&
 6&Side
& \begin{minipage}[c][0.7cm][c]{1.5cm}
\centering
\includegraphics[width=0.7cm, height=0.7cm]{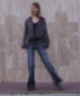}%
\includegraphics[width=0.7cm, height=0.7cm]{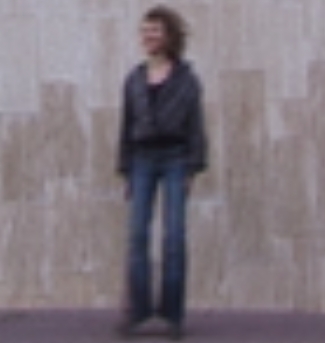}
\end{minipage} & A photo of people side jumping.\\
          2&Jack
&  \begin{minipage}[c][0.7cm][c]{1.5cm}
\centering
\includegraphics[width=0.7cm, height=0.7cm]{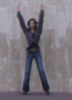}%
\includegraphics[width=0.7cm, height=0.7cm]{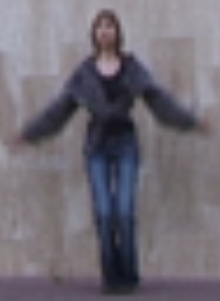}
\end{minipage}& A photo of people jumping jacks.&
7&Skip
& \begin{minipage}[c][0.7cm][c]{1.5cm}
\centering
\includegraphics[width=0.7cm, height=0.7cm]{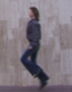}%
\includegraphics[width=0.7cm, height=0.7cm]{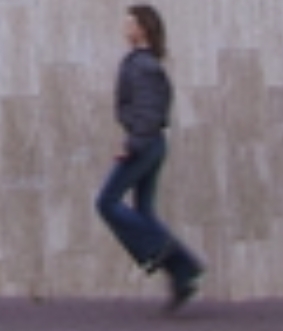}
\end{minipage} & A photo of people skipping jump.\\
          3&Jump
&  \begin{minipage}[c][0.7cm][c]{1.5cm}
\centering
\includegraphics[width=0.7cm, height=0.7cm]{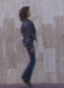}%
\includegraphics[width=0.7cm, height=0.7cm]{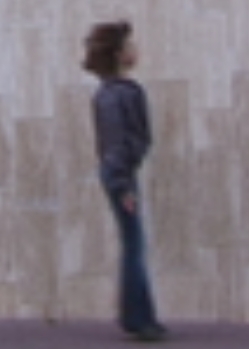}
\end{minipage}& A photo of people jumping.&
8&Walk
&\begin{minipage}[c][0.7cm][c]{1.5cm}
\centering
\includegraphics[width=0.7cm, height=0.7cm]{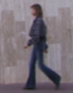}%
\includegraphics[width=0.7cm, height=0.7cm]{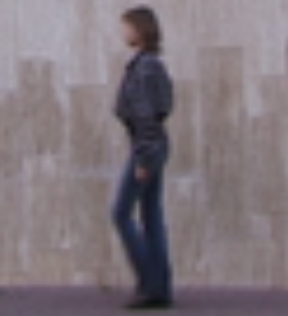}
\end{minipage}  & A photo of people walking.\\
          4&Pjump
& \begin{minipage}[c][0.7cm][c]{1.5cm}
\centering
\includegraphics[width=0.7cm, height=0.7cm]{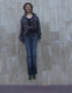}%
\includegraphics[width=0.7cm, height=0.7cm]{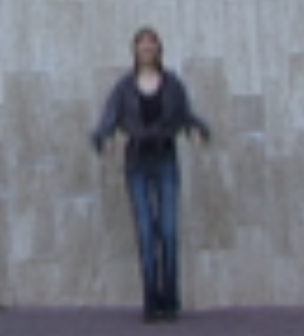}
\end{minipage} & A photo of people jumping in place.&
9&Wave1
& \begin{minipage}[c][0.7cm][c]{1.5cm}
\centering
\includegraphics[width=0.7cm, height=0.7cm]{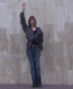}%
\includegraphics[width=0.7cm, height=0.7cm]{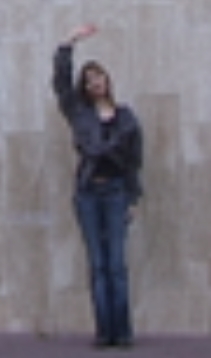}
\end{minipage} & A photo of people waving one hand.\\
          5&Run
& \begin{minipage}[c][0.7cm][c]{1.5cm}
\centering
\includegraphics[width=0.7cm, height=0.7cm]{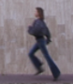}%
\includegraphics[width=0.7cm, height=0.7cm]{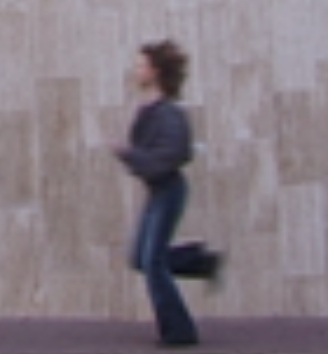}
\end{minipage} & A photo of people running.&    
  10&Wave2& \begin{minipage}[c][0.7cm][c]{1.5cm}
\centering
\includegraphics[width=0.7cm, height=0.7cm]{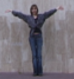}%
\includegraphics[width=0.7cm, height=0.7cm]{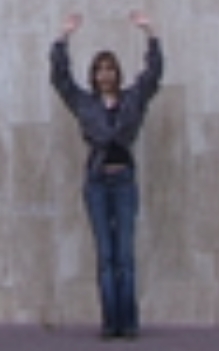}
\end{minipage}&A photo of people waving two hands.\\
\bottomrule
    \end{tabular}}
\end{table}

\subsection{Experimental Details for Temporal Action Segmentation}
\label{sec:action seg}

\myparagraph{Datasets description}
The Breakfast dataset~\cite{Kuehne:CVPR14-Breakfast} consists of $1,712$ videos capturing $52$ participants performing $10$ activities, including making friedegg, sandwich, pancake, \etal.
The YouTube Instructional dataset~\cite{Alayrac:CVPR16-Inria} consists of $150$ videos with $5$ activities capturing complex interactions between people and objects, including changing tire, making coffee, repotting, \etal.
The 50 Salads dataset~\cite{Stein:UC13-Salad} consists of $50$ videos capturing people preparing mixed salads from a top-down perspective.
We follow the baselines for the feature extractor selection of each dataset.
For the Breakfast and 50 Salads dataset, we use the Improved Dense Trajectory (IDT)~\cite{Wang:ICCV13-IDT} features provided by~\cite{Kukleva:CVPR19}; and for YouTube Instructional dataset, we use a concatenation of HOF descriptors~\cite{Laptev:CVPR08} and VGG features~\cite{Simonyan:ICLR15}.

\begin{table*}[tbh]
\caption{\textbf{Hyper-parameters configuration for training \trc{} on temporal action segmentation benchmark datasets.}}
\label{tab:action seg hyperpara}
\begin{center}
\resizebox{0.7\linewidth}{!}{
    \begin{tabular}{P{2.5cm}P{0.9cm}P{0.9cm}P{0.9cm}P{0.9cm}P{0.9cm}P{0.9cm}P{0.9cm}P{1.5cm}}
    \toprule
         \rowcolor{myGray} Dataset &  $d_{pre}$&  $d$&  $T$& $\lambda_1$& $\lambda_2$&  $s$& $\epsilon$&$\eta$\\
         \midrule
         Breakfast&  64&  64&  100& 0.05& 12&  2& 0.1&$10^{-3}$\\
         
         YouTube Instr.&  512&  64&  500& 0.05& 20&2& 0.05&$10^{-2}$\\
         
         50 Salads&  256&  64&  500& 0.05& 15&
         2&0.05&$10^{-2}$\\
 \bottomrule
    \end{tabular}
    }
\end{center}
\end{table*}

\myparagraph{Experimental details}
A significant distinction of the TAS benchmark datasets compared to that of HMS is that it contains a higher number of frames per video (\eg, the average number of frames per video of 50 Salads is $11,788$).
To address this discrepancy while maintaining computational tractability, we down-sample each video before training \trc{}, then up-sample the segmentation result back to the original number of frames.
Commonly used evaluation metrics, namely, Mean over Frames (MoF), F1-score, and Intersection over Union (IoU) are computed following the baselines.
The architecture of neural networks remains consistent with the experiments on HMS.
The hyper-parameters configuration of training \trc{} is listed in Table~\ref{tab:action seg hyperpara}.
When applying the state-of-the-art TAS methods to HMS datasets, we report the best results after tuning hyper-parameters which are picked from the Table~\ref{tab:action_seg_hyperpara_tuning}.

\begin{table}[htbp]
  \centering
  \caption{\textbf{Hyper-parameters tuning for temporal action segmentation methods on HMS datasets.}}
\label{tab:action_seg_hyperpara_tuning}
  \resizebox{0.85\linewidth}{!}{
    \begin{tabular}{lcc}
    \toprule
    \rowcolor{myGray} Method  & Hyper-parameters for Tuning\\
    \midrule
    TWF~\cite{Sarfraz:CVPR21}  & N/A (Automatic Clustering, no parameter to tune)\\
    ASOT~\cite{Xu:CVPR24} & $\alpha\in\{0.2,0.5\},r\in\{0.02,0.04,0.06,0.08,0.1\}, 
     \rho\in\{0.3,0.5,0.7\},\lambda\in\{0.08,0.11,0.14,0.17,0.2\}$ \\
    HVQ~\cite{Spurio:AAAI25}  & $\alpha\in\{1,2,3,4\},\lambda_{\text{rec}}\in\{0.0005,0.002,0.1\}$ \\
    \bottomrule
    \end{tabular}}
\end{table}%
\end{document}

%% file: preamble.tex
\usepackage{algorithm}
\usepackage{algpseudocode}

\usepackage{capt-of}

\newcommand{\myparagraph}[1]{\noindent\textbf{#1.}}

\def\ie{i.e.}
\def\eg{e.g.}
\def\mcr2{$\text{MCR}^2$}
\def\trc{$\text{TR}^2\text{C}$}
\def\boldPi{\boldsymbol{\Pi}}
\def\boldGamma{\boldsymbol{\Gamma}}
\def\x{\boldsymbol{x}}
\def\X{\boldsymbol{X}}
\def\Y{\boldsymbol{Y}}
\def\y{\boldsymbol{y}}
\def\Z{\boldsymbol{Z}}
\def\z{\boldsymbol{z}}

\def\W{\boldsymbol{W}}

\usepackage{amsmath}
\usepackage{amssymb}
\usepackage{mathtools}
\usepackage{amsthm}

\usepackage{multirow}
\usepackage{booktabs}
\usepackage{tabularx}
\newcolumntype{P}[1]{>{\centering\arraybackslash}p{#1}}

\usepackage{subcaption}

\DeclareMathOperator{\Diag}{Diag}

\DeclareMathOperator{\tr}{tr}

\def\mymaketitlesupplementary
   {
   \newpage
   \begin{center}
       \Large
       \textbf{\thetitle}\\
       \vspace{0.5em}
       Supplementary Material \\
       \vspace{1.0em}
   \end{center}
}